\documentclass[pdflatex,sn-mathphys-num]{sn-jnl}% Math and Physical Sciences 
\usepackage{graphicx}%
\usepackage{multirow}%
\usepackage{amsmath,amssymb,amsfonts}%
\usepackage{amsthm}%
\usepackage{mathrsfs}%
\usepackage[title]{appendix}%
\usepackage{xcolor}%
\usepackage{textcomp}%
\usepackage{manyfoot}%
\usepackage{booktabs}%
\usepackage{algorithm}%
\usepackage{algorithmicx}%
\usepackage{algpseudocode}%
\usepackage{listings}%
\usepackage{cleveref}
\usepackage{subcaption}
\usepackage{tabularx}
\usepackage{xurl}
\usepackage{hhline}
\usepackage{multirow}
\usepackage{array}
\usepackage[T1]{fontenc}
\usepackage[utf8]{inputenc}
\theoremstyle{thmstyleone}%
\theoremstyle{thmstyletwo}%

\theoremstyle{thmstylethree}%

\begin{document}

\title[Article Title]{Hidden Axis of Uncertainty: Latent-Posterior Alignment in Graph Neural Networks with Bayesian Output Layers}

% Latent Alignment Governs Uncertainty in Bayesian Neural Networks
% The Hidden Axis of Uncertainty: Latent Alignment in Bayesian Neural Networks
% Beyond the Posterior: Latent Dynamics as the Key to Uncertainty in BNNs

%%=============================================================%%
%% GivenName	-> \fnm{Joergen W.}
%% Particle	-> \spfx{van der} -> surname prefix
%% FamilyName	-> \sur{Ploeg}
%% Suffix	-> \sfx{IV}
%% \author*[1,2]{\fnm{Joergen W.} \spfx{van der} \sur{Ploeg} 
%%  \sfx{IV}}\email{iauthor@gmail.com}
%%=============================================================%%

\author[1]{\fnm{Suk Hoon} \sur{Choi}}\email{inuk97@kist.re.kr}

\author[1]{\fnm{Damdae} \sur{Park}}\email{damdaepark@kist.re.kr}

\author[2]{\fnm{Junhyuk} \sur{Choi}}\email{wnsgurchl@korea.ac.kr}

\author[1]{\fnm{Hyein} \sur{Jung}}\email{976hannah@kist.re.kr}

\author*[1, 3]{\fnm{Changsoo} \sur{Kim}}\email{changs90.kim@kist.re.kr}

\author*[1, 3]{\fnm{Ung} \sur{Lee}}\email{ulee@kist.re.kr}

\author*[1, 3]{\fnm{Kyeongsu} \sur{Kim}}\email{kyeongsu@kist.re.kr}

\affil[1]{Clean Energy Research Center, Korea Institute of Science and Technology, Seoul 02792, Republic of Korea}
\affil[2]{Department of Chemical and Biological Engineering, Korea University, Seoul 02841, Republic of Korea}
\affil[3]{Division of Energy \& Environment Technology, Korea University of Science and Technology, 217 Gajeong-ro, Yuseong-gu, Daejeon 34113, Republic of Korea}

\abstract{
Bayesian Neural Networks (BNNs) with Bayesian output layers provide a principled and tractable framework for quantifying predictive uncertainty, yet the mechanisms shaping that uncertainty remain unclear. While conventional theory attributes uncertainty reduction to posterior contraction, the corresponding assumptions need not hold for deep models. In the Graph Neural Networks (GNNs) with Bayesian output layers studied here, we observe that predictive uncertainty decreases as latent representations shift toward lower-variance posterior directions, even though the posterior variance does not contract. We term this behavior Latent-Posterior Alignment (LPA) and conduct interventional experiments that support its functional role in shaping predictive uncertainty. Building on this insight, we propose Alignment-Guided Learning (AGL), which explicitly promotes this alignment during training. AGL effectively reduces predictive uncertainty while preserving accuracy and improves structural calibration, ensuring that the model confidence faithfully mirrors underlying data density. These findings provide a new perspective on uncertainty dynamics in GNNs with mean-field Bayesian output layers, shifting the focus from the magnitude of the posterior to the geometric interplay between latent and parameter spaces.
}


\maketitle

\section{Introduction}
\label{sec1}
Quantifying predictive uncertainty has become essential as deep learning systems are increasingly deployed in scientific and engineering domains where data are scarce \cite{chakraborti2025personalized, chen2025uncertainty}. The challenges are particularly pronounced in safety-critical and scientific applications, such as autonomous driving \cite{bojarski2016end, codevilla2018end, yurtsever2020survey}, medical diagnostics \cite{esteva2017dermatologist, mckinney2020international}, and accelerated material discovery \cite{jumper2021highly, sanchez2018inverse, merchant2023scaling}. To address these challenges, Bayesian Neural Networks (BNNs) have emerged as a principled framework for modeling the uncertainty of deep learning architectures \cite{neal2012bayesian, lin2023uncertainty, jospin2022hands}. Common approximate Bayesian approaches include Mean-Field Variational Inference (Bayes-by-Backprop) \cite{blundell2015weight} and Monte Carlo dropout \cite{gal2016dropout}, while Deep Ensembles \cite{lakshminarayanan2017simple} provide a widely used non-Bayesian alternative. These approaches share the practical goal of quantifying predictive uncertainty, or “what a model does not know” \cite{kendall2017uncertainties, hullermeier2021aleatoric}.

% Although various approaches exist for implementing BNNs, including Mean-Field Variational Inference (Bayes-by-Backprop) \cite{blundell2015weight}, Monte Carlo dropout (MC-Dropout) \cite{blundell2015weight}, and Deep Ensembles \cite{lakshminarayanan2017simple}, they share a fundamental goal—approximating posterior distributions of the data to quantify “what the model does not know” \cite{kendall2017uncertainties, hullermeier2021aleatoric}.

Conventionally, the reduction of predictive uncertainty in Bayesian models is attributed to the contraction of the posterior distribution. Grounded in the classical Bernstein-von Mises theorem \cite{van2000asymptotic, freedman1999wald}, it is assumed that as data accumulates, the posterior asymptotically converges to a Gaussian distribution centered at the true parameter, causing the variance to vanish \cite{blundell2015weight, wenzel2020good, ghosal2000convergence}. However, this theoretical guarantee is predicated on strict assumptions—that the model is well-specified (i.e., it can perfectly replicate the true data generating process) \cite{kleijn2012bernstein}, the parameter space is finite-dimensional and fixed \cite{freedman1999wald}, and the Fisher information matrix is positive definite \cite{watanabe2009algebraic}—conditions that may not be satisfied in modern deep-learning settings \cite{dar2021farewell, bochkina2019bernstein}. Consequently, the classical assurance that “more data yields tighter posteriors and lower uncertainty” becomes theoretically tenuous in this regime \cite{kleijn2012bernstein, bochkina2019bernstein, masegosa2020learning}.

Recent empirical studies further show that uncertainty behavior in deep Bayesian models can depart from posterior-contraction intuitions. Analyses of loss landscapes reveal that posteriors in deep BNNs are highly complex and multimodal, fundamentally differing from the simple compact distributions assumed by the classical Bayesian theory \cite{izmailov2021bayesian, foong2020expressiveness}. Furthermore, studies on the “Cold Posterior” effect reveal that BNN posteriors do not exhibit the expected contraction and often necessitate artificial sharpening for generalization, underscoring a fundamental mismatch between Bayesian theory and the behavior of deep models \cite{wenzel2020good, aitchison2020statistical, nabarro2022data}. Similarly, out-of-distribution (OOD) benchmarks demonstrate that BNNs frequently exhibit overconfident predictions in regions with low data density, where the uncertainty is expected to be high and the posterior remains diffuse \cite{ovadia2019can, antoran2020depth, mukhoti2023deep, de2025deep, fan2024reducing}. This misalignment violates the fundamental correspondence between data density and uncertainty. Crucially, such failures propagate to downstream applications. In active learning, uncertainty-based acquisition strategies frequently fail to outperform simple random sampling, as they do not faithfully reflect data scarcity \cite{yao2019quality, rakesh2021efficacy}. Likewise, in Bayesian optimization, the lack of distance-awareness obscures informative intermediate regions, directly hindering efficient exploration \cite{foong2019between}.

These limitations are particularly consequential in chemical and materials applications, where predictive models frequently operate under low-data and distribution-shifted regimes. In these domains, uncertainty estimates are not merely auxiliary outputs but often directly influence decision-making processes such as molecular screening, active learning, and experimental prioritization. Among the available architectures, graph neural networks (GNNs) have become a common backbone for machine-learning interatomic potentials (MLIPs) and molecular property prediction tasks due to their ability to encode atomistic structures and chemical interactions. Consequently, deficiencies in uncertainty estimation—such as poor density awareness or overconfident predictions in sparse regions—can propagate through downstream workflows and hinder efficient exploration of chemical space. Motivated by these practical considerations, we focus on Bayesian graph neural networks (BGNNs) employing Bayesian output layers and revisit how predictive uncertainty is formed within these architectures. 

Motivated by these theoretical and practical considerations, we revisit the dynamics of uncertainty in BNNs. We focus on a class of models consisting of a deterministic feature extractor followed by a Bayesian output layer trained via Bayes-by-Backprop, and conduct our analysis within this formulation. Contrary to the conventional expectation that predictive uncertainty decreases through posterior contraction, we observe that predictive uncertainty decreases with increasing data even as the weight posterior variance broadens. This decoupled behavior calls for a re-examination of the underlying mechanism. To this end, we identify a key mechanism, termed Latent–Posterior Alignment (LPA), which explains how predictive uncertainty is reduced. Through analytical derivation and empirical analysis, we show that the model minimizes predictive variance not by tightening its posteriors, but by aligning its latent representation $\mathbf{z}$ along the low-variance directions of the posterior.

We probe the causal role of this mechanism through interventional experiments: structurally disrupting the LPA via latent regularization consistently amplifies predictive uncertainty, supporting a causal contribution of LPA to model confidence. Building on this observation, we introduce Alignment-Guided Learning (AGL), a training framework that explicitly promotes this alignment. We show that AGL reduces predictive uncertainty while preserving accuracy and significantly improves structural calibration, illustrating that model confidence more faithfully reflects the underlying data density. This improved alignment between output uncertainty and data density suggests that AGL can make the model’s predictive confidence more consistent with the local density of the training data. Such density awareness is particularly valuable in chemical and material domain including molecular modeling and MLIPs, where uncertainty estimates can help identify weakly supported structures and prioritize additional calculations or experiments. Collectively, our findings provide a new perspective on uncertainty in models with Bayesian output layers, emphasizing the role of LPA over the magnitude of the posterior.

% \newpage

\section{Results} \label{sec_2_0}
To investigate the mechanisms determining predictive uncertainty, we designed a Bayesian Graph Neural Network (BGNN), as shown in Fig. \ref{fig_1}a. This architecture combines a deterministic Graph Isomorphism Network (GIN) \cite{xu2018powerful, morris2019weisfeiler}, which learns latent representations from molecular structures, with a Bayesian linear output layer for uncertainty quantification \cite{blundell2015weight, tran2019bayesian, dusenberry2020efficient, gawlikowski2023survey}. Previous studies suggest that Bayesian last-layer models can retain predictive performance and provide uncertainty estimates comparable to more extensively Bayesianized networks in the settings examined, at substantially lower computational cost \cite{harrison2024variational, zeng2018relevance}. Although this approach does not capture parameter uncertainty in the deterministic feature extractor, it provides a practical and analytically tractable framework for decomposing predictive variance into representation and output-weight contributions. We therefore adopt this formulation to investigate uncertainty formation in GNNs with Bayesian output layers. To ensure the robustness and generality of our observations, we conducted experiments across six molecular property prediction benchmarks. In the main text, we present results for the partition coefficient dataset \cite{mansouri2018opera} as a representative example, while results for the remaining datasets are provided in the Supplementary Information.

In this section, we present a series of findings that revisit the conventional understanding of uncertainty in this class of BNNs. We begin by identifying a counterintuitive behavior, where predictive uncertainty decreases with increasing data density despite the expansion of the posterior variance (Section \ref{sec_paradox}). To explain this phenomenon, we introduce LPA as the central mechanism. We show that the geometry of the latent representation—rather than the magnitude of the posterior variance—plays a central role in determining predictive uncertainty (Section \ref{sec_laa}). We then probe the causal contribution of LPA through interventional experiments that perturb the latent-space structure and demonstrate that this alignment-based mechanism persists under diverse probabilistic settings (Section \ref{sec_test}). Finally, we propose Alignment-Guided Learning (AGL), a training strategy that explicitly promotes this geometric alignment, reducing average predictive uncertainty and strengthening density-uncertainty dependence while preserving predictive accuracy (Section \ref{sec_agl}).

\subsection{Rethinking uncertainty behavior in Bayesian neural networks} \label{sec_paradox}
% \begin{figure}[]
%     \centerline{\includegraphics[width=\columnwidth]{fig/1/img_1_paradox_laa_new.pdf}}
%     \caption{Uncertainty behavior and Latent-Posterior Alignment (LPA). (a) Schematic of the BGNN architecture. (b–g) Deviation from conventional expectation. As training data increases, predictive error (b) and uncertainty decrease (c), whereas weight posterior uncertainty expands (d), leading to broader distributions in the data-rich regime (f, red) compared to the data-poor regime (f, blue). The posterior means remain near zero on average (e), while both the NLL and KL terms decrease (g). (h-m) Latent-Posterior Alignment. Normalized latent vectors ($\tilde{\mathbf{z}}$, blue) progressively align with low-variance weight dimensions ($\tilde{\sigma}_{w}$, orange) as data accumulates (1\% $\to$ 80\%). Grey bars indicate the relative uncertainty contribution. (k) High-uncertainty samples (green) exhibit amplified latent magnitudes compared to the full data distribution (blue) while maintaining the alignment. (l) t-SNE projection of latent vectors colored by predictive uncertainty. (m) Monotonic increase in the Latent-Posterior Alignment Score (LPAS), quantifying the alignment mechanism. Shaded areas represent standard deviation across $n=30$ independent runs. “Number of data (\%)” denotes the percentage of the available training set used to train each model. }
%     \label{fig_1} 
% \end{figure}

\begin{figure}[]
    \centerline{\includegraphics[width=\columnwidth]{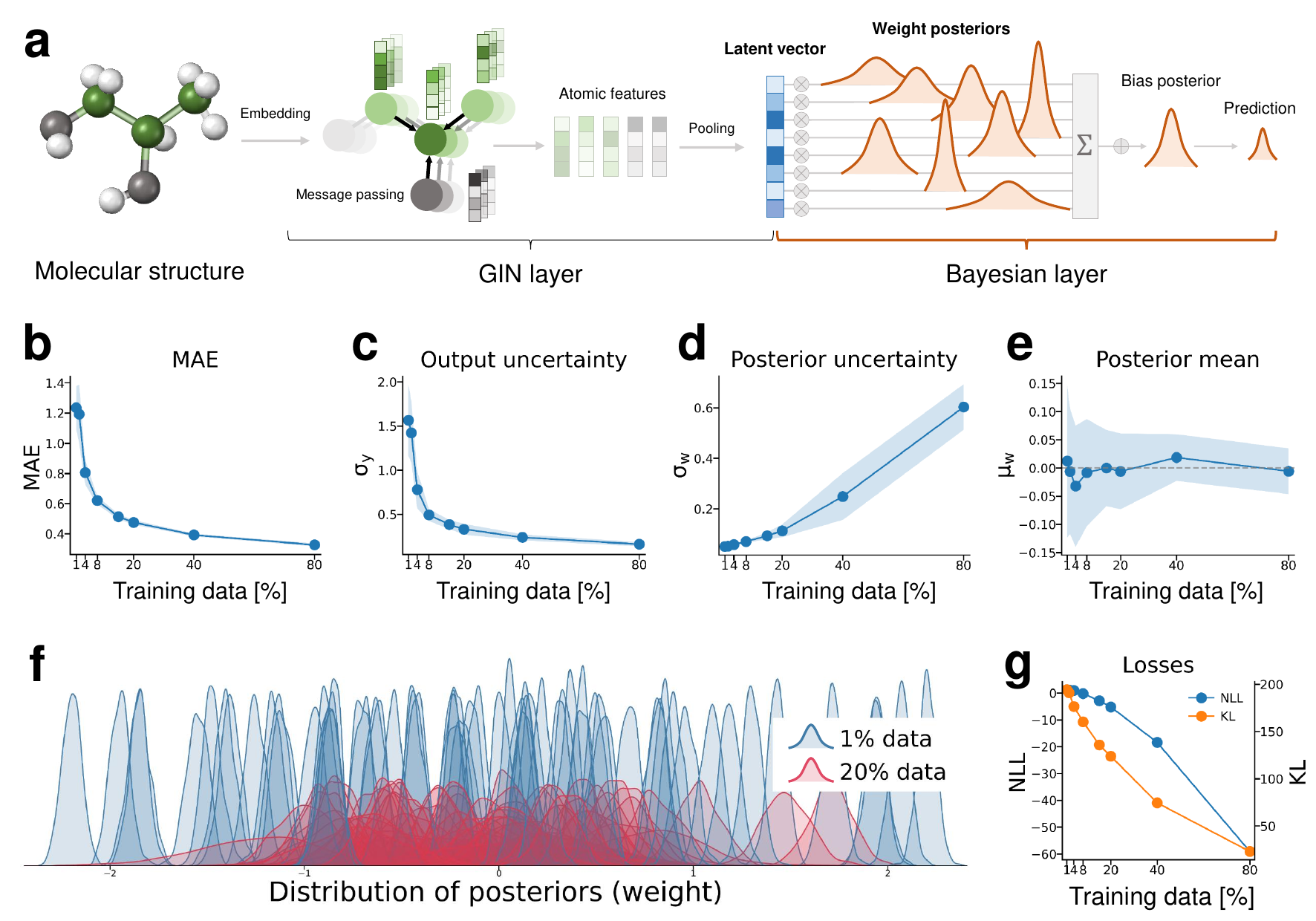}}
    \caption{Uncertainty behavior of BGNN. (a) Schematic of the BGNN architecture. (b–g) Deviation from conventional expectation. As training data increases, predictive error (b) and uncertainty decrease (c), whereas weight posterior uncertainty expands (d), leading to broader distributions in the data-rich regime (f, red) compared to the data-poor regime (f, blue). The posterior means remain near zero on average (e), while both the NLL and KL terms decrease (g). Shaded areas represent standard deviation across $n=30$ independent runs. “Training data [\%]” denotes the percentage of the available training set used to train each model.}
    \label{fig_1} 
\end{figure}

Consistent with the Bernstein-von Mises theorem \cite{van2000asymptotic, freedman1999wald} and fundamental Bayesian theory \cite{bishop2006pattern, mackay1992practical, graves2011practical}, we first confirm that as the amount of training data increases, the model's predictive performance (MAE, Fig. \ref{fig_1}b) improves and its overall predictive uncertainty monotonically decreases ($\sigma_{y}$, Fig. \ref{fig_1}c). Conventionally, this reduction is attributed to posterior contraction $p(\theta \vert \mathcal{D})$, where $\theta$ and $\mathcal{D}$ denote the model parameters and dataset, respectively. As empirical evidence accumulates, the posterior variance is theoretically expected to shrink, making the model increasingly confident with regard to its parameter estimates \cite{blundell2015weight, wenzel2020good, ghosal2000convergence}. Following the standard uncertainty decomposition framework \cite{kendall2017uncertainties, hullermeier2021aleatoric, depeweg2018decomposition}, the predictive variance $\mathrm{Var}\left[y \vert x, \mathcal{D} \right]$, given an input $x$ and target output $y$, can be decomposed as:

\begin {align} 
\mathrm{Var}\left[y \vert x, \mathcal{D} \right] &= \underbrace{\mathbb{E}_{p(\theta \vert \mathcal{D})} \left[\mathrm{Var} \left[y \vert x, \theta \right] \right]}_{\text{Aleatoric}} + \underbrace{\mathrm{Var}_{p(\theta \vert \mathcal{D})} \left[ \mathbb{E} \left[y \vert x, \theta \right] \right]}_{\text{Epistemic}} 
\end {align}

Here, the first term represents aleatoric uncertainty, while the second captures the epistemic uncertainty arising from uncertainty in model parameters. Under the standard Bayesian paradigm, the posterior variance $\mathrm{Var} [\theta \vert \mathcal{D}]$ is expected to diminish as the dataset grows, causing the epistemic term to vanish \cite{graves2011practical, hullermeier2021aleatoric}. Thus, the reduction in predictive uncertainty is conventionally understood to result from this tightening of the posterior \cite{blundell2015weight, wenzel2020good, ghosal2000convergence, hullermeier2021aleatoric}. However, many modern deep BNNs do not clearly satisfy the regularity conditions under which such asymptotic contraction is guaranteed \cite{dar2021farewell, bochkina2019bernstein}, potentially leading to behaviors that diverge from classical expectations \cite{kleijn2012bernstein, bochkina2019bernstein} (see Supplementary Note \ref{si_analysis} for more details).

% However, it is crucial to note that modern deep BNNs violate the fundamental conditions required for such asymptotic contraction \cite{dar2021farewell, bochkina2019bernstein}, potentially leading to behaviors that diverge from classical expectations \cite{kleijn2012bernstein, bochkina2019bernstein} (see Supplementary Note \ref{si_analysis} for more details).

Consistent with this theoretical perspective, our analysis of the Bayesian output layer reveals a systematic deviation from classical expectations. Contrary to the conventional expectation that larger datasets yield constricted posteriors, we observed that the uncertainty of the weight posterior—computed as the mean standard deviation across all hidden dimensions—increases with dataset size (Fig. \ref{fig_1}d). This posterior broadening is further visualized in Fig. \ref{fig_1}f, where the posterior uncertainty increases from 0.05 (1$\%$ data, blue) to 0.11 (20$\%$ data, red). One possible interpretation is that the learned representation and the variational output posterior co-adapt during training. As additional data improve the representation, predictive information can be redistributed toward directions that are less sensitive to uncertain output weights. This interpretation motivates examining the coupled latent-posterior geometry, although the present observation alone does not establish why the posterior scale broadens. Although the bias posteriors fluctuate, they exhibit no consistent trend and show negligible influence on predictive uncertainty; consequently, our subsequent analysis primarily focuses on the weight posteriors (see Supplementary Note \ref{si_bias}). This decoupling is also reflected in the training dynamics shown in Fig. \ref{fig_1}g. With increasing training data, the Negative Log-Likelihood (NLL) decreases, reflecting improved predictive performance and confidence. At the same time, the Kullback-Leibler (KL) divergence term decreases, indicating that the variational posterior moves closer to the $\mathcal{N}(0, 1)$ prior in KL divergence. Notably, the posterior means remain centered around zero across all dataset sizes (Fig. \ref{fig_1}e), indicating that the reduction in KL divergence is primarily driven by posterior variance expanding toward the unit variance imposed by the prior.

Taken together, these findings indicate a decoupling between posterior uncertainty and predictive uncertainty: the model achieves lower predictive uncertainty even as the posterior becomes more diffuse. This decoupling implies that the reduction in predictive uncertainty cannot be attributed to the constriction of the weight posterior, necessitating an alternative governing mechanism. Importantly, this behavior is consistently observed across five additional molecular-property datasets (see Supplementary Fig. \ref{si_fig_paradox}), demonstrating its robustness.

% \clearpage
\subsection{Latent-posterior alignment mechanism} \label{sec_laa}
To explain the observed decoupling between posterior behavior and predictive uncertainty, we examine the analytical form of the predictive variance. The output $y$ of the BGNN, derived from the Bayesian linear layer, is given by \cite{blundell2015weight}:

\begin{align} 
y = \mathbf{z}^\top \mathbf{w} + b, \quad \mathbf{w} \sim q_{\phi}(\mathbf{w}), \quad b \sim q_{\phi}(\mathbf{b})
\end{align}

where the $d$-dimensional latent vector $\mathbf{z} \in \mathbb{R}^d$, computed by the GIN layers, serves as the input to the Bayesian layer, and the weight $\mathbf{w}$ and bias $b$ are sampled from their respective posterior distributions, $q_{\phi}(\mathbf{w})$ and $q_{\phi}(\mathbf{b})$. By assuming a mean-field approximation \cite{blundell2015weight, blei2017variational}, where the weights are independent across dimensions, and noting the negligible contribution of the bias (Supplementary Note \ref{si_bias}), the predictive variance can be approximated as:

\begin{align} 
\mathrm{Var}\left[y \vert \mathbf{z}, \mathcal{D} \right] &\approx \mathbf{z}^\top \mathrm{Cov}\left[\mathbf{w} \vert \mathcal{D}\right] \mathbf{z} \\
&\approx \sum_{i=1}^{d} z_{i}^{2} \sigma_{w,i} ^ {2} \label{eq_pred_uncert}
\end{align}

where $\sigma_{w, i}$ and $d$ denote the standard deviation of the $i$-th weight posterior and the number of latent dimensions, respectively. This expression shows that predictive uncertainty depends on the interaction between the latent representation $\mathbf{z}$ and the posterior variance $\sigma_{w}^{2}$. In particular, uncertainty can remain low even when the overall posterior variance is large, provided that $\mathbf{z}$ is concentrated along low-variance directions. Although Eq. \ref{eq_pred_uncert} directly implies that predictive variance depends jointly on latent activations and posterior variance, the emergence of systematic latent reorganization during training is not obvious. Based on this observation, we propose \textit{Latent-Posterior Alignment (LPA)} as the primary mechanism that governs predictive uncertainty. This means that the model reduces uncertainty not by uniformly decreasing posterior variances ($\sigma_{w, i}^{2}$), but by strategically aligning the latent representation $\mathbf{z}$ with stable (low-variance) posterior directions. Concretely, the model learns to amplify components of $\mathbf{z}$ in dimensions with small $\sigma_{w, i}$ (i.e., “reliable” dimensions), while suppressing those associated with large variance (i.e., “unstable” dimensions).

To quantify this structural behavior, we define the \textit{Latent-Posterior Alignment Score (LPAS)} as:

\begin{align} 
\text{Latent-Posterior Alignment Score (LPAS)} = 1-\frac{\sum_{i=1}^{d} |z_{i}|\sigma_{i}}{\sum_{i=1}^{d}|z_{i}|\sum_{i=1}^{d}\sigma_{i}} 
\end{align}

A higher LPAS indicates strong alignment, where the latent vector selectively utilizes stable, low-variance dimensions to reduce uncertainty. Further insights into the relationship between the LPAS and uncertainty are provided in Supplementary Note \ref{si_LPAS_uncertainty}. We use LPAS as the primary alignment score because higher values intuitively indicate stronger alignment. For baseline-normalized AGL comparisons presented later, relative improvement is reported as the reduction in its complement, $1-\mathrm{LPAS}$, which contains the same information in the opposite direction and avoids compressed percentage changes when LPAS is close to one.

\begin{figure}[]
    \centerline{\includegraphics[width=\columnwidth]{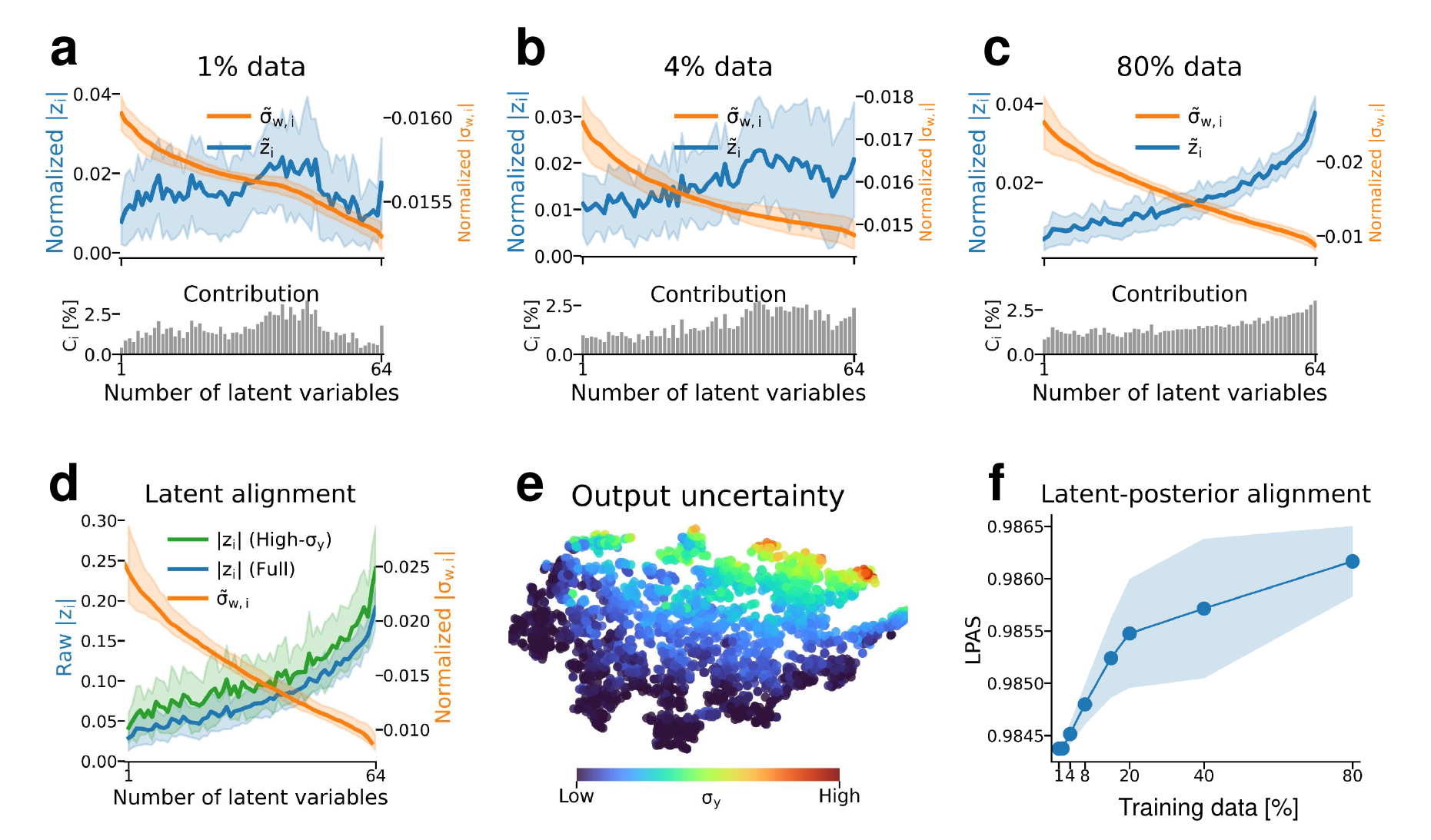}}
    \caption{Latent-Posterior Alignment (LPA) of BGNN. (a-c) Normalized latent vectors ($\tilde{\mathbf{z}}$, blue) progressively align with low-variance weight dimensions ($\tilde{\sigma}_{w}$, orange) as data accumulates (1\% $\to$ 80\%). Grey bars indicate the relative uncertainty contribution. (d) High-uncertainty samples (green) exhibit amplified latent magnitudes compared to the full data distribution (blue) while maintaining the alignment. (e) t-SNE projection of latent vectors colored by predictive uncertainty. (f) Monotonic increase in the Latent-Posterior Alignment Score (LPAS), quantifying the alignment mechanism. Shaded areas represent standard deviation across $n=30$ independent runs.}
    \label{fig_1_2} 
\end{figure}

Empirical results strongly support this alignment mechanism. Fig. \ref{fig_1_2}a-c visualize the normalized absolute latent vectors $\tilde{z}_{i}$ and posterior standard deviations $\tilde{\sigma}_{w,i}$ (see Methods) across $d=64$ hidden dimensions for models trained with 1\%, 4\%, and 80\% of the dataset, respectively. The $\tilde{\sigma}_{w,i}$ values are sorted in descending order for visual clarity. As the amount of training data increases, the latent representations progressively shift toward dimensions with lower posterior variance, leading to stronger LPA. This structural change of the latent representations directly impacts predictive uncertainty. We analyze the contribution of each dimension to the total uncertainty $C_{i}$ (see Methods), shown as gray bars in Fig. \ref{fig_1_2}a–c. As alignment strengthens, the contribution becomes increasingly concentrated in low-variance dimensions, reflecting a shift in how predictive uncertainty is distributed across latent directions.

Importantly, this global alignment does not eliminate local variability. As shown in Fig. \ref{fig_1_2}d, after training each model, we ranked all test samples within each independent run by posterior-predictive standard deviation and defined the top 5\% as the high-uncertainty subset (green). Selected post hoc from the predictions of the same trained model, this subset follows the overall LPA trend of the full test set (blue) but exhibits larger latent-vector magnitudes; no separate model was trained for this analysis. A t-SNE projection (Fig. \ref{fig_1_2}e) reveals that these high-uncertainty samples lie near the boundaries of the data manifold (red), where data density is low and sparsely populated. This suggests that while the LPA controls the overall uncertainty reduction, the magnitude of the latent vector is adjusted to capture instance-level uncertainty. This behavior is reflected in the alignment score. As shown in Fig. \ref{fig_1_2}f, the LPAS increases with data density, showing stronger alignment that closely tracks the reduction in predictive uncertainty. Taken together, these results demonstrate that LPA acts as the major mechanism for uncertainty regulation, rather than a secondary effect. Additional experiments across five independent datasets further confirm the consistency of this behavior, showing similar trends in LPA and corresponding changes in predictive uncertainty (see Supplementary Note \ref{si_anti_alignment}).

% \clearpage
\subsection{Probing the causal evidence for latent–posterior alignment} \label{sec_test}
Having established the association between LPA and predictive uncertainty, we next test whether LPA contributes causally to predictive uncertainty by deliberately perturbing the alignment structure. To this end, we perform an interventional experiment that directly perturbs the latent structure. Specifically, we introduce explicit regularization on the latent representation $\mathbf{z}$, including L1, L2, and anti-alignment penalties (see Supplementary Note \ref{si_z_regularization_formulation} for details). These constraints are designed to disrupt the alignment between the latent representation and the low-variance posteriors. If LPA governs the uncertainty, then preventing such alignment should lead to an increase in predictive uncertainty.

\begin {align}
\mathcal{L}_{L1} = -ELBO + \lambda_{l1} \| z \|_{1}
\end {align}

\begin {align}
\mathcal{L}_{L2} = -ELBO + \lambda_{l2} \| z \|_{2}
\end {align}

\begin {align}
\mathcal{L}_{anti-align} = -ELBO - \lambda_{anti-align} \frac{\sum_{i=1}^{d} |z_{i}|\sigma_{i}}{\sum_{i=1}^{d}|z_{i}|\sum_{i=1}^{d}\sigma_{i}}
\end {align}

\begin{figure}[H]
    \centerline{\includegraphics[width=\columnwidth]{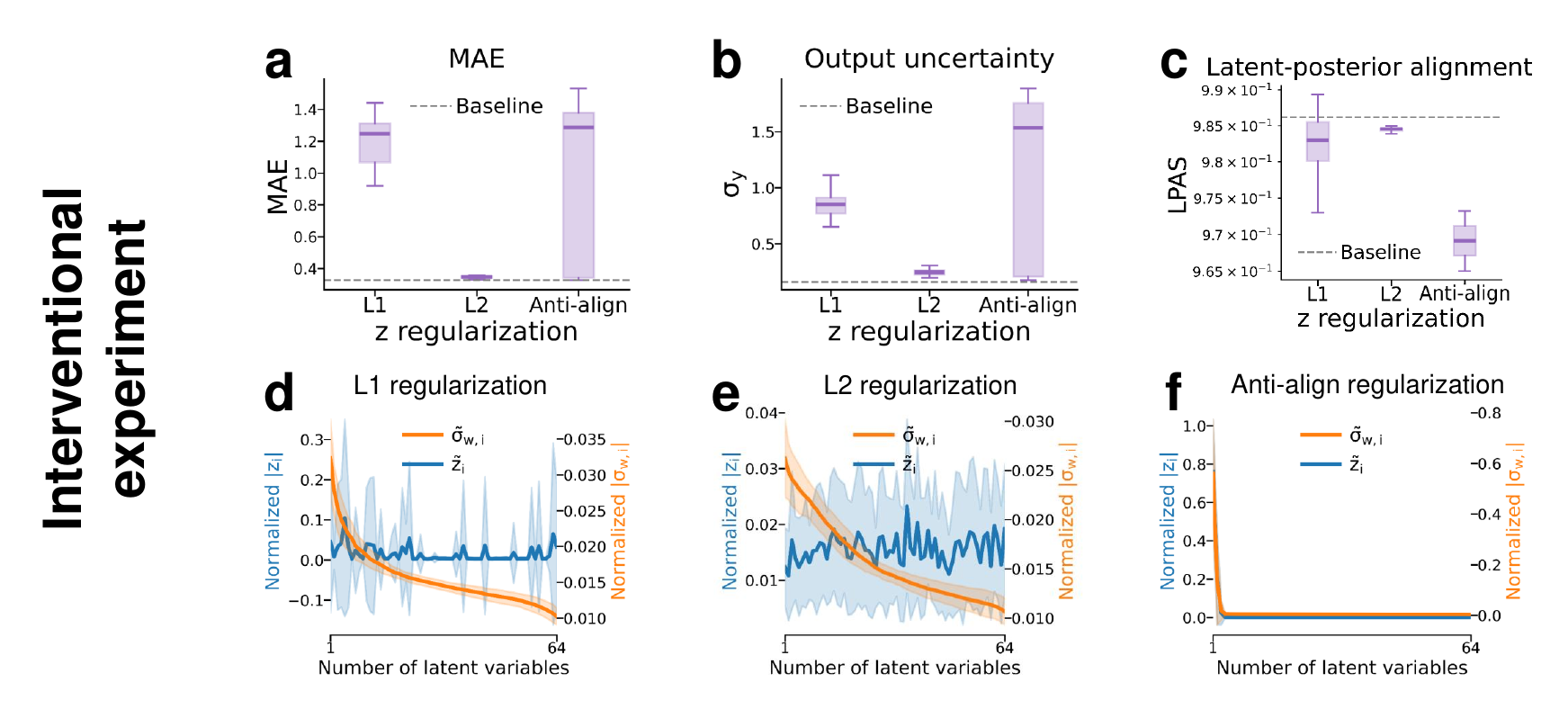}}
    \caption{Interventional evidence for a causal contribution of Latent–Posterior Alignment (a-f). Structurally disrupting the LPA via L1, L2, and anti-alignment regularization degrades accuracy (a) and amplifies predictive uncertainty (b), accompanied by the decrease in the alignment score (LPAS) (c) compared to baseline (dashed line). Visualizations of latent geometry (d–f) illustrate this structural breakdown: unlike the baseline, the regularized models fail to maintain LPA. Shaded areas represent standard deviation across $n=30$ independent runs.}
    \label{fig_2} 
\end{figure}

The results provide convergent interventional evidence that LPA contributes causally to predictive uncertainty (Fig. \ref{fig_2}a–f). Models trained with these latent constraints exhibit a consistent degradation in predictive accuracy (Fig. \ref{fig_2}a) and increase in predictive uncertainty (Fig. \ref{fig_2}b) compared to the unconstrained baseline, despite identical training conditions. Correspondingly, the alignment score (LPAS) is significantly reduced (Fig. \ref{fig_2}c), and the latent representations no longer align with low-variance posteriors (Fig. \ref{fig_2}d-f). Further details of the analysis are provided in Supplementary Note \ref{si_z_regularization_formulation}.

These interventions consistently weaken LPA and increase predictive uncertainty, providing evidence that the alignment has a causal contribution beyond a simple correlation. The anti-alignment penalty provides the most direct perturbation of the proposed mechanism, whereas L1 and L2 regularization and the additional prior and KL-weight experiments provide complementary evidence across distinct perturbations. Their effects differ because these interventions also modify other properties of the learned system: L2 regularization induces relatively diffuse shrinkage across dimensions, whereas L1 regularization promotes sparsity and more strongly restricts the effective dimensional capacity of the representation. Because latent norm, sparsity, representational capacity, and optimization dynamics are not held fixed, these experiments do not identify an isolated causal effect of LPA. We therefore interpret them as convergent evidence for a causal contribution of LPA rather than definitive proof that LPA is the exclusive driver of predictive uncertainty. Consistent results across additional benchmark datasets are provided in the Supplementary Note \ref{si_z_regularization}. Importantly, this behavior is robust to variations in posterior regularization and prior specification. Even when these constraints are relaxed or tightened, the model continues to rely on the alignment structure to regulate predictive uncertainty (see Supplementary Note \ref{si_robustness}).

% \clearpage
\subsection{Alignment-guided learning for density-aware uncertainty} \label{sec_agl}
Motivated by evidence that Latent–Posterior Alignment (LPA) contributes to predictive uncertainty, we introduce a training strategy that explicitly promotes this coupling and examine whether it improves the structural relationship between uncertainty and representation support. Specifically, we introduce Alignment-Guided Learning (AGL), which encourages LPA during training. AGL incorporates the LPAS into the training objective as a regularization term:

\begin{align}
\mathcal{L}_{total} = -\text{ELBO} - \gamma \cdot \text{LPAS} \label{LPAS}
\end{align}

By penalizing misalignment, AGL promotes a latent alignment that favors low-variance posterior directions, thereby reducing predictive uncertainty. The regularization weight $\gamma$ is tuned to maintain predictive performance (MAE) comparable to the baseline model (see Supplementary Table \ref{si_table_lambda}).

To evaluate the effectiveness of AGL, we analyze the model across multiple metrics capturing predictive performance, uncertainty, and structural calibration. In addition to LPAS, MAE, posterior standard deviation ($\sigma_{w}$), and output uncertainty ($\sigma_{y}$), we consider Expected Calibration Error (ECE) \cite{kuleshov2018accurate, laves2020well} which evaluates consistency between model confidence and accuracy, and Density Uncertainty Criterion (DUC) \cite{park2024density}, which measures how well uncertainty reflects the underlying data density. ECE and DUC are essential for assessing uncertainty estimates, as they capture whether uncertainty is accurate, and is structurally consistent with the data distributions, respectively. All values are normalized relative to the baseline (set to 100). For LPAS, the normalized value represents the proportional reduction in $1-\mathrm{LPAS}$. We further compare AGL across data-poor (1–4\%) and data-rich (20–80\%) regimes to examine how alignment-driven uncertainty behaves under varying data density.

\begin{figure}[]
    \centerline{\includegraphics[width=\columnwidth]{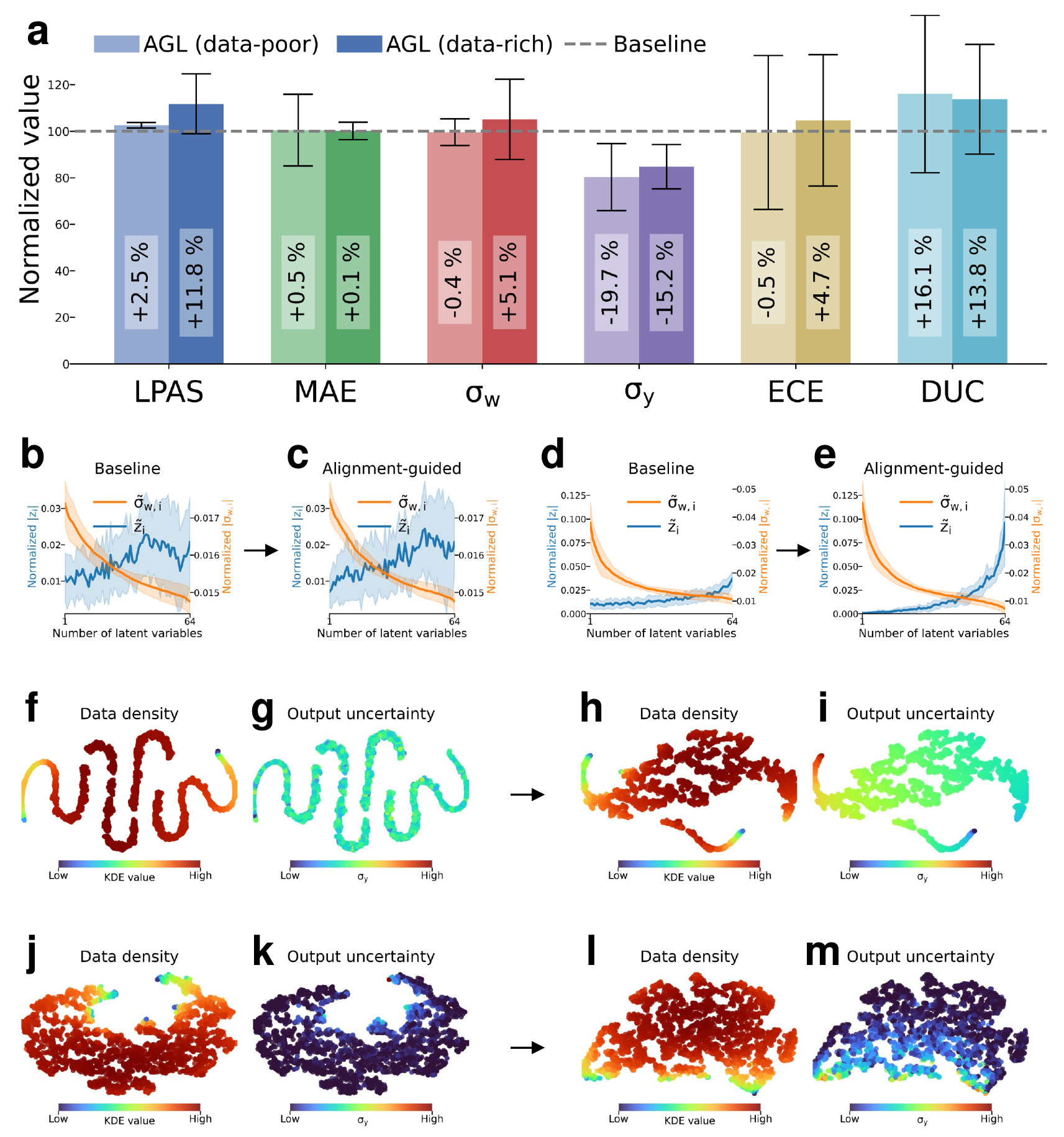}}
    \caption{Alignment-Guided Learning (AGL) and uncertainty calibration analysis. (a) Normalized performance metrics. For LPAS, normalized values represent the proportional reduction in $1-\mathrm{LPAS}$ relative to the baseline; values above 100 indicate stronger alignment. AGL improves structural calibration (DUC) and increases the Latent-Posterior Alignment Score (LPAS) across both regimes without compromising accuracy (MAE) compared to baseline (dashed line). (b–e) Latent alignment. AGL enforces latent vectors ($\tilde{\mathbf{z}}$) to shift toward low-variance posterior dimensions. This geometric reconfiguration is more pronounced in the data-rich regime (d, e) compared to the data-poor regime (b, c). (f-m) Qualitative visualization of density awareness. Each pair displays estimated density (left, KDE) and predictive uncertainty (right, $\sigma_{y}$) on the same t-SNE coordinates. In the data-poor regime (f–i), the baseline shows no correlation between density (f) and uncertainty (g). AGL establishes this missing relationship, aligning low-density regions (h, blue in KDE) with high uncertainty (i, red in $\sigma_{y}$). In the data-rich regime (j-m), AGL refines this structural correspondence, sharpening the boundary so that uncertainty (m) is precisely concentrated along the low-density edges of the data manifold (l), compared to the baseline (j, k).}
    \label{fig_3} 
\end{figure}

As shown in Fig. \ref{fig_3}a, AGL consistently increases LPAS across both regimes, confirming that the alignment is effectively reinforced. This effect is more pronounced in the data-rich setting, where sufficient data allows a more distinct reorganization of the latent space (Fig. \ref{fig_3}d,e). In contrast, latent space reorganization under low data density settings is limited due to information constraints (Fig. \ref{fig_3}b,c).

Meanwhile, the predictive accuracy (MAE) remains unchanged, indicating that AGL preserves the model performance (Fig. \ref{fig_3}a). Importantly, while the posterior uncertainty ($\sigma_{w}$) barely changed, the predictive uncertainty ($\sigma_{y}$) is significantly reduced. This decoupling further supports our central claim: LPA plays a key role in determining predictive uncertainty, rather than the magnitude of posterior variance. By explicitly shaping the latent geometry, AGL reduces output uncertainty without requiring posterior contraction.

Calibration analysis highlights a key advantage of AGL. As shown in Fig. \ref{fig_3}a, the ECE remains largely unchanged, indicating that the alignment between prediction confidence and prediction error is preserved. In contrast, the DUC is substantially improved for both data regimes. This illustrates that AGL is capable of calibrating the structural relationship between uncertainty and data density, while preserving the prediction error. Specifically, uncertainty is evaluated to be high in data scarce regions and low for regions where data is abundant, in contrast to conventional models.

The paired t-SNE maps \cite{maaten2008visualizing} in Fig. \ref{fig_3}f-m provide complementary qualitative evidence for the expected direction of this dependence. The density (estimated via Kernel Density Estimation (KDE); see Methods) and uncertainty ($\sigma_{y}$) maps use the same two-dimensional coordinates, allowing their spatial patterns to be compared directly within each model. In the data-poor regime, the baseline shows weak spatial correspondence (Fig. \ref{fig_3}f, g), whereas AGL more frequently places elevated uncertainty in regions with lower estimated density (Fig. \ref{fig_3}h, i). In the data-rich regime, a partial correspondence is already visible in the baseline (Fig. \ref{fig_3}j, k) and becomes more pronounced under AGL (Fig. \ref{fig_3}l, m), particularly near the representation boundary. Taken together, the DUC score and paired maps indicate that AGL strengthens density-uncertainty dependence and redistributes uncertainty toward less-supported regions of the learned representation. Similar qualitative patterns are observed across the additional datasets (Supplementary Note \ref{si_agl}), illustrating that AGL strengthens LPA and generally increases the association between predictive uncertainty and low-support regions of the learned representation.

% Extended evaluations across five additional datasets show that AGL strengthens LPA and generally increases the association between predictive uncertainty and low-support regions of the learned representation (Supplementary Note \ref{si_agl}).

% \newpage

\section{Discussion}\label{sec_discussion}
The central observation of this study is that predictive uncertainty can decrease even while the Bayesian output-layer posterior becomes broader. As shown in Fig. \ref{fig_1}c, d, and f, output uncertainty declines as the average posterior standard deviation of the output-layer weights increases. Within the mean-field formulation of Eq. \ref{eq_pred_uncert}, latent representations can compensate for a broader posterior by shifting predictive information toward dimensions with smaller posterior variance. The increasing LPAS in Fig. \ref{fig_1_2}f and the perturbation results in Fig. \ref{fig_2} therefore suggest that predictive uncertainty should be interpreted in terms of the jointly trained representation-posterior system, rather than the posterior alone. Because the latent regularizers also alter latent norm, sparsity, and representational capacity, these experiments support a functional contribution of LPA but do not isolate it as the sole causal determinant of predictive uncertainty. Note that this apparent lack of posterior contraction should not be read as a failure of classical posterior-contraction theory. Classical results describe exact posterior concentration in a pre-specified statistical model under regularity conditions, whereas the distribution examined here is a mean-field variational posterior over the output layer coupled to a deterministic representation learned from the same data \cite{van2000asymptotic, freedman1999wald, kleijn2012bernstein, watanabe2009algebraic, dar2021farewell, bochkina2019bernstein, masegosa2020learning}.

The results further show that AGL strengthens density awareness even though data density is not explicitly included in its objective. The increased DUC scores indicate a stronger relationship between representation scarcity and predictive uncertainty, while the paired t-SNE maps show that elevated uncertainty more frequently coincides with lower-density regions after AGL (Fig. \ref{fig_3}). Together, these results support both the strengthened relationship and its expected low-density/high-uncertainty direction. This behavior can be understood from the AGL objective: by promoting latent utilization of lower-posterior-scale directions through Eq. \ref{LPAS}, AGL reduces uncertainty associated with unnecessary activation of higher-posterior-scale dimensions in well-supported regions. Samples in sparse or boundary regions, however, may retain larger latent magnitudes or more atypical combinations of latent components, as suggested by Fig. \ref{fig_1_2}d and e, and therefore remain relatively uncertain even after the global alignment is strengthened. AGL thus does not simply lower uncertainty uniformly, but redistributes it according to support in the learned representation.

% This density-aware allocation of uncertainty is relevant to scientific workflows in which predictions must be accepted or deferred for additional validation. In molecular modeling and MLIPs, for example, elevated uncertainty in sparsely supported latent regions could flag configurations that warrant further DFT calculations or experiments \cite{vandermause2020fly, wen2020uncertainty, kulichenko2023uncertainty}. This role complements error-based calibration: ECE evaluates agreement between confidence and observed errors on labeled data, whereas density awareness indicates whether a new input is supported by the learned representation. Because low latent density does not necessarily imply large error, density-aware uncertainty should guide rather than replace conventional calibration and prospective validation \cite{ovadia2019can, tan2023single}.

Although not evaluated prospectively here, this density-aware allocation of uncertainty may be useful in scientific workflows in which predictions must be accepted or deferred for additional validation. MLIP applications provide a representative example, where users must often decide whether a prediction for a newly encountered atomic configuration is sufficiently reliable or whether an additional DFT calculation should be performed \cite{vandermause2020fly, wen2020uncertainty, kulichenko2023uncertainty}. ECE is valuable for evaluating calibration on a labeled validation distribution, but it cannot by itself determine whether an individual unlabeled configuration lies outside the empirical support of the model. A density-aware uncertainty model is more directly actionable in this setting because latent sparsity and predictive uncertainty can be evaluated before the DFT result is available. In this sense, DUC can serve as a complementary criterion to ECE in MLIP applications. Because AGL increases DUC while maintaining ECE, the resulting uncertainty becomes more informative about the density of the learned latent representation without compromising aggregate error calibration. In practice, a high uncertainty estimate from an AGL-trained model can be interpreted as a signal that the queried atomic configuration may be weakly supported by the training distribution in latent space, providing a rationale for requesting an additional DFT calculation. The same property may also be useful in active learning and Bayesian optimization, where uncertainty is used to prioritize experiments or calculations in insufficiently explored regions. This density-aware interpretation should complement rather than replace error-based calibration, because low latent density does not necessarily imply large error and high latent density does not exclude systematic bias \cite{ovadia2019can, tan2023single}.

Several questions remain unresolved. The present analysis is restricted to deterministic GNN feature extractors combined with mean-field Bayesian output layers, and it is unknown whether the same alignment behavior will emerge under fully Bayesian architectures, correlated posterior approximations, dropout, Laplace methods, or ensembles. In addition, the molecular-property benchmarks used here evaluate uncertainty on fixed test distributions and therefore do not capture the scaffold shifts, prospective chemical-space exploration, or the trajectory-dependent distribution shifts that can arise in molecular simulations. In this context, future studies should examine the training-time evolution of latent directions and posterior scales, and evaluate whether LPA remains informative under alternative uncertainty frameworks. A further important direction is to test whether AGL-based density awareness can support prospective active-learning decisions by prioritizing costly DFT calculations or experiments while reducing the risk of undetected high-error configurations. Within the model class examined here, these findings identify latent–posterior geometry as a central component of predictive uncertainty and provide a practical route for making model confidence more responsive to data support.

\clearpage

\section{Methods}\label{sec_methods}
\subsection{Bayesian GNN framework and training}

\subsubsection{Bayesian Graph Neural Network (BGNN)}
We constructed a backbone GNN composed of five sequential Graph Isomorphism Networks (GIN) \cite{xu2018powerful}, followed by pooling and a fully connected Bayesian layer. Each molecule is represented as a graph where nodes encode 36 atomic features (e.g., element type, hybridization) and edges represent covalent bonds. A complete list of the detailed model architecture and input features are summarized in Supplementary Note \ref{si_model_architecture}.

\subsubsection{Model training}
The model parameters were optimized by minimizing the negative Evidence Lower Bound (ELBO) using the AdamW optimizer. To ensure statistical robustness, all experiments were repeated across 30 independent runs. Specific training configurations, including learning rates, batch sizes, and stopping criteria, are detailed in Supplementary Note \ref{si_model_training}. Additionally, parity plots assessing the predictive performance for each physicochemical property are presented in Supplementary Fig. \ref{si_fig_method_result}.

\subsection{Datasets}
Training datasets were compiled from various publicly available sources, including OPERA \cite{mansouri2018opera} and QM9* \cite{tang2024qm9star}, and carefully curated to ensure internal consistency. For molecules reported with minor discrepancies, values were averaged, while substantial inconsistencies were removed. We allocated 80\% of the data for training and 20\% for testing. Detailed descriptions of all datasets, including their sources, sizes, and preprocessing protocols, are provided in Supplementary Note \ref{si_datasets}.

\subsection{Quantification of uncertainty dynamics}
To ensure the statistical robustness and reproducibility of our findings, all reported metrics represent the aggregate performance across $R=30$ independent training runs with distinct random seeds.

\subsubsection{Predictive metrics (MAE, $\sigma_{y}$)}
Predictive accuracy and output uncertainty are computed by averaging over all $N$ samples in the test dataset and across all $R$ independent runs. The Mean Absolute Error (MAE) and the average output uncertainty $\sigma_y$ are defined as:

\begin{align}
\text{MAE} &= \frac{1}{R} \sum_{r=1}^{R} \left( \frac{1}{N} \sum_{i=1}^{N} |y_{i}^{(r)} - \hat{y}_{i}^{(r)}| \right) \\
\sigma_y &= \frac{1}{R} \sum_{r=1}^{R} \left( \frac{1}{N} \sum_{i=1}^{N} \sigma_{y, i}^{(r)} \right)
\end{align}

where $y_{i}^{(r)}$, $\hat{y}_{i}^{(r)}$, and $\sigma_{y, i}^{(r)}$ denote the ground truth, predicted mean, and predicted standard deviation (or output uncertainty) for the $i$-th sample in the $r$-th run, respectively.

\subsubsection{Posterior metrics ($\mu_{w}$, $\sigma_{w}$)}
To analyze the global behavior of the posteriors, statistics of the posteriors are aggregated across all $d=64$ hidden dimensions and the $R$ runs. The global posterior mean $\mu_w$ and posterior uncertainty $\sigma_w$ are calculated as:

\begin{align}
\mu_{w} &= \frac{1}{R \cdot d} \sum_{r=1}^{R} \sum_{j=1}^{d} \mu_{w_{j}}^{(r)} \\
\sigma_{w} &= \frac{1}{R \cdot d} \sum_{r=1}^{R} \sum_{j=1}^{d} \sigma_{w_{j}}^{(r)} 
\end{align}

where $\mu_{w_{j}}^{(r)}$ and $\sigma_{w_{j}}^{(r)}$ represent the mean and standard deviation of the posterior distribution for the $j$-th weight parameter in the $r$-th run.

\subsubsection{Latent alignment metrics ($\tilde{z}_{i}$, $\tilde{\sigma}_{w, i}$)}
To directly examine the alignment of latent vectors, we quantify the relative magnitude of each latent component and each posterior scale within a normalized space. For each data sample $n$, the absolute latent activation $|z_{i}^{(n)}|$ is normalized across all $d$ hidden dimensions, while the posterior standard deviation $\sigma_{w, i}$ is normalized once per model:

\begin{align}
\tilde{z}_{i} &= \frac{1}{N} \sum_{n=1}^{N} \left( \frac{| z_{i}^{(n)} |}{\sum_{j=1}^{d} | z_{j}^{(n)} |} \right) \\
\tilde{\sigma}_{w,i} &= \frac{\sigma_{w,i}}{\sum_{j=1}^{d} \sigma_{w,j}}
\end{align}

These normalized quantities capture the relative contribution of each latent direction and its associated posterior uncertainty. To visualize the geometric relationship between latent representations and posterior scales, $\tilde{z}_{i}$ values are plotted after sorting the corresponding $\tilde{\sigma}_{w,i}$ in descending order. This enables direct inspection of whether latent directions align with high-variance axes, low-variance axes, or exhibit no systematic structure.

\subsubsection{Relative uncertainty contribution ($C_{i}$)}
To dissect the sources of predictive uncertainty (Eq. \ref{eq_pred_uncert}), we evaluate the proportion of total variance attributable to the $i$-th latent dimension. This is computed for each dimension and averaged across all $N$ samples:

\begin{align}
% C_{i} =  \frac{1}{N} \sum_{l=1}^{N} \left(  \frac{(z_{l,i}^{(r)} \sigma_{w_i}^{(r)})^2}{\sum_{k=1}^{d} (z_{l,k}^{(r)} \sigma_{w_k}^{(r)})^2} \right) \times 100 \: (\%) \\
C_{i} =
\frac{1}{R}\sum_{r=1}^{R}
\frac{1}{N}\sum_{l=1}^{N}
\left(
\frac{(z_{l,i}^{(r)} \sigma_{w_i}^{(r)})^2}
{\sum_{k=1}^{d}(z_{l,k}^{(r)}\sigma_{w_k}^{(r)})^2}
\right)\times100 \: (\%)
\end{align}

This metric quantifies how much each $i$-th latent dimension and its corresponding weight posterior variance contribute to the final output uncertainty. We utilized this to visualize the shift in uncertainty allocation under different data regimes (Fig. \ref{fig_1_2}a--c).

\subsection{Calibration metrics}

\subsubsection{Expected Calibration Error (ECE)}
To assess the reliability of predictive uncertainty, we utilized ECE for regression, which measures the discrepancy between predicted confidence intervals and empirical coverage probabilities. Unlike ECE for classification \cite{naeini2015obtaining, guo2017calibration, nixon2019measuring}, which bins predictions by confidence score, regression ECE evaluates whether the $p$-confidence interval centered at the predicted mean $\mu(x)$ contains the true value $y$ with probability $p$ \cite{kuleshov2018accurate, laves2020well, gustafsson2020evaluating}. Assuming a Gaussian predictive distribution $\mathcal{N}(\mu(x), \sigma^2(x))$, the expected $p$-confidence interval for a target confidence level $p \in (0, 1)$ is defined as:

\begin{align}
I_p(x) = \left[ \mu(x) - \Phi^{-1}\left(\frac{1+p}{2}\right)\sigma(x), \quad \mu(x) + \Phi^{-1}\left(\frac{1+p}{2}\right)\sigma(x) \right]
\end{align}

where $\Phi^{-1}$ denotes the probit function (inverse cumulative distribution function of the standard normal distribution). Using this confidence interval, an empirical coverage $\hat{p}$ can be calculated as the proportion of data samples for which the true value falls within the interval:

\begin{align}
\hat{p} = \frac{1}{N} \sum_{i=1}^{N} \mathbb{I}\left(y_i \in I_p(x_i)\right)
\end{align}

where $N$ is the total number of samples and $\mathbb{I}(\cdot)$ is the indicator function. We considered a range of target confidence levels, $p \in \{0.1, 0.2, \dots, 0.9, 0.99\}$, and calculated the calibration gap between the target confidence $p_{j}$ and the empirical coverage $\hat{p}_{j}$ for each $j$-th confidence interval. The final ECE is computed by approximating the integral of the calibration gap $e_{j}=\vert p_{j} - \hat{p}_{j} \vert$ over the domain of $p$ using the trapezoidal rule:
\begin{align}
\text{ECE} \approx \sum_{j} \frac{|e_{j}| + |e_{j+1}|}{2} \cdot (p_{j+1} - p_{j}), \:\: \text {where  } e_j = \hat{p}_j - p_j
\end{align}
This metric quantifies the total misalignment area between theoretical confidence and observed coverage, with lower values indicating better calibration.

\subsubsection{Density Uncertainty Criterion (DUC)}
To evaluate whether the model’s predictive uncertainty faithfully reflects the underlying data distribution—specifically, assigning high uncertainty to regions of low data support—we utilized the Density Uncertainty Criterion (DUC) \cite{liu2020simple, van2020uncertainty, ovadia2019can}. This metric quantifies the statistical dependence between data scarcity and predictive uncertainty. First, we estimated the probability density function $p(\mathbf{z})$ of the latent representations using Kernel Density Estimation (KDE) with a Gaussian kernel and a bandwidth of 0.2. To measure the scarcity of a data point, we then computed the negative log-likelihood (also known as surprisal):
\begin{align}
s(\mathbf{z}) = -\log p(\mathbf{z})
\end{align}

% where a higher $s(\mathbf{z})$ indicates a lower density region. Finally, the DUC is defined as the mutual information (MI) between this sparsity score $s(\mathbf{z})$ and the model's predictive uncertainty $\sigma(\mathbf{y}|\mathbf{z})$:

A higher DUC indicates stronger statistical dependence between data scarcity and predictive uncertainty. Because MI is unsigned, DUC alone does not determine the direction of this dependence; the expected low-density/high-uncertainty direction is assessed separately using the paired KDE–uncertainty maps in Fig. \ref{fig_3} and Supplementary Figs. \ref{si_fig_agl_data_poor} and \ref{si_fig_agl_data_rich}.

\begin{align}
\text{DUC} = MI(s(\mathbf{z}); \sigma(\mathbf{y}|\mathbf{z}))
\end{align}

where $MI(X; Y)$ denotes the mutual information between random variables $X$ and $Y$. We estimated this value with a continuous MI estimator based on nearest neighbor distances using scikit-learn v1.6.1 \cite{pedregosa2011scikit}. A higher DUC indicates a stronger alignment between the model's uncertainty estimates and the geometry of the data manifold, confirming that the model effectively assigns higher uncertainty to regions where data is scarce.

\section*{Acknowledgements}

This work was supported by the Nano-Material Technology Development Program (RS-2026-25534767), STEAM Project (2022M3C1A3092056), Outstanding Junior Researcher Program (RS-2024-00348230), and the institutional research program of Korea Institute of Science and Technology (26E0323) of the National Research Foundation (NRF) grant funded by the Korea Ministry of Science and ICT. This work was also supported by the National Supercomputing Center with supercomputing resources including technical support (KSC-2023-CRE-0559).

% \backmatter

% \section*{Declarations}

\bibliography{sn-bibliography}

% \clearpage
% \section{SI}
% \input{SI}

\clearpage
\section*{Supplementary Information}

% Restart section numbering for the Supplementary Information
\setcounter{section}{0}
\setcounter{subsection}{0}
\setcounter{figure}{0}   
\setcounter{table}{0}
\setcounter{equation}{0}

\renewcommand*{\theHsection}{SI.section.\arabic{section}}
\renewcommand*{\theHsubsection}
    {SI.subsection.\arabic{section}.\arabic{subsection}}
\renewcommand*{\theHsubsubsection}
    {SI.subsubsection.\arabic{section}.\arabic{subsection}.\arabic{subsubsection}}
\renewcommand*{\theHfigure}{SI.figure.\arabic{figure}}
\renewcommand*{\theHtable}{SI.table.\arabic{table}}
\renewcommand*{\theHequation}{SI.equation.\arabic{equation}}

\def\panelwidth{0.24\textwidth}
\def\onewidth{0.6\textwidth}
\def\fourwidth{0.24\textwidth}

\section{Experimental details} \label{si_details}
\subsection{Model architecture} \label{si_model_architecture}

\begin{figure}[H]
    \centering
    \begin{subfigure}{\textwidth}
        \centering
        \includegraphics[width=\textwidth]{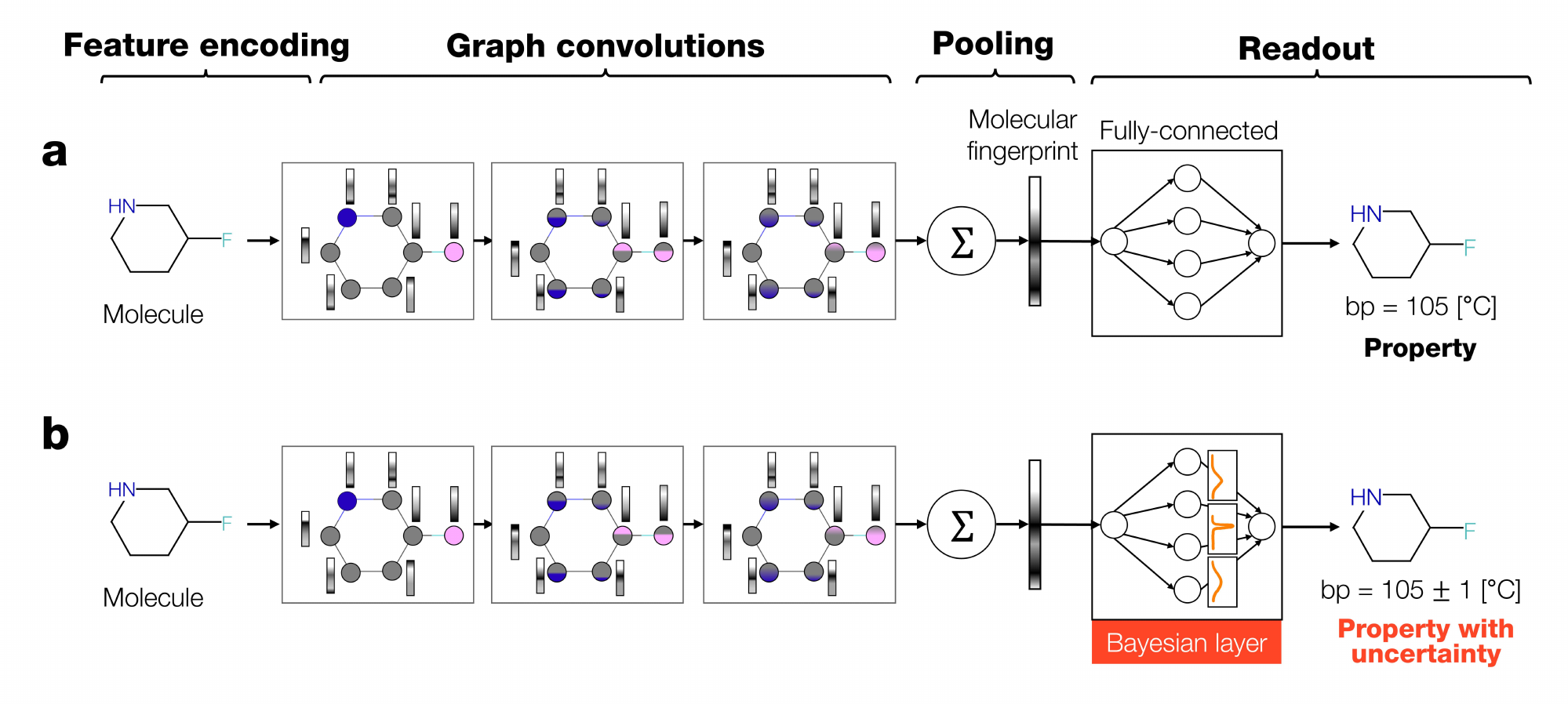}
    \end{subfigure}   
    \caption{Schematic of Bayesian GNN models developed in this study. (a) Backbone GNN model and (b) BGNN model.}
    \label{fig_method_bgnn}
\end{figure}

To develop the Bayesian GNN, we first construct a backbone GNN composed of five sequential graph-convolutional layers, followed by a pooling operation and a fully connected layer (Fig. \ref{fig_method_bgnn}). Each molecule is represented as a graph, where the nodes encode 37 atomic features listed in Table \ref{table_method_datafeatures}, and edges represent the covalent bonds between atoms. The node features are propagated and aggregated across local neighborhoods through convolutional operations using a Graph Isomorphism Network (GIN) \cite{xu2018powerful, morris2019weisfeiler}. The resulting graph embeddings are then pooled to generate a molecular fingerprint (latent representation), which is subsequently fed into a fully connected layer to predict the target physicochemical property. Architectural details, including the number of layers and dimensions of hidden channels, are summarized in Table \ref{table_methods_model_detail}.

\begin{table}[h]
    \centering
    \renewcommand{\arraystretch}{1.3}
    \begin{tabularx}{\textwidth}{|p{2.5cm}|X|p{3.5cm}|}
        \hline
        \textbf{Feature} & \textbf{Note} & \textbf{Implementation} \\
        \hline

        Element type &
        Ten elemental types corresponding to atomic numbers (1, 5, 6, 7, 8, 15, 16, 17, 35, and 53) are represented using a one-hot encoding. For instance, a carbon atom is assigned the vector [0, 0, 1, 0, 0, 0, 0, 0, 0, 0]. & atom\_GetAtomicNum() \\
        \hline

        Formal charge &
        The formal charge of each atom is one-hot encoded over the set \{-2, -1, 0, +1, +2\}; any value outside this range (e.g., –3) is mapped to an all-zero vector. & atom\_GetFormalCharge() \\
        \hline

        Hybridization &
        The hybridization state of each atom is encoded over the set \{s, sp, sp$^2$, sp$^3$\}, with each state mapped to its corresponding one-hot vector. & atom\_GetHybridization() \\
        \hline

        Total degree &
        The number of bonded neighbors is encoded using a one-hot scheme over the values \{1, 2, 3, 4\}. & atom\_GetTotalDegree() \\
        \hline

        Total number of neighboring hydrogen atoms &
        The total number of attached hydrogen atoms is represented by a one-hot vector defined over the set \{0, 1, 2, 3\}. & atom\_GetTotalNumHs() \\
        \hline

        Aromatic &
        Whether an atom is aromatic is encoded as a binary feature taking values from \{0, 1\}. & atom\_GetIsAromatic() \\
        \hline

        Total number of valence electrons &
        The total number of valence electrons is encoded using a one-hot vector spanning the set \{1, 2, 3, 4, 5, 6\}. & atom\_GetTotalValence() \\
        \hline

        Within a ring &
        Whether an atom is part of a ring is encoded as a binary feature taking values from \{0, 1\}. & atom\_IsInRing() \\
        \hline

        Reaction center &
        For pKa prediction, atoms involved in protonation or deprotonation are indicated with a value of 1; otherwise, 0 is assigned. \newline
        For BDE prediction, atoms constituting the terminal points of the bond undergoing cleavage are marked with a value of 1; otherwise, 0 is assigned.
        & atom\_GetIdx() == target\_idx \\
        \hline
    \end{tabularx}
    \caption{Description of utilized atomic features.}
    \label{table_method_datafeatures}
\end{table}

\begin{table}[h]
    \centering
    \renewcommand{\arraystretch}{1.3}
    \begin{tabularx}{\textwidth}{|p{3.2cm}|X|p{2cm}|}
        \hline
        \textbf{Layer} & \textbf{Parameter} & \textbf{Value} \\
        \hline
        \multirow{4}{=}{\textbf{Graph convolutions with GIN}} 
        & Number of layers & 5 \\
        \hhline{|~|-|-|}
        & Number of hidden channels & 128 \\
        \hhline{|~|-|-|}
        & Dimensionality of fingerprint channel \newline (latent dimension) & 64 \\
        \hhline{|~|-|-|}
        & Dropout probability & 0.3 \\
        \hline
        \textbf{Bayesian layer}
        & Number of layers & 1 \\
        \hline
    \end{tabularx}
    \caption{Detailed description of model architecture.}
    \label{table_methods_model_detail}
\end{table}

The Bayesian GNN is implemented by reparameterizing each weight in the final fully connected layer as a Gaussian distribution (Fig. \ref{fig_method_bgnn}b):

\begin{align} 
\mathbf{w} &= \mu_{\mathbf{w}} + \sigma_{\mathbf{w}} \cdot \epsilon\\
\mathbf{b} &= \mu_{\mathbf{b}} + \sigma_{\mathbf{b}} \cdot \epsilon
\end{align}

where $\epsilon\sim\mathcal{N}\left(0,1^2\right)$ denotes random samples drawn from a standard normal distribution. Under this formulation, each forward pass generates a distinct realization of the weights by sampling $\epsilon$. This stochastic construction yields an implicit ensemble effect within a single BGNN and enhances robustness, particularly when available data are sparse or contaminated with noise \cite{blundell2015weight, izmailov2021bayesian, pang2021evaluating}. A separate BGNN model is constructed for each physicochemical property, following standard property-specific modeling practices. All models used in this study were implemented using PyTorch v2.7.0 \cite{paszke2019pytorch}, and molecular processing was performed using RDKit v2024.09.06 \cite{rdkit}.

\subsection{Model training} \label{si_model_training}

\begin{figure}[H]
    \centering
    \begin{subfigure}{\textwidth}
        \centering
        \includegraphics[width=\textwidth]{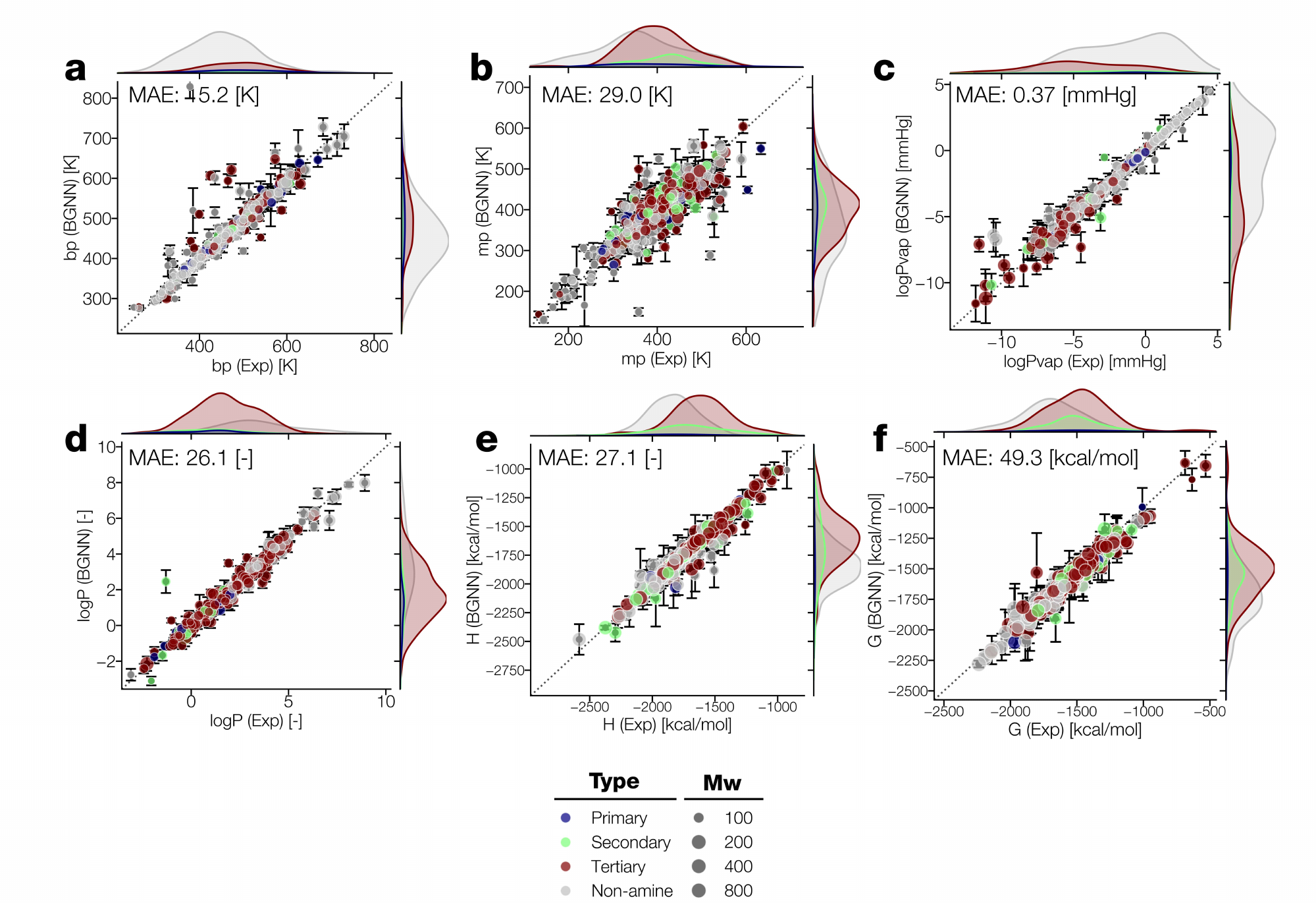}
    \end{subfigure}   
    \caption{Parity plots for the physicochemical properties examined in this study: (a) boiling point (bp), (b) melting point (mp), (c) vapor pressure at 25 $^{\circ}C$ (Pvap), (d) partition coefficient (P), (e) standard enthalpy of formation (H), and (f) standard Gibbs free energy of formation (G). The mean absolute error (MAE) for each property is annotated. Marker color denotes the amine type, and marker size represents molecular weight [g/mol]. Amine types are assigned priority in the order: tertiary $>$ secondary $>$ primary, when a compound belongs to multiple classes.}
    \label{si_fig_method_result}
\end{figure}

To approximate the intractable posterior distributions of the weights and biases, we employed Variational Inference (VI) with a mean-field approximation \cite{blundell2015weight, tran2019bayesian, dusenberry2020efficient}. The model parameters were optimized by minimizing the negative Evidence Lower Bound (ELBO), which consists of a reconstruction error term (negative log-likelihood) and a regularization term (KL-divergence) \cite{kingma2013auto}. The total training objective $\mathcal{L}(\theta)$ is defined as:

\begin{align} 
\mathcal{L}(\theta) = \underbrace{\mathbb{E}_{q_{\theta}(w, b)}\left[-\log{p(\mathcal{D} \vert w, b)}\right]}_{\text{NLL}} + \beta \underbrace{\left[ KL(q_{\theta}(w) \Vert p(w)) + KL(q_{\theta}(b) \Vert p(b)) \right]}_{\text{Regularization}} 
\end{align}
where $\mathcal{D}$ denotes the training dataset, and $\beta$ is a weighting factor balancing the regularization strength. The detailed derivation of the ELBO is provided in Supplementary Note \ref{si_elbo}. To accelerate convergence and enhance training stability, the backbone GNN model is first trained, after which its learned weights are transferred to the BGNN to provide a warm start. To ensure statistical robustness and reproducibility, all experiments were repeated across 30 independent runs for every configuration. Detailed training configurations are summarized in Table \ref{table_model_training}. The parity plots and prediction errors for each physicochemical property are presented in Fig. \ref{si_fig_method_result}.

\begin{table}[h]
    \centering
    \renewcommand{\arraystretch}{1.4} 
    \begin{tabularx}{\textwidth}{|p{4.5cm}|X|}
        \hline
        \textbf{Parameter} & \textbf{Value} \\
        \hline
        
        Input data scaling method &
        Normalization to (0, 1) range \\
        \hline
        
        Output data scaling method &
        Standardization to $\mathcal{N}(0,1)$ \\
        \hline
        
        Number of epochs & 
        500 \\
        \hline
        
        Batch size &
        64 for $0 < N < 10^4$; \newline
        256 for $10^4 \le N < 10^5$; \newline
        1024 for $10^5 \le N$ \\
        \hline
        
        Optimizer &
        AdamW \\
        \hline
        
        Initial learning rate &
        $10^{-3}$ \\
        \hline
        
        Minimum learning rate &
        $10^{-6}$ \\
        \hline
        
        Learning rate scheduling &
        \texttt{ReduceLROnPlateau} (factor = 0.8, patience = 10) \\
        \hline
        
        Gradient clipping &
        (-10, 10) \\
        \hline
        
        Stopping criteria &
        Current loss / initial loss $< 10^{-3}$ \newline
        MAE / span of property values in training dataset $< 10^{-3}$ \newline
        Patience of 60 epochs for improvement in the loss \\
        \hline
        
        Loss function for GNN &
        Mean squared error (MSE) $\times 10^3$ \\
        \hline

        Independent runs &
        30 \\
        \hline
        
    \end{tabularx}
    \caption{Detailed description of model training configuration.}
    \label{table_model_training}
\end{table}

\subsection{Datasets} \label{si_datasets}
Training datasets were compiled from publicly accessible sources and carefully curated to ensure internal consistency. For molecules reported with only minor discrepancies in property values, the corresponding entries were averaged, whereas data points exhibiting substantial inconsistencies were removed. Eighty percent of the collected data points were allocated for model training, with the remaining portion reserved for testing. A detailed description of all datasets used in this study, including their sources, sizes, and preprocessing transformations, is provided in Table \ref{table_methods_datasets}.

\begin{table}[h]
    \centering
    \renewcommand{\arraystretch}{1.4}
    \tiny 
    \begin{tabularx}{\textwidth}{|p{2.5cm}|X|p{3.2cm}|>{\centering\arraybackslash}p{1.6cm}|}
        \hline
        \textbf{Property \newline (abbreviation) [unit]} &
        \textbf{Data sources} &
        \textbf{Note} &
        \textbf{Total number of data points} \\
        \hline

        Boiling point \newline (bp) [K] &
        Sanchez-Lengeling et al.\cite{sanchez2017optimizing} \newline
        \url{https://github.com/aspuru-guzik-group/ORGANIC/blob/master/data/datasets} \newline\newline
        OPERA dataset v2.9 \cite{mansouri2018opera} \newline
        \url{https://github.com/kmansouri/OPERA/tree/master}
        & -- &
        5,656 \\
        \hline

        Melting point \newline (mp) [K] &
        Jean-Claude Bradley Open Melting Point Dataset \newline
        \url{https://figshare.com/articles/dataset/Jean_Claude_Bradley_Open_Melting_Point_Dataset/1031637?file=1503990} \newline\newline
        OPERA dataset v2.9 \cite{mansouri2018opera} \newline
        \url{https://github.com/kmansouri/OPERA/tree/master}
        & -- &
        21,414 \\
        \hline

        Partition coefficient  \newline  (P) [-] &
        OPERA dataset v2.9 \cite{mansouri2018opera} \newline
        \url{https://github.com/kmansouri/OPERA/tree/master}
        & Polarity of a compound: \newline
        $\log P = \log_{10}(\frac{[\text{solute}]_{\text{oct}}}{[\text{solute}]_{\text{water}}})$
        & 13,962 \\
        \hline

        Vapor pressure \newline (Pvap) [mmHg] &
        Sanchez-Lengeling et al.\cite{sanchez2017optimizing} \newline
        \url{https://github.com/aspuru-guzik-group/ORGANIC/blob/master/data/datasets} \newline\newline
        OPERA dataset v2.9 \cite{mansouri2018opera} \newline
        \url{https://github.com/kmansouri/OPERA/tree/master}
        & Vapor pressure at 25$^\circ$C
        & 2,925 \\
        \hline

        Enthalpy \newline (H) [kcal/mol] &
        \multirow{2}{=}{QM9* star dataset \cite{tang2024qm9star} \newline \url{https://github.com/gentlej1999/qm9star_query}}
        & Standard enthalpy of formation &
        \multirow{2}{*}{2,008,806} \\
        \hhline{-~-~} 
        
        Gibbs free energy \newline (G) [kcal/mol] &
        & Standard Gibbs free energy of formation &
        \\
        \hline
    \end{tabularx}
    \caption{Description of datasets and their properties.}
    \label{table_methods_datasets}
\end{table}

\clearpage

\section{Theoretical context for posterior contraction in the present setting} \label{si_analysis}

In this section, we summarize the classical Bayesian asymptotic setting and discuss its applicability to the model studied here. We first outline the assumptions of the Bernstein-von Mises (BvM) theorem and then describe how the present setting, which uses a mean-field variational approximation over a Bayesian output layer coupled to a deterministic representation learned from the same data, differs from the pre-specified exact-posterior setting considered by the classical theorem.

\subsection{The Bernstein-von Mises Theorem}
Consider a parametric model family $\{p(x|\boldsymbol{\theta}): \boldsymbol{\theta} \in \Theta \subset \mathbb{R}^d\}$ and a dataset $\mathcal{D}_n = \{x_1, \dots, x_n\}$ consisting of independent and identically distributed (i.i.d.) samples from a true distribution $p_0(x)$, where $d$ and $n$ denote the number of model parameters and data points, respectively. Under standard regularity conditions (detailed in Section \ref{si_bvm_assumptions}), the Bernstein-von Mises theorem states that as $n \to \infty$, the posterior distribution $p(\boldsymbol{\theta}|\mathcal{D}_n)$ converges in total variation (TV) distance to a multivariate Gaussian distribution centered at the maximum likelihood estimator $\hat{\boldsymbol{\theta}}_n$ \cite{van2000asymptotic, freedman1999wald}:

\begin{align}
\| p(\boldsymbol{\theta}|\mathcal{D}_n) - \mathcal{N}(\hat{\boldsymbol{\theta}}_n, \Sigma_n) \|_{TV} \to 0
\label{eq_bvm_var}
\end{align}

where the asymptotic covariance $\Sigma_n$ is given by the inverse of the Fisher Information Matrix (FIM), scaled by the sample size:
$$\Sigma_n = \frac{1}{n} I(\boldsymbol{\theta}_0)^{-1}$$
Here, $I(\boldsymbol{\theta}_0)$ is the Fisher Information Matrix at the true parameter $\boldsymbol{\theta}_0$, defined as:

\begin{align}
I(\boldsymbol{\theta}_0) = \mathbb{E}_{x \sim p_0} \left[ -\nabla_{\boldsymbol{\theta}}^2 \log p(x|\boldsymbol{\theta}) \big|_{\boldsymbol{\theta}=\boldsymbol{\theta}_0} \right]
\label{eq_bvm_fim}
\end{align}

In this classical regime, the posterior variance scales as $O(1/n)$, and the posterior mass increasingly concentrates around $\boldsymbol{\theta}_0$ as $n$ increases. This result provides the theoretical basis for the conventional expectation that more data yield a tighter posterior under the regularity conditions of the theorem \cite{van2000asymptotic, freedman1999wald, blundell2015weight, wenzel2020good}.

\subsection{Applicability of classical assumptions to the present model} \label{si_bvm_assumptions}

The convergence guarantee of the BvM theorem relies on three structural assumptions \cite{van2000asymptotic, freedman1999wald}: (1) the model is well-specified, (2) the parameter space is finite-dimensional and fixed, and (3) the model is regular, with a positive-definite Fisher Information Matrix. These assumptions are not automatically inherited by the present training formulation, and the following differences limit a direct application of the classical result.

1. \textbf{Possible misspecification (difference from the ``well-specified'' setting):} Classical theory assumes that the model family $\{p(x|\boldsymbol{\theta}) : \boldsymbol{\theta} \in \Theta\}$ contains a parameter $\boldsymbol{\theta}_0$ that represents the data-generating distribution $p_0(x)$. Neural-network models used for real-world molecular data may not satisfy this assumption exactly \cite{bronstein2021geometric, bengio2013representation}. Under misspecification, however, a posterior may still concentrate around a pseudo-true parameter rather than remain diffuse \cite{kleijn2012bernstein, masegosa2020learning}. Misspecification alone therefore does not imply persistent posterior broadening, but it changes the target and conditions of posterior concentration.

2. \textbf{Over-parameterization and learned representations (difference from the ``finite-dimensional and fixed'' setting):} The deterministic GNN feature extractor can be highly parameterized, whereas the variational distribution analyzed in this study is placed only over the fixed-dimensional Bayesian output layer. Full-network over-parameterization therefore does not by itself establish a failure of posterior contraction in the output layer. A more relevant distinction is that the effective design supplied to the Bayesian readout, namely the learned representation $\mathbf{z}$, changes with the training data and optimization. The output-layer posterior is consequently coupled to a data-dependent representation rather than to a pre-specified fixed design.

3. \textbf{Regularity and approximate inference (difference from the classical exact-posterior setting):} Deep neural networks can be singular when their complete parameterization is considered, owing to non-identifiability and parameter symmetries \cite{watanabe2009algebraic, wei2022deep}. In the present model, however, the analyzed distribution is a mean-field variational approximation over the Bayesian output layer rather than an exact posterior over the full network. Conditional on a learned representation, the linear output layer may be regular when its effective design is full rank. The singularity of the complete deterministic network therefore does not by itself prove that the output-layer posterior must remain diffuse. Instead, the classical BvM result does not directly require monotonic contraction of a $\beta$-weighted mean-field variational posterior coupled to a jointly learned representation. The posterior broadening observed here should accordingly be interpreted as an empirical property of the tested architecture and training objective, rather than as a violation of classical posterior-contraction theory.

\clearpage

\section{Derivation of the Evidence Lower Bound (ELBO) objective} \label{si_elbo}

In the Bayesian neural network model employed in this study, both weights $w$ and biases $b$ are treated as random variables with prior distributions $p(w)$ and $p(b)$. Given a dataset $\mathcal{D} = \left\{ x_{i}, y_{i}\right\}_{i=1}^{N}$, the joint distribution is:

\begin {align}
p_{\theta}(\mathcal{D}, w, b) = p(w)p(b)\prod_{i=1}^{N}p_{\theta}(y_{i}\vert x_{i}, w, b)
\end {align}

Since the true posterior $p(w, b\vert \mathcal{D})$ is intractable, we introduce a variational approximation $q_{\Phi}(w, b) = q_{\Phi}(w)q_{\Phi}(b)$. Then, the evidence lower bound (ELBO) is formulated as \cite{blei2017variational, jordan1999introduction}:

\begin {align}
\log{p_{\theta}(\mathcal{D})} & = \log{\int p_{\theta}(\mathcal{D}, w, b)dwdb} \\
& \geq \mathbb{E}_{q_{\Phi}} \left[ \log{p_{\theta}(\mathcal{D}\vert w, b)} \right] - KL(q_{\Phi}(w, b) \Vert p(w, b))
\end {align}

Expanding the terms gives:
\begin {align}
ELBO = \sum_{i=1}^{N} \mathbb{E}_{q_{\Phi}(w, b)} \left[ \log{p_{\theta}(y_{i} \vert x_{i}, w, b)} \right] - KL(q_{\Phi}(w)\Vert p(w)) - KL(q_{\Phi}(b)\Vert p(b))
\end {align}

In practice, a weighting factor $\beta$ is applied to the KL term to balance regularization strength when minimizing the negative ELBO:

\begin {align}
\mathcal{L}(\theta) = \mathbb{E}_{q_{\theta}(w, b)}\left[-\log{p(\mathcal{D} \vert w, b)}\right] + \beta\left[ KL(q_{\theta}(w) \Vert p(w)) + KL(q_{\theta}(b) \Vert p(b)) \right] \label{eq_2}
\end {align}

\newpage

\section{Negligible impact of bias posterior distributions} \label{si_bias}
\begin{figure}[]
    \centering
    \begin{subfigure}{\textwidth}
        \centering
        \includegraphics[width=\panelwidth]{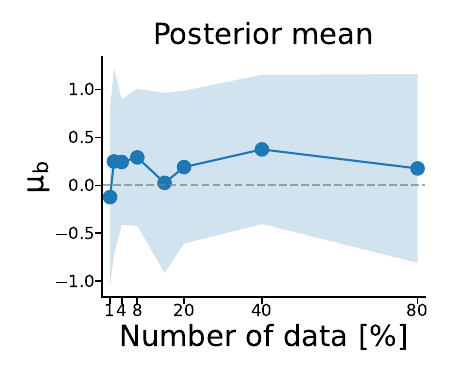}
        \includegraphics[width=\panelwidth]{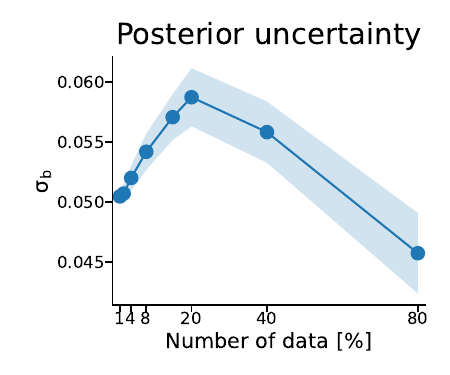}
        \includegraphics[width=\panelwidth]{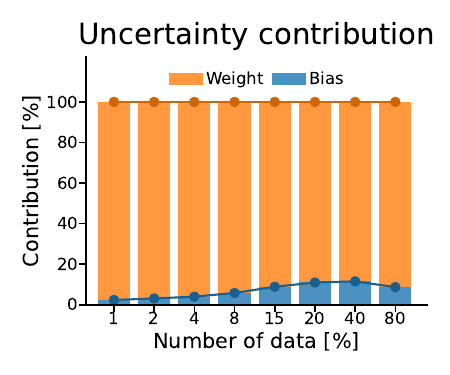}
        \caption{Boiling point (bp)} % (b) 
        \label{si_1_a}
    \end{subfigure}
    \begin{subfigure}{\textwidth}
        \centering
        \includegraphics[width=\panelwidth]{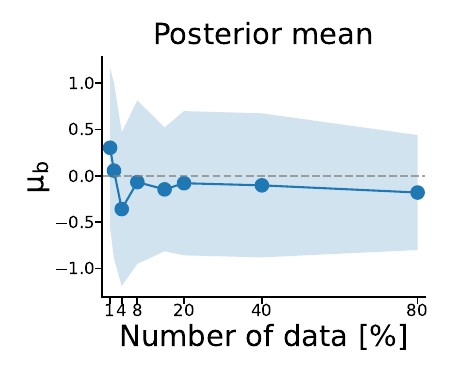}
        \includegraphics[width=\panelwidth]{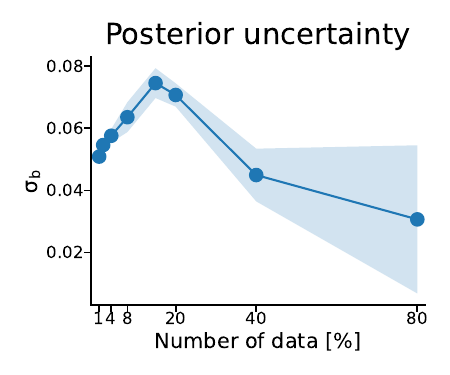}
        \includegraphics[width=\panelwidth]{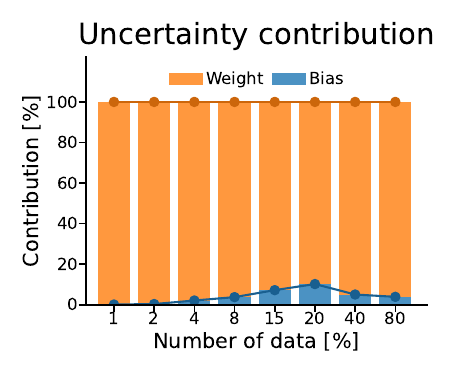}
        \caption{Melting point (mp)} % (b) 
        \label{si_1_b}
    \end{subfigure}
    \begin{subfigure}{\textwidth}
        \centering
        \includegraphics[width=\panelwidth]{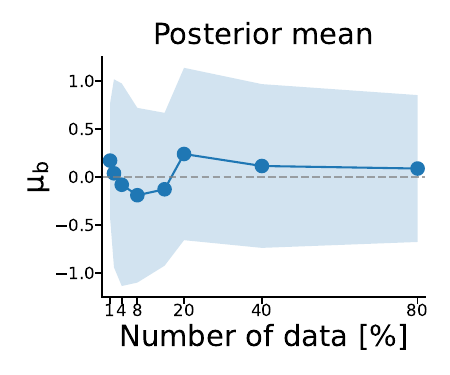}
        \includegraphics[width=\panelwidth]{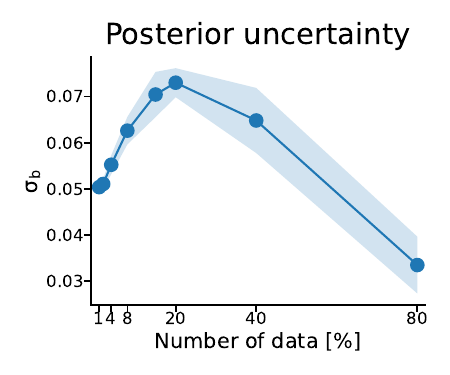}
        \includegraphics[width=\panelwidth]{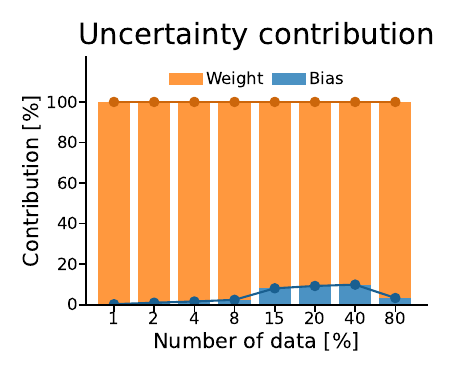}
        \caption{Partition coefficient (P)} % (b) 
        \label{si_1_c}
    \end{subfigure}
    \begin{subfigure}{\textwidth}
        \centering
        \includegraphics[width=\panelwidth]{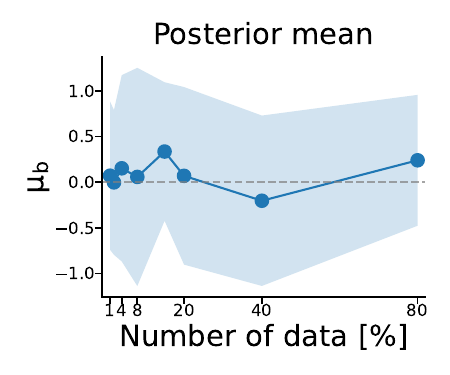}
        \includegraphics[width=\panelwidth]{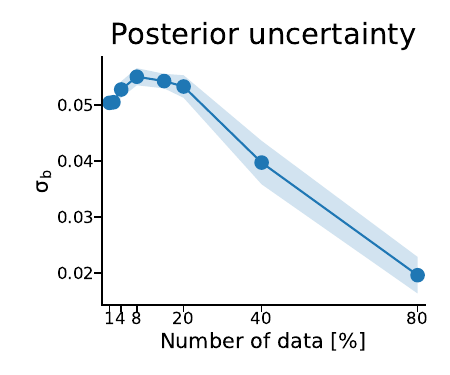}
        \includegraphics[width=\panelwidth]{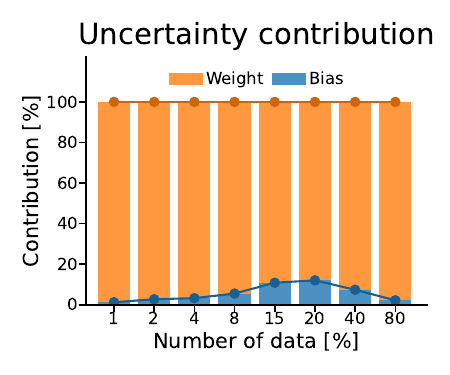}
        \caption{Vapor pressure (Pvap)} % (b) 
        \label{si_1_d}
    \end{subfigure}
    \begin{subfigure}{\textwidth}
        \centering
        \includegraphics[width=\panelwidth]{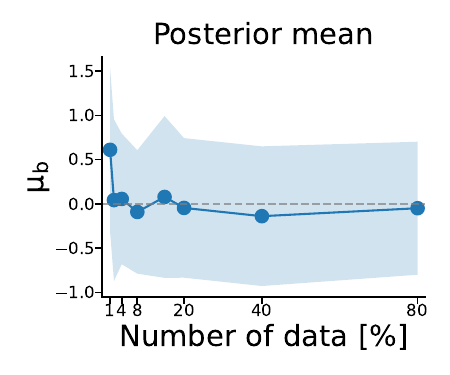}
        \includegraphics[width=\panelwidth]{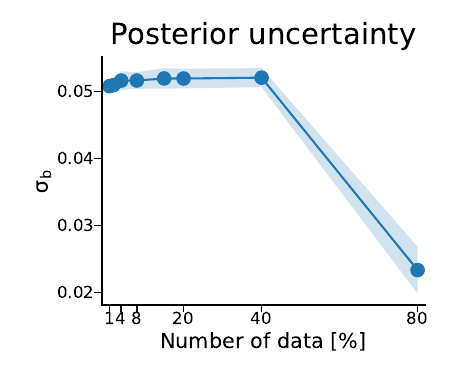}
        \includegraphics[width=\panelwidth]{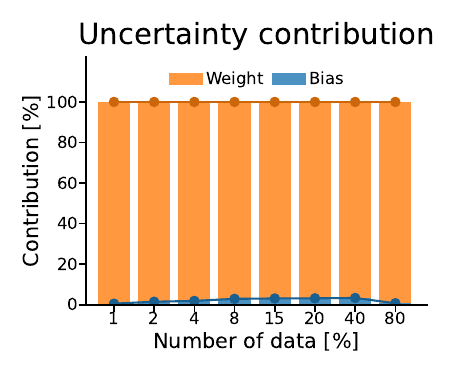}
        \caption{Enthalpy (H)} % (b) 
        \label{si_1_e}
    \end{subfigure}
    \begin{subfigure}{\textwidth}
        \centering
        \includegraphics[width=\panelwidth]{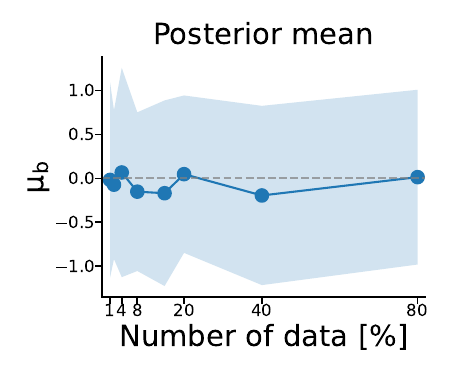}
        \includegraphics[width=\panelwidth]{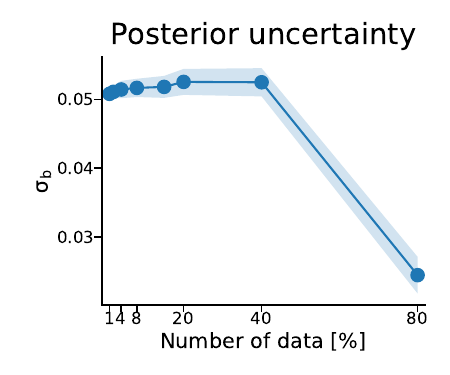}
        \includegraphics[width=\panelwidth]{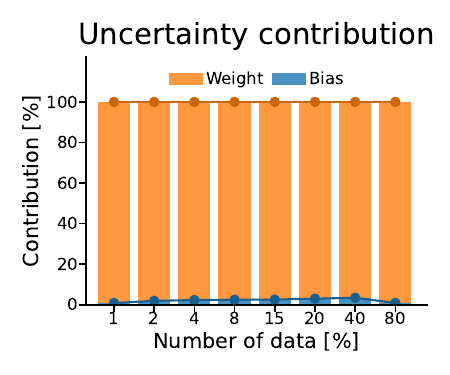}
        \caption{Gibbs free energy (G)} % (b) 
        \label{si_1_f}
    \end{subfigure}
    
    \caption{\textbf{Negligible influence of bias posteriors on predictive uncertainty.} Analyses are presented for six different datasets (a-f), where each panel displays the evolution of bias posterior mean, posterior standard deviation, and relative uncertainty contribution (blue for bias, orange for weight) as a function of training data size.}
    \label{si_fig_bias}
\end{figure}

To ensure that our analysis of uncertainty dynamics is not confounded by the bias posterior, we examined the behavior of the bias distribution across multiple datasets. We observed that the posterior mean ($\mu_{b}$, Fig. \ref{si_fig_bias}) converges closely to zero, while the standard deviation ($\sigma_b$) exhibits negligible magnitude and inconsistent fluctuations with respect to dataset size. Unlike the weight posteriors, bias uncertainty shows no clear pattern correlating with predictive uncertainty, suggesting it does not play a governing role in data–uncertainty dynamics. We further quantified this observation by analytically decomposing the total predictive uncertainty. The predictive variance over a dataset $\mathcal{D}$ can be expressed as the sum of contributions from the weights and the bias:

\begin{align}
\mathrm{Var}(y \mid \mathbf{z}, \mathcal{D}) 
  & =  \mathbf{z}^\top \mathrm{Cov}\left[\mathbf{w}\vert \mathcal{D}\right] \mathbf{z} + \sigma_{b}^{2} \\
  & = \underbrace{\sum_{i=1}^{d}z_{i}^{2}\sigma_{i}^{2}}_{weight} + \underbrace{\sigma_{b}^{2}}_{bias} 
\label{var_avg_1}
\end{align}

where $d$ is the latent dimension and $\sigma_{i}^{2}$ is the variance of the $i$-th weight parameter. In this decomposition, the first term represents the contribution of the weight posterior coupled with the latent representation, while the second term ($\sigma_{b}^{2}$) represents the bias contribution. As illustrated in Fig. \ref{si_fig_bias}, the weight contribution (orange bar) consistently dominates the bias contribution (blue bar) by a substantial margin across all datasets and training sizes. Consequently, the bias posterior exerts a practically insignificant influence on output uncertainty, justifying our focus on the weight posterior and latent alignment as the primary drivers of uncertainty.
\newpage

\section{Cross-dataset consistency of posterior–predictive decoupling} \label{si_paradox}
\begin{figure}[H]
    \centering
    \begin{subfigure}{\textwidth}
        \centering
        \includegraphics[width=\panelwidth]{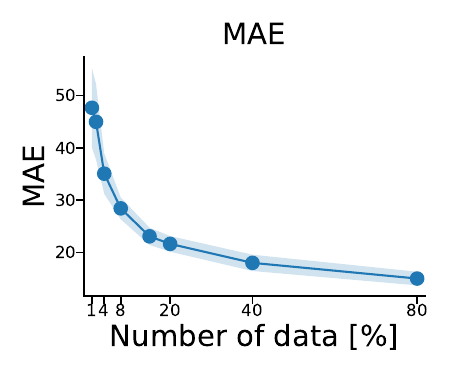}
        \includegraphics[width=\panelwidth]{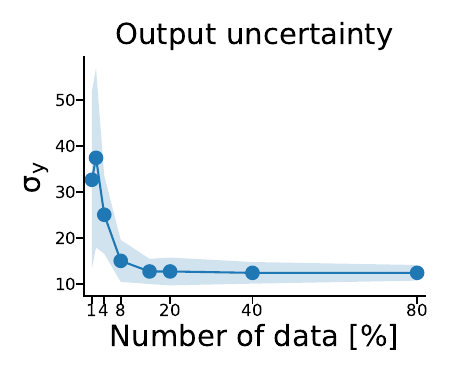}
        \includegraphics[width=\panelwidth]{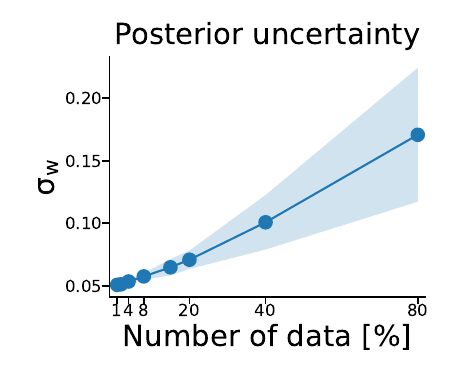}
        \includegraphics[width=\panelwidth]{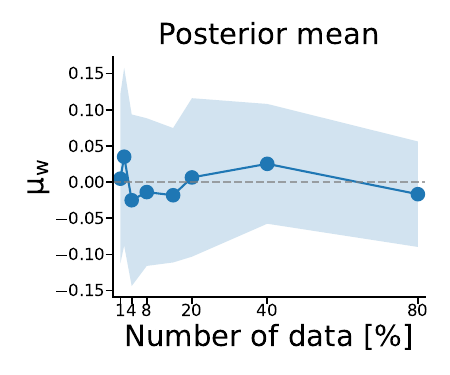}
        \caption{Boiling point (bp)} % (b) 
        \label{si_2_a}
    \end{subfigure}
    \begin{subfigure}{\textwidth}
        \centering
        \includegraphics[width=\panelwidth]{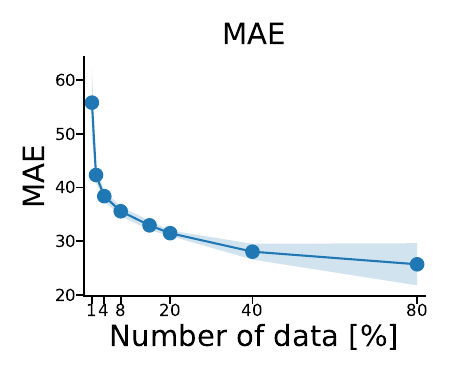}
        \includegraphics[width=\panelwidth]{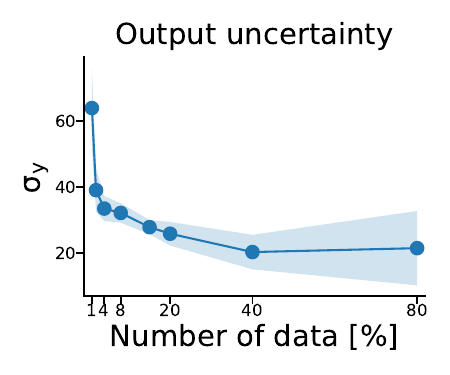}
        \includegraphics[width=\panelwidth]{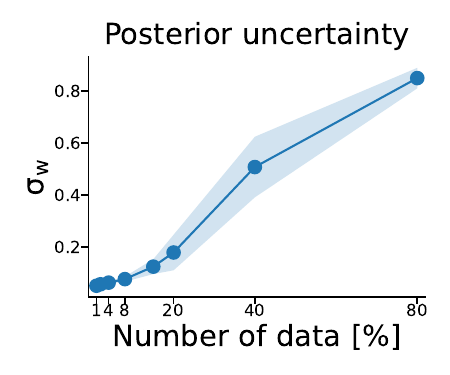}
        \includegraphics[width=\panelwidth]{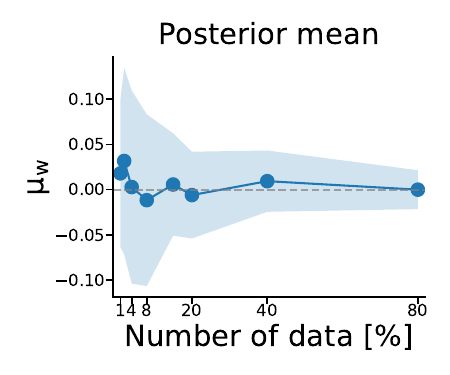}
        \caption{Melting point (mp)} % (b) 
        \label{si_2_b}
    \end{subfigure}
    % \begin{subfigure}{\textwidth}
    %     \centering
    %     \includegraphics[width=\panelwidth]{fig/si/2_paradox/logP_MAE.pdf}
    %     \includegraphics[width=\panelwidth]{fig/si/2_paradox/logP_pred_sigma.pdf}
    %     \includegraphics[width=\panelwidth]{fig/si/2_paradox/logP_sigma.pdf}
    %     \includegraphics[width=\panelwidth]{fig/si/2_paradox/logP_mu.pdf}
    %     \caption{} % (b) 
    %     \label{si_2_c}
    % \end{subfigure}
    \begin{subfigure}{\textwidth}
        \centering
        \includegraphics[width=\panelwidth]{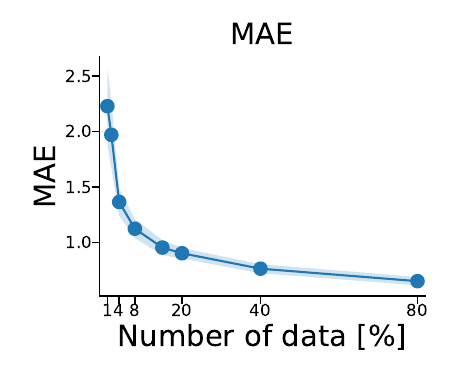}
        \includegraphics[width=\panelwidth]{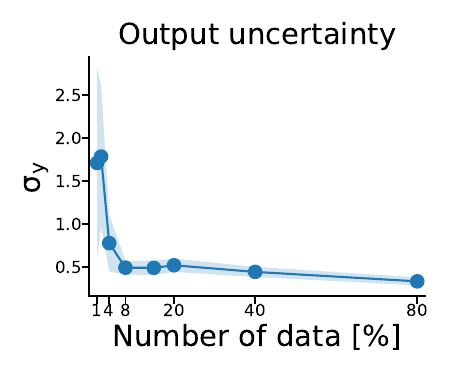}
        \includegraphics[width=\panelwidth]{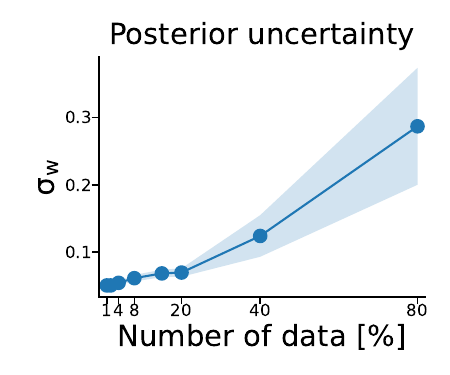}
        \includegraphics[width=\panelwidth]{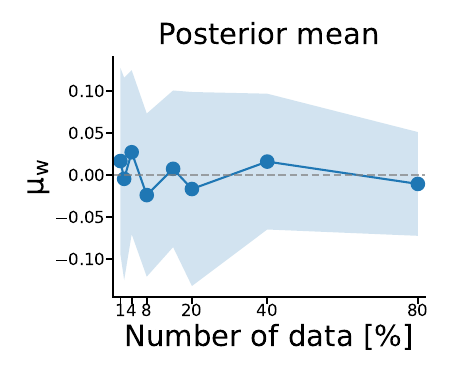}
        \caption{Vapor pressure (Pvap)} % (b) 
        % \label{si_2_d}
    \end{subfigure}
    \begin{subfigure}{\textwidth}
        \centering
        \includegraphics[width=\panelwidth]{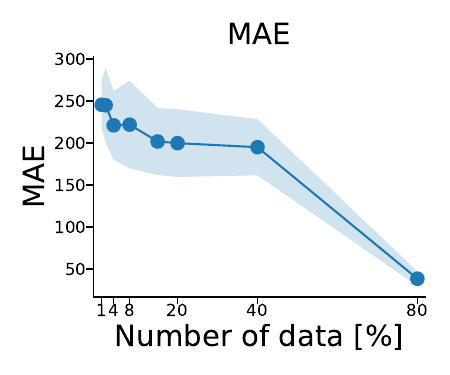}
        \includegraphics[width=\panelwidth]{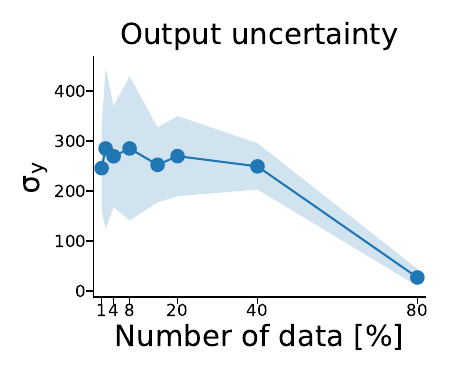}
        \includegraphics[width=\panelwidth]{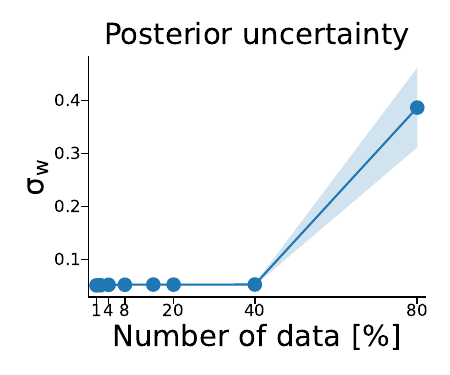}
        \includegraphics[width=\panelwidth]{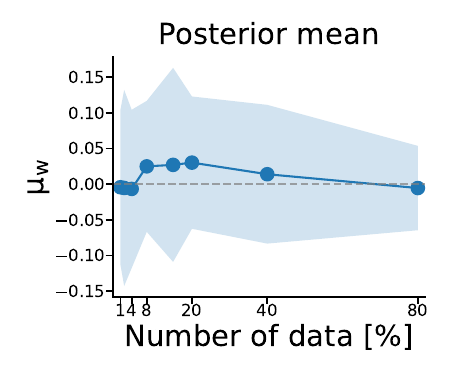}
        \caption{Enthalpy (H)} % (b) 
        % \label{si_2_e}
    \end{subfigure}
    \begin{subfigure}{\textwidth}
        \centering
        \includegraphics[width=\panelwidth]{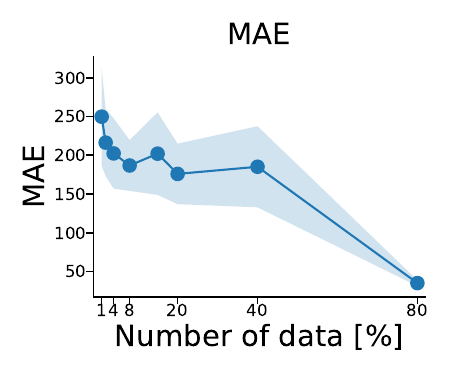}
        \includegraphics[width=\panelwidth]{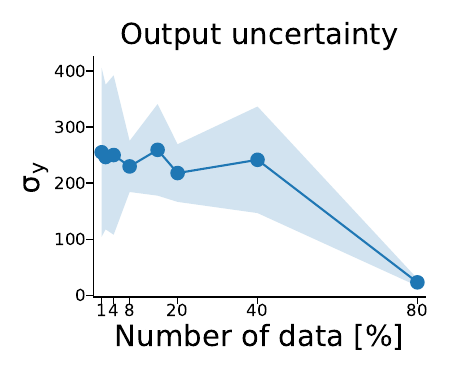}
        \includegraphics[width=\panelwidth]{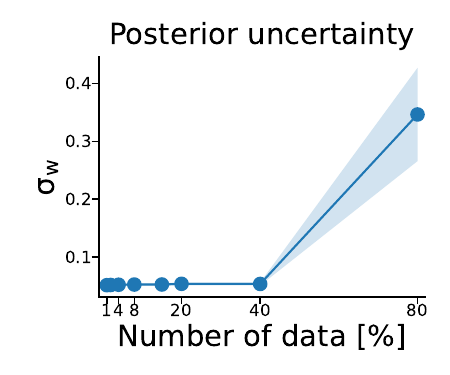}
        \includegraphics[width=\panelwidth]{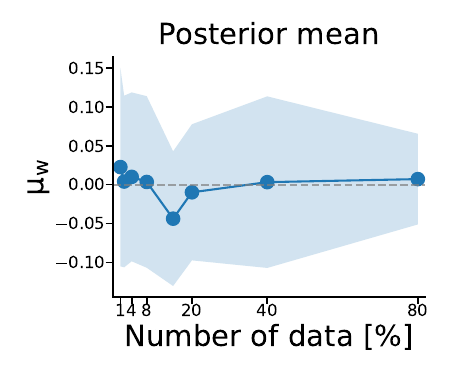}
        \caption{Gibbs free energy (G)} % (b) 
        % \label{si_2_f}
    \end{subfigure}
    
    \caption{\textbf{Verification of the non-classical uncertainty behavior.} Each row (a-e) corresponds to a distinct dataset, while columns show metrics as a function of training data size (from left to right): Mean Absolute Error (MAE), predictive uncertainty ($\sigma_{y}$), weight posterior uncertainty ($\sigma_{w}$), and weight posterior mean ($\mu_{w}$). Consistent with the main analysis, predictive uncertainty (second column) decreases while posterior uncertainty (third column) increases across all tasks.}
    \label{si_fig_paradox}
\end{figure}

To corroborate the decoupled behavior discovered in the main text, we repeated the posterior analysis for the other property datasets. We tracked the evolution of predictive performance (MAE), output uncertainty ($\sigma_{y}$), posterior uncertainty, and posterior mean as a function of training data size. As shown in Fig. \ref{si_fig_paradox}, all other datasets exhibit behavioral trends consistent with the primary analysis. Both the MAE and overall output uncertainty decrease as the training set grows, indicating improved model performance and confidence. However, the average standard deviation of weight posteriors systematically increases with data accumulation, directly contradicting the conventional expectation of posterior concentration. Throughout the process, the posterior means remain stable near zero. These results confirm that the non-classical behavior—the decoupling of predictive uncertainty from posterior uncertainty—is not an artifact of a specific task but a reproducible pattern across the six tested molecular-regression datasets under the same architecture and training protocol.

\section{Latent-Posterior Alignment and predictive uncertainty} \label{si_LPAS_uncertainty}

For $N$ latent vectors of dimension $d$,

\begin{align}
Z = [z_{1}, ... z_{N}]^\top \in \mathbb{R}^{N\times d}, \: C_{z}=\frac{1}{N}Z^{\top}Z, \: \Sigma_{w} = \text{diag}(\sigma_{w}^{2}).
\end{align}

Averaging the predictive uncertainty over samples gives
\begin{align}
\frac{1}{N}\sum^{N}_{i=1}\mathrm{Var}(y_{i} \mid z_{i}) 
  & = tr(C_{z}\Sigma_{w}) + \sigma_{b}^2 \\
  & = \sum_{j=1}^{d}(\frac{1}{N}\sum_{i=1}^{N}z_{i, j}^{2})\sigma_{j}^{2} + \sigma_{b}^{2} \\
  & = \sum_{j=1}^{d}m_{j}\sigma_{j}^{2} + \sigma_{b}^{2},
\label{var_avg_2}
\end{align}
where $m_{j} = \frac{1}{N}\sum_{i}z_{i,j}^{2}$ denotes the latent energy along coordinate $j$. Since $\sigma_{b}^{2}$ is empirically negligible, the uncertainty is dominated by the coupling term $\sum_{j}m_{j}\sigma_{j}^{2}$. To analyze how the Latent-Posterior Alignment (LPA) affects overall uncertainty, we consider two practical cases:

\begin{enumerate}
    \item Fixed posterior variance spectrum (eigenvalue-preserving): the multiset $\{\sigma_{j}^{2}\}$ is fixed.
    \item Budget constraints: either $\sum_{j}\sigma_{j}^{2}=C_{1}$ (fixed total variance) or $\sum_{j}\sigma_{j}=C_{2}$ (fixed total standard deviation).
\end{enumerate}

Both assumptions are practically consistent with how BNNs behave during training: the KL term in the variational objective acts as a soft regularizer toward the prior, keeping either the spectral shape of $\Sigma_{w}$ stable (Case 1) or the overall posterior variance $\sum_{j}\sigma_{j}^{2}$ approximately constant (Case 2). Hence, treating the posterior as having a fixed or budget-constrained variance profile is not an artificial simplification but reflects a realistic equilibrium of variational Bayesian training. These cases are idealized constraints used to isolate the effect of latent–posterior pairing; the training objective does not enforce either condition exactly.

\textbf{Case 1. Fixed posterior variance spectrum}

If $m_{j}$ and $\sigma_{j}^{2}$ are fixed nonnegative sequences, von Neumann's trace inequality provides bounds on the overall uncertainty \cite{mirsky1975trace}:
\begin{align}
\sum_{j=1}^{d}m_{j}^{\uparrow}\sigma_{j}^{\downarrow 2} \leq tr(C_{z}\Sigma_{w}) = \sum_{j=1}^{d}m_{j}\sigma_{j}^{2} \leq \sum_{j=1}^{d}m_{j}^{\uparrow}\sigma_{j}^{\uparrow 2}
\label{von_neumann}
\end{align}

The lower bound—corresponding to minimal predictive uncertainty—is achieved when large $m_{j}$ are paired with small $\sigma_{j}^{2}$, i.e., when the latent covariance $C_{z}$ and weight variance matrix $\Sigma_{w}$ are aligned. This condition formalizes the notion of the latent alignment, where the model reduces predictive uncertainty by directing high-energy latent dimensions toward low-uncertainty weight directions.

\textbf{Case 2. Budget constraints}

When total variance is fixed,
\begin{align}
\sum_{j=1}^{d}m_{j}\sigma_{j}^{2} \quad \text{s.t.} \quad \sum_{j}\sigma_{j}^{2} = C_{1}
\label{budget_1}
\end{align}
the optimization is a linear program over a simplex. Its optimum lies at an extreme point, concentrating all variance on the dimension $j^{\star}\in \text{argmin}_{j}m_{j}$. Thus, predictive variance is minimized when posterior uncertainty is concentrated along latent directions with the smallest energy—an extreme form of the alignment.

Under fixed total standard deviation,
\begin{align}
\sum_{j=1}^{d}m_{j}\sigma_{j}^{2} \quad \text{s.t.} \quad \sum_{j}\sigma_{j} = C_{2}
\label{budget_2}
\end{align}
the objective is strictly convex. From the KKT conditions for equality constraints, $2m_{j}\sigma_{j} + \lambda=0$ where $\lambda$ is a Lagrange multiplier, implying $\sigma_{j}$ and $m_{j}$ are inversely proportional to minimize uncertainty, thereby concentrating $\sigma_{j}$ on small $m_{j}$. Hence, both budget-constrained formulations yield the same aligned solution, confirming that predictive uncertainty is minimized when posterior variance aligns inversely with latent energy. These results indicate that the model structurally reorganizes the latent representations during training to minimize the predictive uncertainty.
\clearpage

\section{Cross-dataset consistency of Latent-Posterior Alignment} \label{si_anti_alignment}

\begin{figure}[H]
    \centering
    \begin{subfigure}{\textwidth}
        \centering
        \includegraphics[width=\panelwidth]{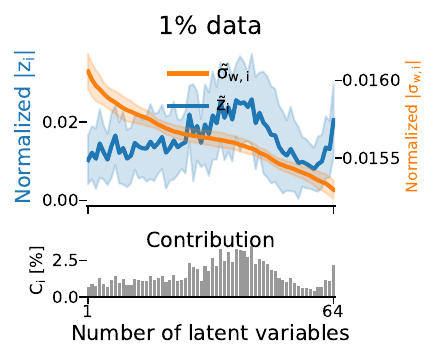}
        \includegraphics[width=\panelwidth]{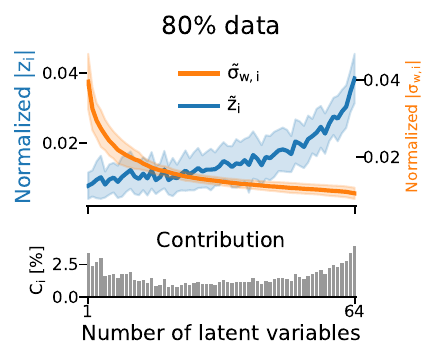}
        \includegraphics[width=\panelwidth]{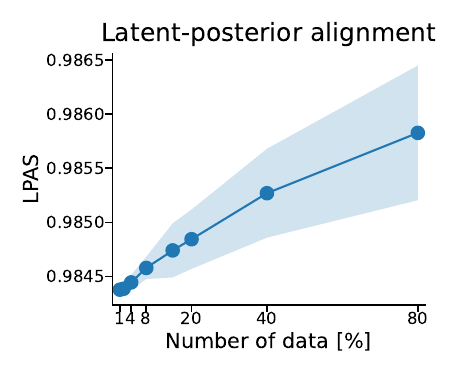}
        \includegraphics[width=\panelwidth]{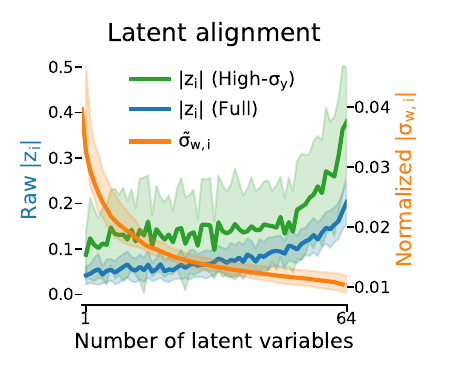}
        \caption{Boiling point (bp)} % (b) 
        \label{si_3_a}
    \end{subfigure}
    \begin{subfigure}{\textwidth}
        \centering
        \includegraphics[width=\panelwidth]{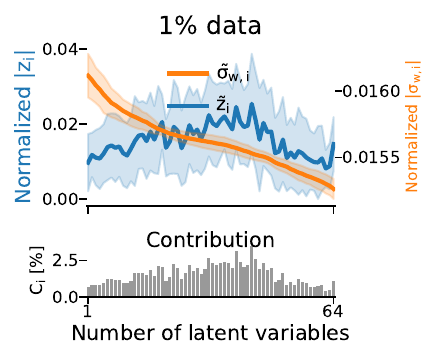}
        \includegraphics[width=\panelwidth]{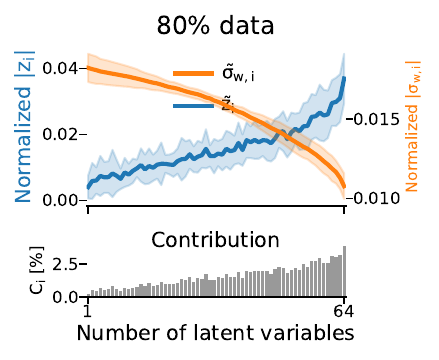}
        \includegraphics[width=\panelwidth]{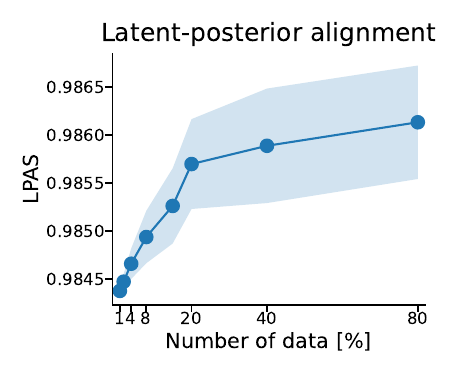}
        \includegraphics[width=\panelwidth]{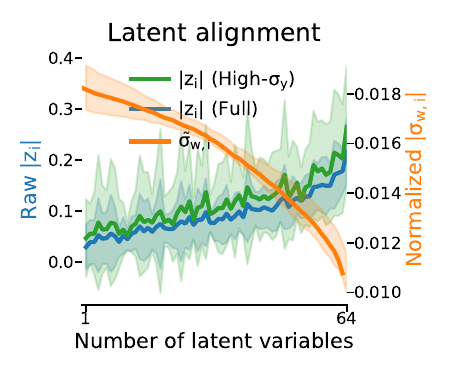}
        \caption{Melting point (mp)} % (b) 
        \label{si_3_b}
    \end{subfigure}
    % \begin{subfigure}{\textwidth}
    %     \centering
    %     \includegraphics[width=\panelwidth]{fig/si/3_alignment/logP_alignment_0.1.pdf}
    %     \includegraphics[width=\panelwidth]{fig/si/3_alignment/logP_alignment_8.pdf}
    %     \includegraphics[width=\panelwidth]{fig/si/3_alignment/logP_LPAS.pdf}
    %     \includegraphics[width=\panelwidth]{fig/si/3_alignment/logP_top.pdf}
    %     \caption{} % (b) 
    %     \label{si_2_c}
    % \end{subfigure}
    \begin{subfigure}{\textwidth}
        \centering
        \includegraphics[width=\panelwidth]{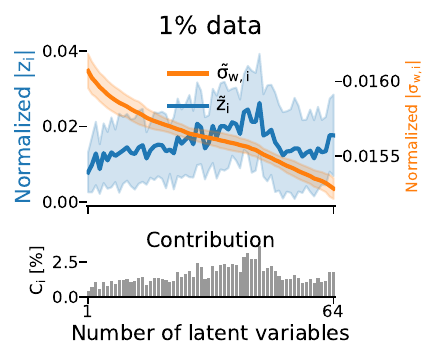}
        \includegraphics[width=\panelwidth]{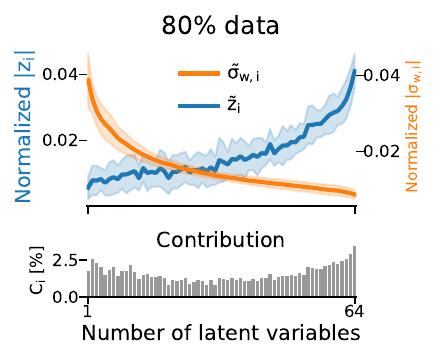}
        \includegraphics[width=\panelwidth]{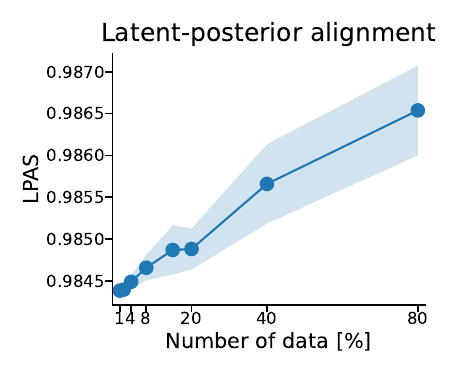}
        \includegraphics[width=\panelwidth]{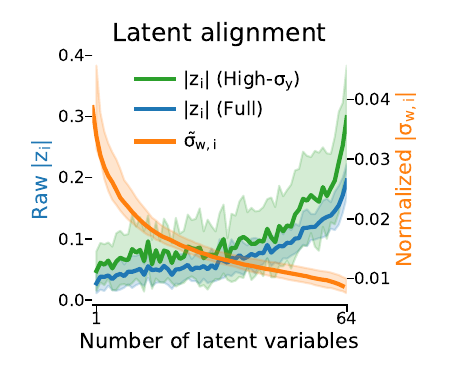}
        \caption{Vapor pressure (Pvap)} % (b) 
        % \label{si_2_d}
    \end{subfigure}
    \begin{subfigure}{\textwidth}
        \centering
        \includegraphics[width=\panelwidth]{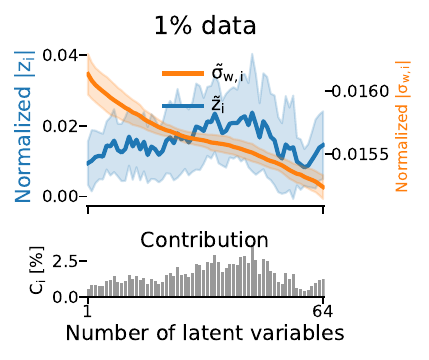}
        \includegraphics[width=\panelwidth]{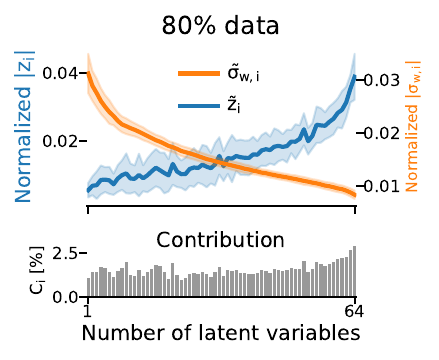}
        \includegraphics[width=\panelwidth]{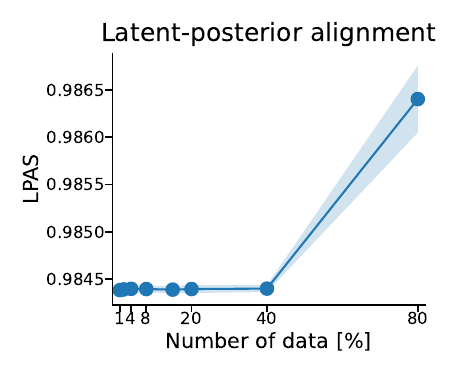}
        \includegraphics[width=\panelwidth]{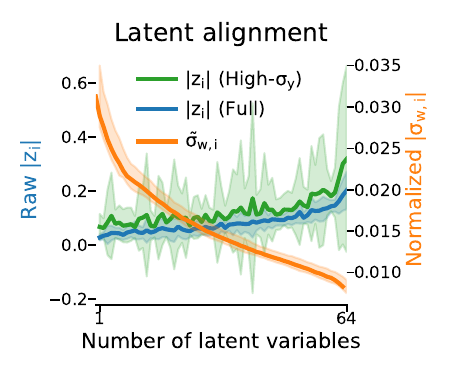}
        \caption{Enthalpy (H)} % (b) 
        % \label{si_2_e}
    \end{subfigure}
    \begin{subfigure}{\textwidth}
        \centering
        \includegraphics[width=\panelwidth]{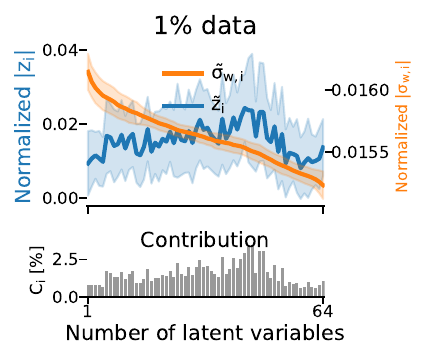}
        \includegraphics[width=\panelwidth]{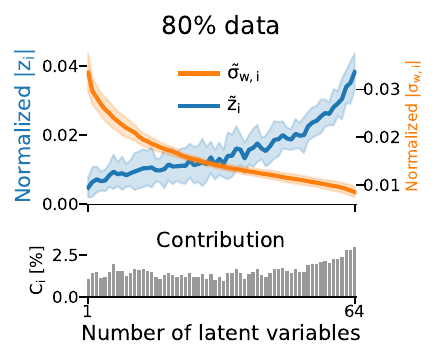}
        \includegraphics[width=\panelwidth]{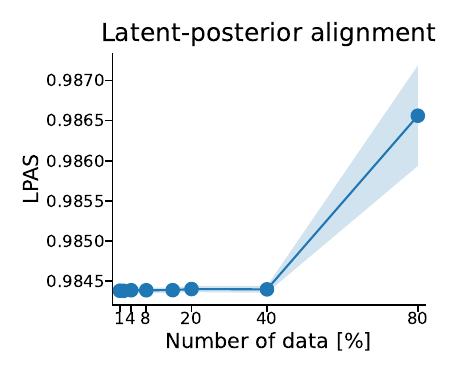}
        \includegraphics[width=\panelwidth]{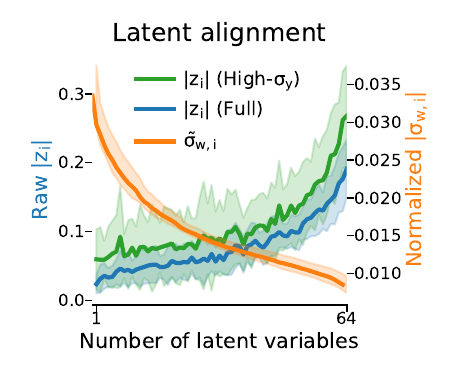}
        \caption{Gibbs free energy (G)} % (b) 
        % \label{si_2_f}
    \end{subfigure}
    
    \caption{\textbf{Cross-dataset consistency of Latent-Posterior Alignment.} Results are presented for five different datasets (a-e). Columns 1–2: Comparison of latent geometry between data-poor (1\%) and data-rich (80\%) regimes. The plots overlay the normalized latent vector ($\tilde{\mathbf{z}}$, blue), posterior standard deviation ($\tilde{\sigma}_{w}$, orange), and relative uncertainty contribution (grey bars). Column 3: The Latent-Posterior Alignment Score (LPAS) versus data size. Column 4: Comparison of latent structure between the overall data distribution (blue) and samples with the highest predictive uncertainty (green).}
    \label{si_fig_alignment}
\end{figure}

To confirm that Latent-Posterior Alignment (LPA) is a consistent pattern, we extended our analysis to other property datasets (Fig. \ref{si_fig_alignment}). First, a visual comparison between the data-poor (1\%, first column) and data-rich (80\%, second column) regimes reveals a consistent emergence of Latent-Posterior Alignment across all tasks. In the data-rich regime, the normalized latent vectors ($\tilde{\mathbf{z}}$, blue) shift toward dimensions where the posterior standard deviation ($\tilde{\sigma}_{w}$, orange) is minimized. This geometric reorganization confirms that the model actively learns to utilize stable weight dimensions as data accumulates.

Second, this shift fundamentally alters the composition of predictive uncertainty. The relative contribution analysis ($C_{i}$, grey bars) demonstrates that as the model utilizes more data, the uncertainty budget is increasingly reallocated toward these low-variance dimensions. While the large magnitudes of some high-variance dimensions inevitably contribute to total uncertainty, the dominant trend is a systematic shift of contribution toward “reliable” axes, confirming that the model anchors its predictions on stable dimensions.

Third, this structural evolution is quantitatively validated by the Latent-Posterior Alignment Score (LPAS, third column). Across all datasets, the LPAS increases with training-set size across the tested datasets, providing robust evidence that LPA is the key strategy.

Finally, the analysis of high-uncertainty samples corroborates the dual-geometric mechanism proposed in the main text. We compared the latent representations of the entire dataset (blue) against the top 5\% of samples with the highest predictive uncertainty (green). Crucially, high-uncertainty samples maintain the same aligned directionality while exhibiting significantly amplified latent magnitudes, as shown in the fourth column of Fig. \ref{si_fig_alignment}. This confirms that the model regulates uncertainty through a two-tiered strategy: global alignment sets a stable baseline, while latent magnitude encodes instance-specific epistemic risk.

\section{Formulation of latent regularization objectives} \label{si_z_regularization_formulation}
To test the causal role of Latent-Posterior Alignment (LPA) in predictive uncertainty, we introduced L1, L2, and anti-alignment penalties directly into the standard ELBO objective to inhibit the formation of LPA as follows:

\begin {align}
\mathcal{L}_{L1} = -\text{ELBO} + \lambda_{l1} \| z \|_{1}
\end {align}

\begin {align}
\mathcal{L}_{L2} = -\text{ELBO} + \lambda_{l2} \| z \|_{2}
\end {align}

\begin {align}
\mathcal{L}_{anti-align} = -\text{ELBO} - \lambda_{anti-align} \frac{\sum_{i=1}^{d} |z_{i}|\sigma_{i}}{\sum_{i=1}^{d}|z_{i}|\sum_{i=1}^{d}\sigma_{i}}
\end {align}

The first two interventions (L1 and L2) aim to suppress the capacity of the latent vector $\mathbf{z}$ to form strong directional concentrations or large magnitudes, which are essential for LPA. Specifically, the L1 term enforces a strict sparsity penalty, while the L2 term penalizes the Euclidean norm of $\mathbf{z}$, thereby hindering the formation of sharp alignment. The anti-alignment penalty term utilizes the negative value of the Latent-Posterior Alignment Score (LPAS) to directly test causality. This constraint actively pushes the latent representation into uncertain dimensions where $\sigma_{w}$ is large, forbidding the model from utilizing the reliable subspaces it naturally seeks. Since the magnitudes of the ELBO and regularization terms vary across different datasets and properties, the regularization strength $\lambda$ was empirically tuned for each task. Regularization strengths were selected using a pre-specified grid and validation data only. The test set was not used to select $\lambda$. The specific hyperparameters used for all experiments are shown in Table \ref{si_table_lambda} while $\gamma$ is the regularization weight for Alignment-Guided Learning.

\begin{table}[h]
\centering
\begin{tabular}{lcccc}
\toprule
\textbf{Dataset (Property)} & \textbf{$\lambda_{l1}$} & \textbf{$\lambda_{l2}$} & \textbf{$\lambda_{anti-align}$} & \textbf{$\gamma$}\\
\midrule
bp   & 0.05 & 0.2 & 0.1 & 0.05\\
mp   & 0.3 & 1.5 & 0.5 & 0.02\\
P    & 0.05 & 0.3  & 0.05 & 0.02\\
Pvap & 0.02 & 0.1  & 0.01 & 0.005\\
H    & 0.02  & 0.2  & 0.1 & 0.05\\
G    & 0.02  & 0.1  & 0.05 & 0.02\\
\bottomrule
\end{tabular}
\caption{\textbf{Hyperparameters for latent regularization experiments and Alignment-Guided Learning.} The regularization and AGL strengths ($\lambda$ and $\gamma$) were adjusted for each dataset.}
\label{si_table_lambda}
\end{table}

\section{Extended interventional evidence across datasets} \label{si_z_regularization}
\begin{figure}[H]
    \centering
    \begin{subfigure}{\textwidth}
        \centering
        \includegraphics[width=0.16\textwidth]{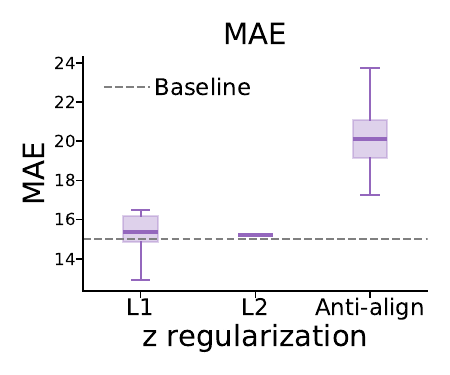}
        \includegraphics[width=0.16\textwidth]{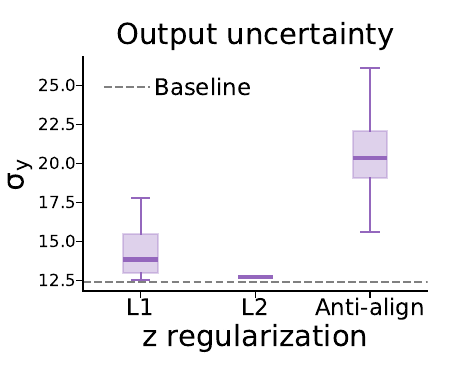}
        \includegraphics[width=0.16\textwidth]{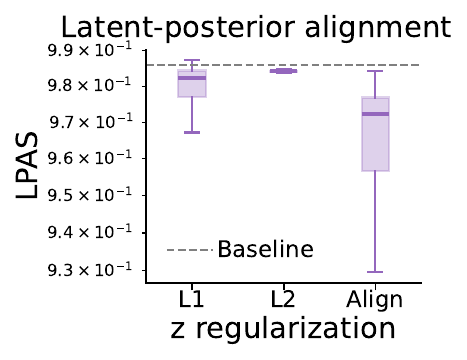}
        \includegraphics[width=0.16\textwidth]{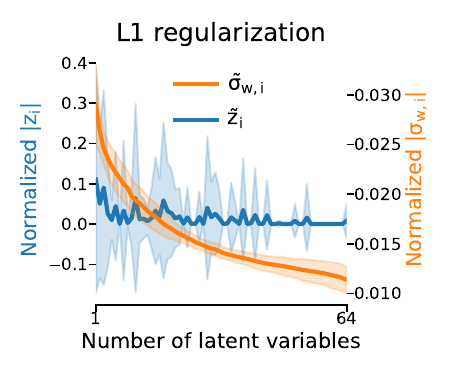}
        \includegraphics[width=0.16\textwidth]{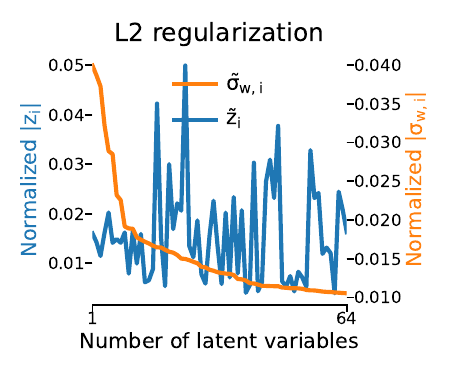}
        \includegraphics[width=0.16\textwidth]{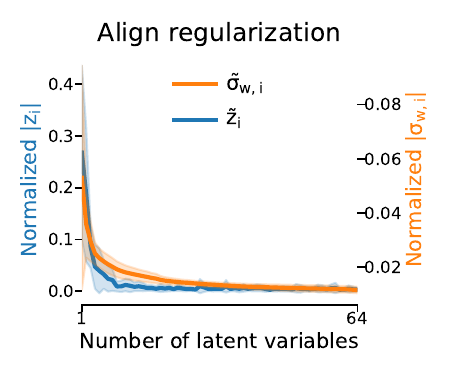}
        \caption{Boiling point (bp)} 
        \label{si_4_a}
    \end{subfigure}
    \begin{subfigure}{\textwidth}
        \centering
        \includegraphics[width=0.16\textwidth]{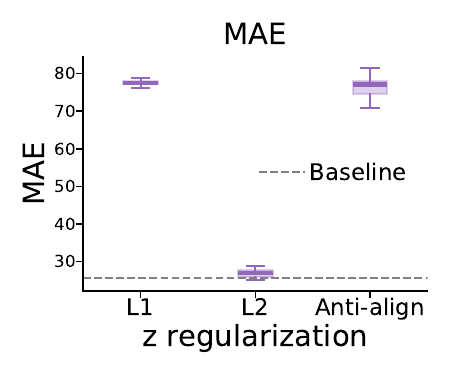}
        \includegraphics[width=0.16\textwidth]{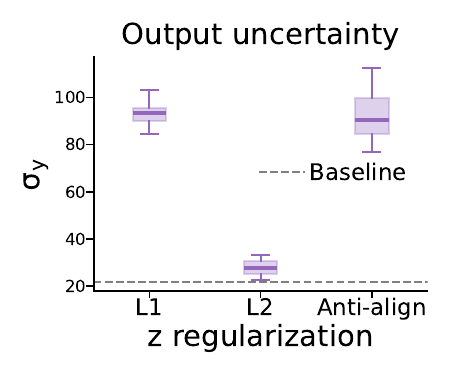}
        \includegraphics[width=0.16\textwidth]{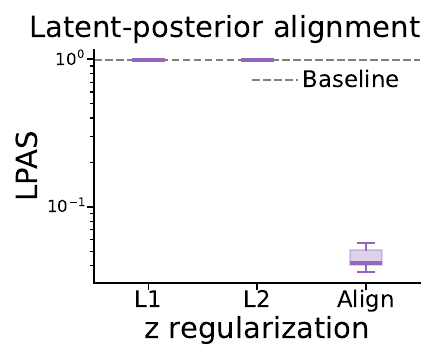}
        \includegraphics[width=0.16\textwidth]{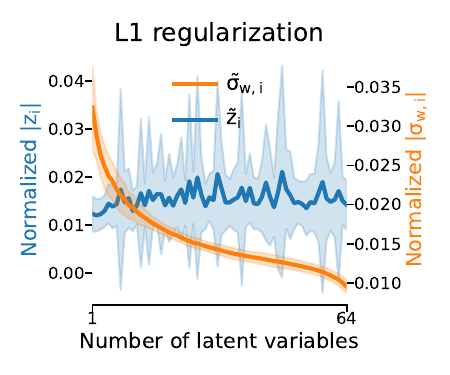}
        \includegraphics[width=0.16\textwidth]{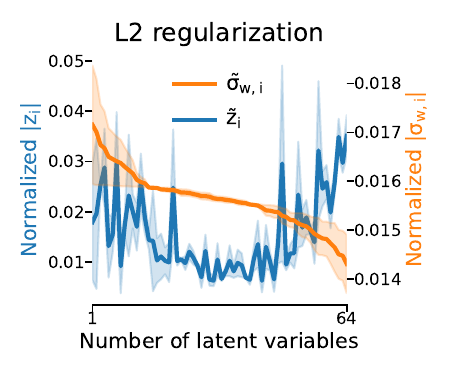}
        \includegraphics[width=0.16\textwidth]{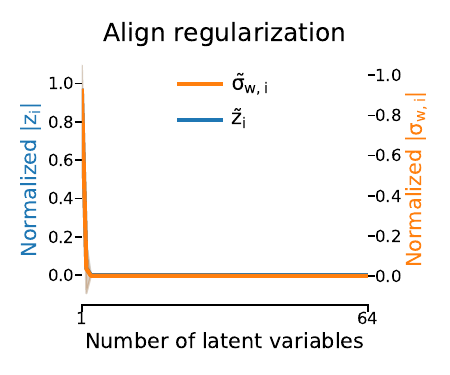}
        \caption{Melting point (mp)} % (b) 
        \label{si_4_b}
    \end{subfigure}
    % \begin{subfigure}{\textwidth}
    %     \centering
    %     \includegraphics[width=0.16\textwidth]{fig/si/4_z_test/logP_mae.pdf}
    %     \includegraphics[width=0.16\textwidth]{fig/si/4_z_test/logP_sigma.pdf}
    %     \includegraphics[width=0.16\textwidth]{fig/si/4_z_test/logP_LPAS.pdf}
    %     \includegraphics[width=0.16\textwidth]{fig/si/4_z_test/logP_l_1.pdf}
    %     \includegraphics[width=0.16\textwidth]{fig/si/4_z_test/logP_l_2.pdf}
    %     \includegraphics[width=0.16\textwidth]{fig/si/4_z_test/logP_align.pdf}
    %     \caption{} % (b) 
    %     \label{si_4_c}
    % \end{subfigure}
    \begin{subfigure}{\textwidth}
        \centering
        \includegraphics[width=0.16\textwidth]{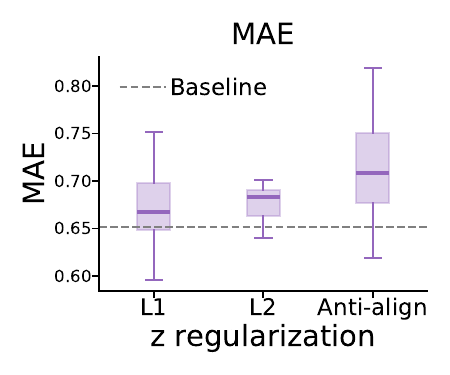}
        \includegraphics[width=0.16\textwidth]{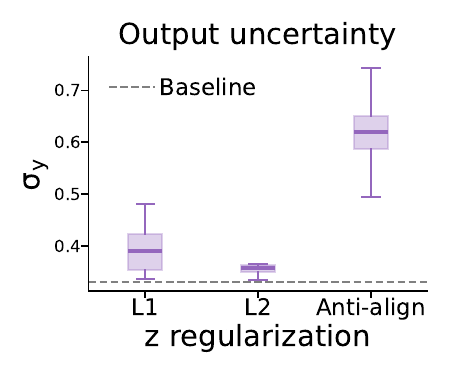}
        \includegraphics[width=0.16\textwidth]{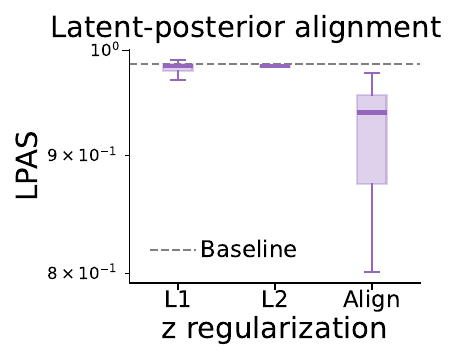}
        \includegraphics[width=0.16\textwidth]{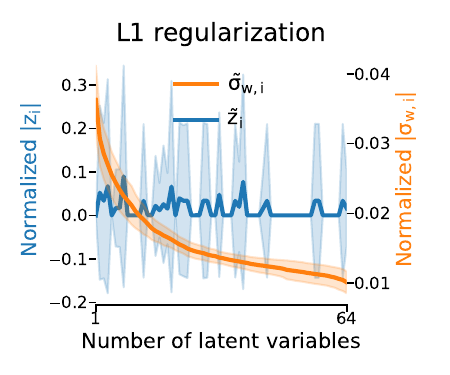}
        \includegraphics[width=0.16\textwidth]{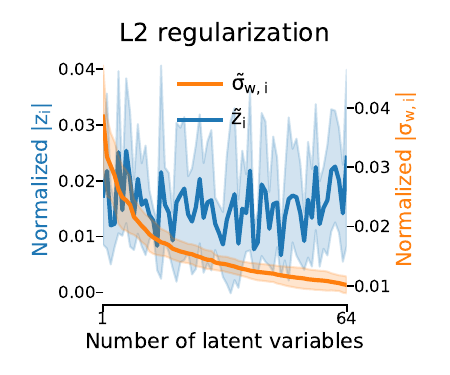}
        \includegraphics[width=0.16\textwidth]{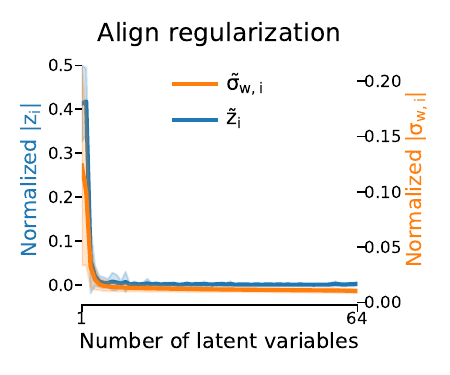}

        \caption{Vapor pressure (Pvap)} % (b) 
        \label{si_4_d}
    \end{subfigure}
    \begin{subfigure}{\textwidth}
        \centering
        \includegraphics[width=0.16\textwidth]{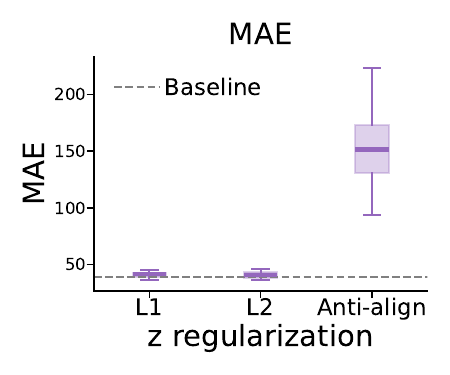}
        \includegraphics[width=0.16\textwidth]{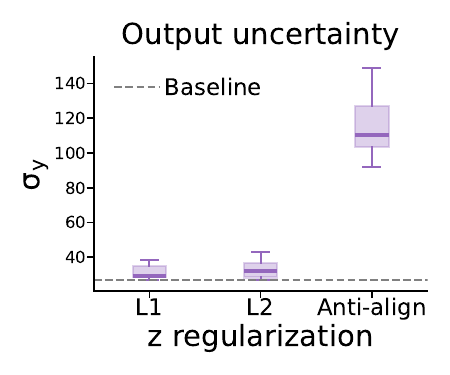}
        \includegraphics[width=0.16\textwidth]{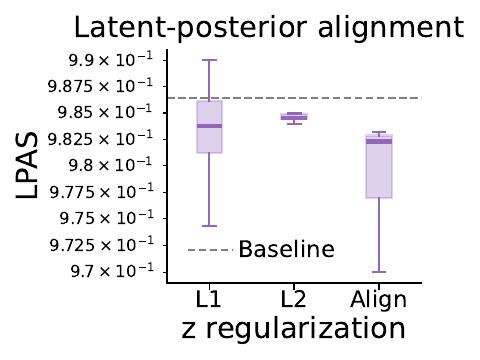}
        \includegraphics[width=0.16\textwidth]{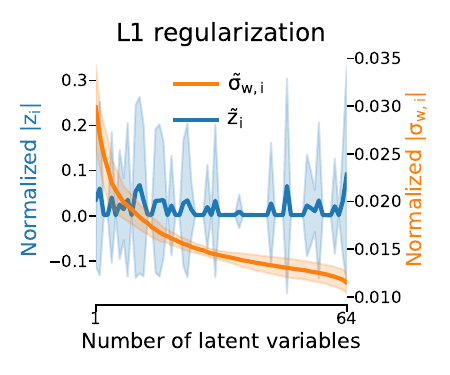}
        \includegraphics[width=0.16\textwidth]{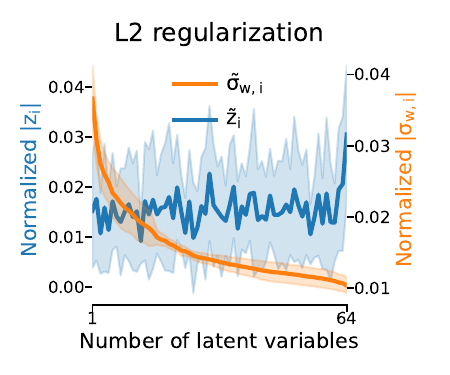}
        \includegraphics[width=0.16\textwidth]{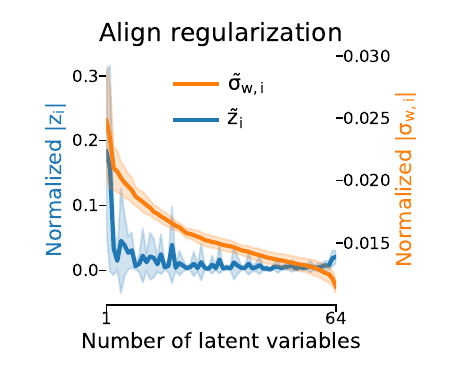}
        \caption{Enthalpy (H)} % (b) 
        \label{si_4_e}
    \end{subfigure}
    \begin{subfigure}{\textwidth}
        \centering
        \includegraphics[width=0.16\textwidth]{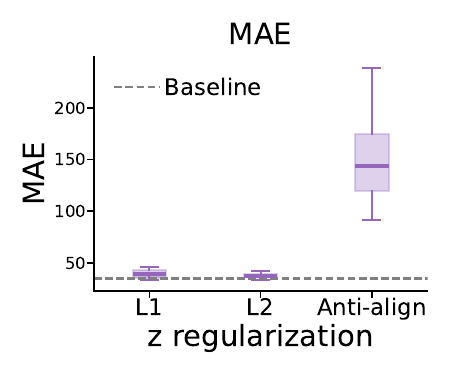}
        \includegraphics[width=0.16\textwidth]{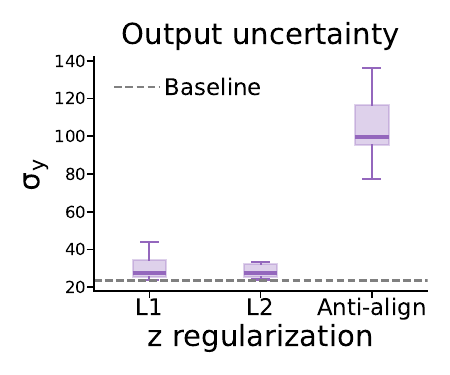}
        \includegraphics[width=0.16\textwidth]{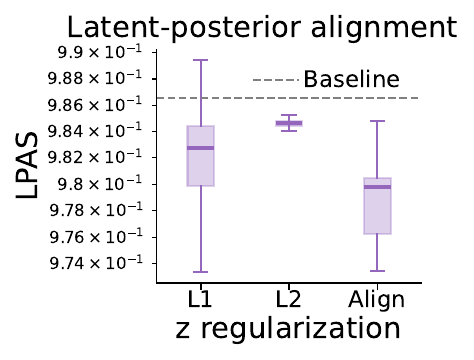}
        \includegraphics[width=0.16\textwidth]{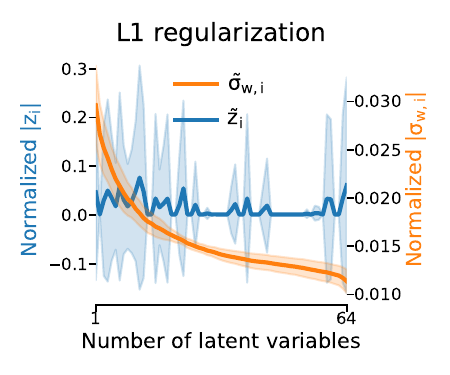}
        \includegraphics[width=0.16\textwidth]{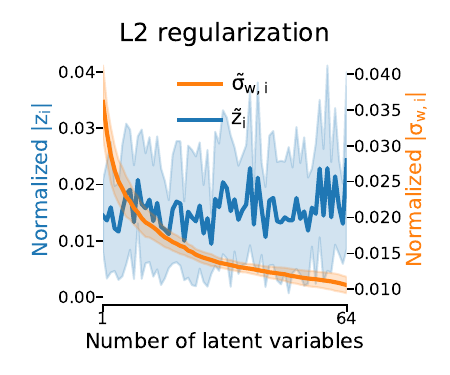}
        \includegraphics[width=0.16\textwidth]{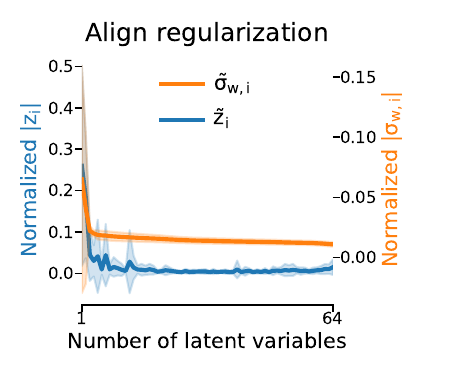}
        \caption{Gibbs free energy (G)} % (b) 
        \label{si_4_f}
    \end{subfigure}
    
    \caption{Impact of latent regularization on predictive uncertainty and latent geometry. Rows (a-e) correspond to five different datasets. Columns 1-3: Comparison of MAE, predictive uncertainty, and LPAS between baseline (dashed line) and regularized models. Columns 4-6: Visualization of latent vector and posterior standard deviation under L1, L2, and anti-alignment penalties, respectively.}
    \label{si_fig_z_regularization}
\end{figure}

Using the formulation in Section \ref{si_z_regularization_formulation}, we verified the causal role of LPA by extending the interventional experiments to five additional datasets. Fig. \ref{si_fig_z_regularization} shows the impact of three regularization schemes (L1, L2, and anti-alignment penalty) on model performance and latent geometry. Consistent with the main analysis, all regularized models exhibit degradation in predictive performance (MAE, Column 1) and a substantial increase in predictive uncertainty ($\sigma_{y}$, Column 2) compared to the unconstrained baseline model. This strictly correlates with the breakdown of the LPA mechanism, as evidenced by decreased Latent-Posterior Alignment Score values (Column 3) and the disruption of latent structures (Columns 4–6). This confirms that preventing the formation of LPA fundamentally deteriorates the model’s ability to minimize uncertainty.

We also observed a hierarchy in the degradation, generally ordered as anti-alignment $>$ L1 $>$ L2 (Anti-alignment penalty and L1 induced greater uncertainty and MAE than L2). This trend can be attributed to the geometric nature of the methods. The L2 penalty tends to induce a diffuse distribution, where the latent vector retains non-zero magnitudes across most hidden dimensions (Column 5). In contrast, L1 and anti-alignment penalties force the latent representation into a highly sparse or sharp regime, activating only a few dimensions (Columns 4 and 6). This severe restriction on latent geometry appears to drastically hinder the model's capacity to fully utilize dimensional information. Note that the precise ranking of degradation may vary depending on hyperparameter tuning ($\lambda$) specific to each dataset.

\clearpage

\section{Robustness analysis} \label{si_robustness}

\begin{figure}[]
    \centerline{\includegraphics[width=\columnwidth]{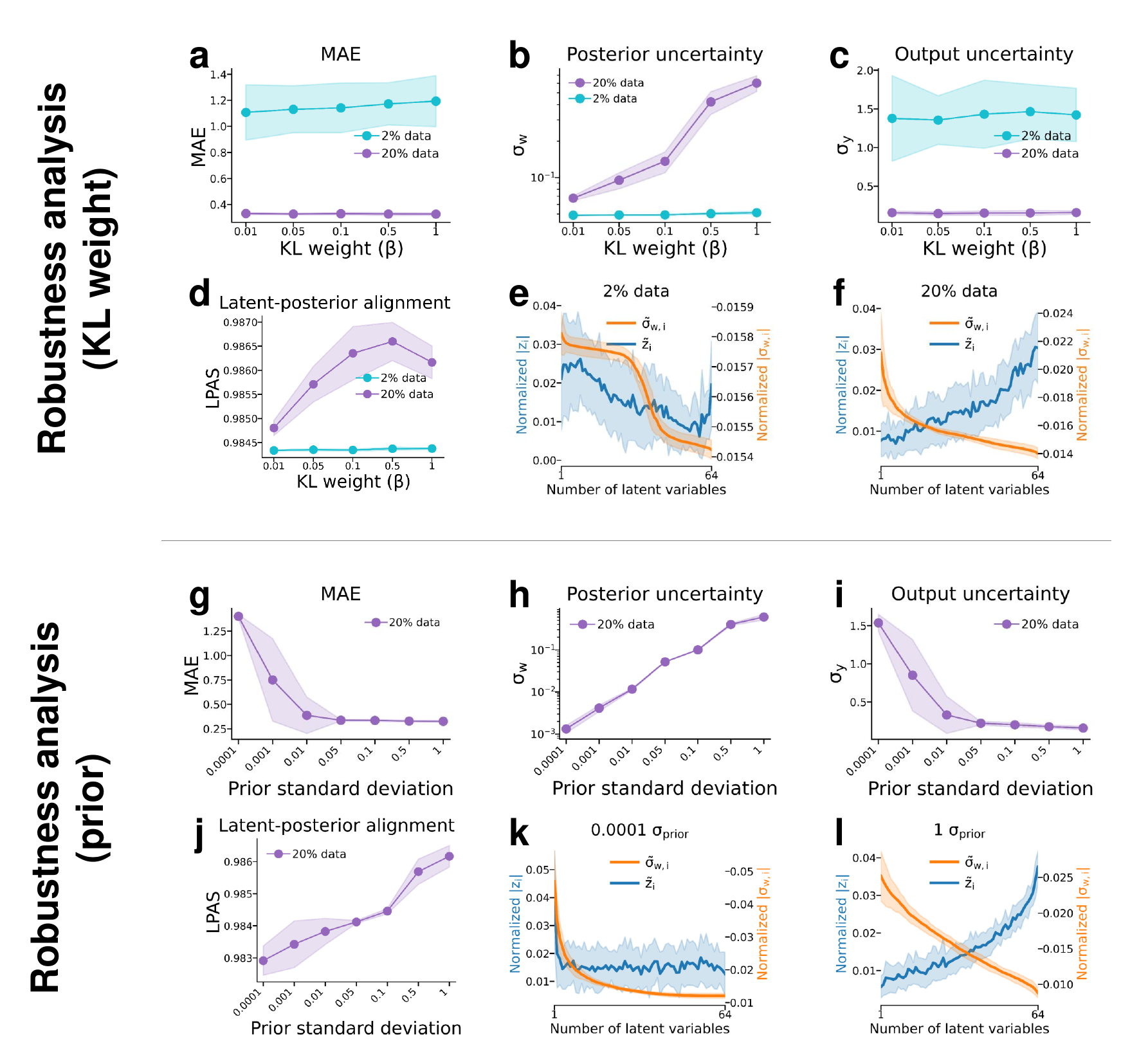}}
    \caption{Robustness of latent-posterior alignment mechanism.
    (a–f) Robustness to KL weight. Reducing $\beta$ leads to smaller $\sigma_{w}$ (b). However, this variation has negligible impact on accuracy (a) and output uncertainty (c). This invariance arises because LPA serves as a robust geometric mechanism: data-rich models maintain a consistently high alignment score (d, purple) and a clearly aligned latent-posterior geometry (f) regardless of $\beta$. In contrast, data-poor models exhibit low LPAS (d, cyan) and fail to form such structure (e), confirming that predictive confidence is governed by geometric alignment rather than posterior magnitude. (g-l) Impact of prior variance constraints. Restricting the prior standard deviation (x-axis) forces posterior contraction (h) but paradoxically increases predictive error (g) and output uncertainty (i). This failure arises because tight priors restrict the parameter-space flexibility required to maintain LPA, as evidenced by the decay of alignment score (j) and disrupted geometry (k) compared to the relaxed prior (l). Shaded bands denote standard deviation across $n=30$ independent runs.}
    \label{si_5_6_robustness} 
\end{figure}

We evaluated the robustness of this alignment mechanism to determine whether it represents a general principle or is contingent upon specific training configurations. We first investigated the influence of the KL-divergence weight ($\beta$), which governs the strength of posterior regularization. We hypothesized that the posterior's failure to govern uncertainty might stem from excessive regularization pressure. By reducing $\beta$, we relaxed this constraint to determine whether the posterior would reclaim its role, or if the model would consistently persist in utilizing the Latent-Posterior Alignment as its dominant strategy.

As shown in Fig. \ref{si_5_6_robustness}a--f, we compared representative data-poor (2\% data, high uncertainty, cyan) and data-rich (20\% data, low uncertainty, purple) regimes, analyzing how their respective strategies evolved under varying $\beta$. As expected, lowering $\beta$ weakens the regularization, leading to a reduction in the posterior standard deviation (Fig. \ref{si_5_6_robustness}b). This effect is particularly pronounced in the data-rich regime, where the baseline posterior variance was originally substantial. Crucially, however, this concentration of the posterior did not translate to the model's predictive behavior. Both predictive accuracy (MAE) (Fig. \ref{si_5_6_robustness}a) and output uncertainty (Fig. \ref{si_5_6_robustness}c) remained largely invariant to changes in $\beta$. Specifically, the data-rich models maintained a low uncertainty profile despite the drastic tightening of their posteriors. This suggests that the posterior variance itself is not the primary driver of predictive uncertainty. Instead, the uncertainty dynamics were consistently governed by the alignment mechanism.

The LPAS in Fig. \ref{si_5_6_robustness}d reveals a fundamental distinction between the regimes that persists regardless of $\beta$. In the data-poor case, LPAS values remain consistently low and invariant to $\beta$, indicating a persistent absence of the alignment mechanism (Fig. \ref{si_5_6_robustness}e). Conversely, in the data-rich case, the models robustly maintain substantially higher LPAS values; although minor variations exist depending on $\beta$, they exhibit strong alignment (Fig. \ref{si_5_6_robustness}f). This confirms that the adoption of Latent-Posterior Alignment is a robust strategy, independent of the posterior's flexibility. We verified that this resilience to $\beta$ variations is consistent across diverse molecular benchmarks (see Supplementary Fig. \ref{si_fig_kl_test}).

We further extended this robustness analysis by investigating the effect of the prior distribution for the data-rich case (20\% data) as shown in Fig. \ref{si_5_6_robustness}g--l. We hypothesized that imposing a highly restrictive and low-variance prior on the weights would force the posterior variance $\sigma_{w}^{2}$ to shrink. If posterior variance were indeed the dominant factor, this should directly reduce predictive uncertainty. As expected, narrowing the prior caused the posterior standard deviations to decrease (Fig. \ref{si_5_6_robustness}h). However, despite this reduction in the posterior distribution, predictive uncertainty gradually increased (Fig. \ref{si_5_6_robustness}i). This counterintuitive trend arises from the disruption of latent alignment: as the standard deviation of the prior decreases, the alignment score decreases (Fig. \ref{si_5_6_robustness}j), revealing that the latent vectors begin to concentrate toward directions where the posterior variance is large (Fig. \ref{si_5_6_robustness}k), thereby amplifying uncertainty. Similar results under restrictive priors were reproduced in other tasks (Supplementary Fig. \ref{si_fig_prior_test}).

While the model can tolerate moderate constraints, our experiments revealed a critical failure point (Fig. \ref{si_5_6_robustness}g--j). When the prior standard deviation is excessively reduced (e.g., below 0.01), the model fails to maintain coherent latent alignment (Fig. \ref{si_5_6_robustness}k) compared to the relaxed model (Fig. \ref{si_5_6_robustness}l), leading to a sharp increase in predictive uncertainty and error. This finding underscores that the model requires a sufficient degree of “exploratory variance” (parameter-space flexibility) to discover and leverage effective alignment pathways. Artificially suppressing this flexibility proves counterproductive, ultimately disrupting the very mechanism the model relies on for reliable predictions.

Relaxing posterior regularization did not diminish the role of LPA, whereas attempting to circumvent it via narrow priors proved detrimental. Supported by our extensive validation across multiple datasets (see Supplementary Figs.~\ref{si_fig_z_regularization}, \ref{si_fig_kl_test}, \ref{si_fig_prior_test}), these findings confirm that LPA remains associated with predictive uncertainty, supporting its robust contribution within the deterministic GNN feature extractors and mean-field Bayesian output layers examined here.

\subsection{Effect of KL-divergence weight on LPA} \label{si_kl_test}
\begin{figure}[H]
    \centering
    \begin{subfigure}{\textwidth}
        \centering
        \includegraphics[width=0.16\textwidth]{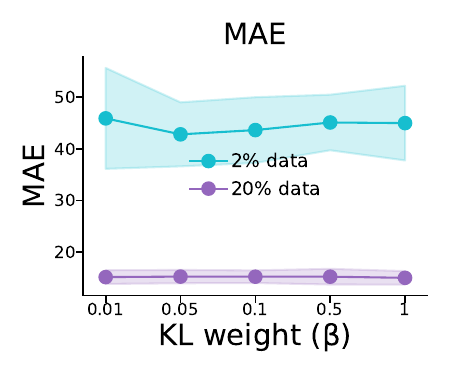}
        \includegraphics[width=0.16\textwidth]{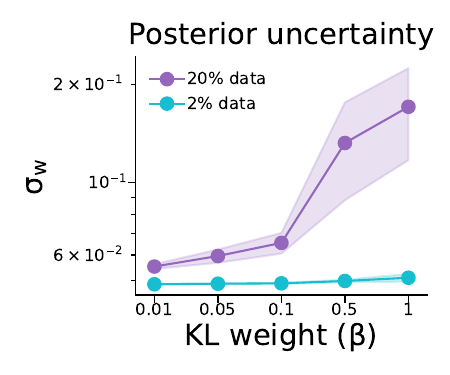}
        \includegraphics[width=0.16\textwidth]{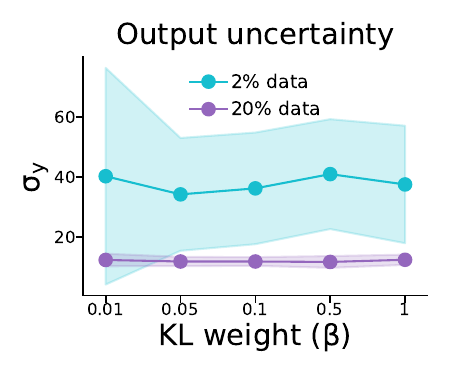}
        \includegraphics[width=0.16\textwidth]{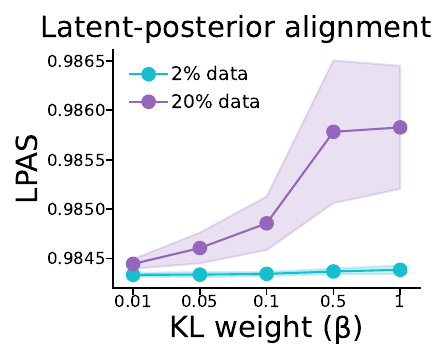}
        \includegraphics[width=0.16\textwidth]{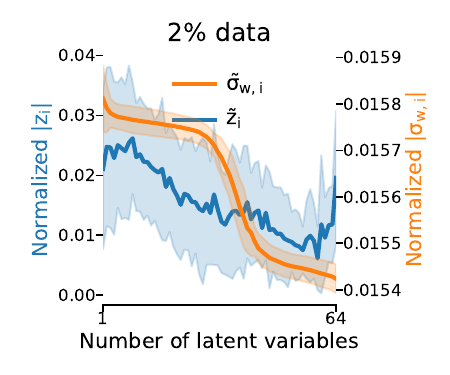}
        \includegraphics[width=0.16\textwidth]{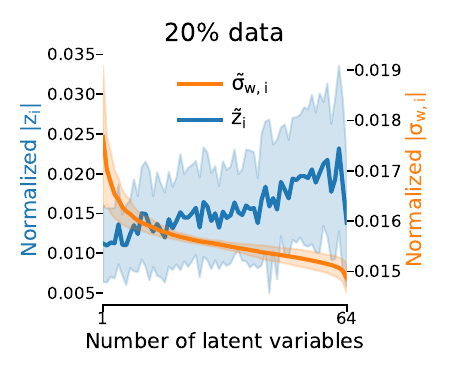}
        \caption{Boiling point (bp)} % (b) 
        % \label{si_5_a}
    \end{subfigure}
    \begin{subfigure}{\textwidth}
        \centering
        \includegraphics[width=0.16\textwidth]{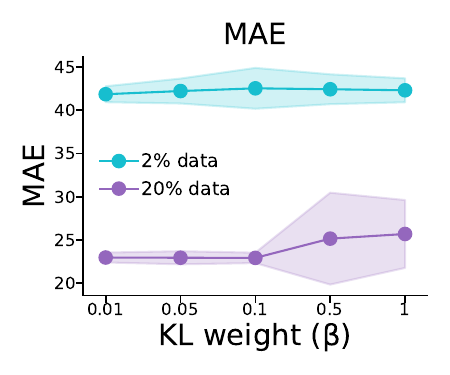}
        \includegraphics[width=0.16\textwidth]{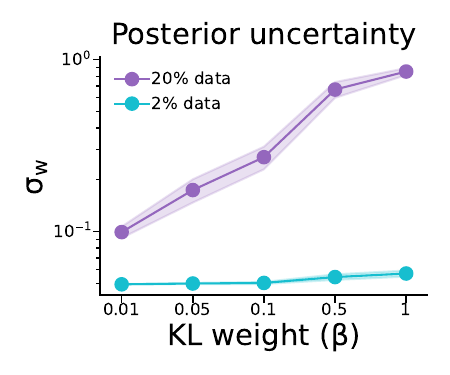}
        \includegraphics[width=0.16\textwidth]{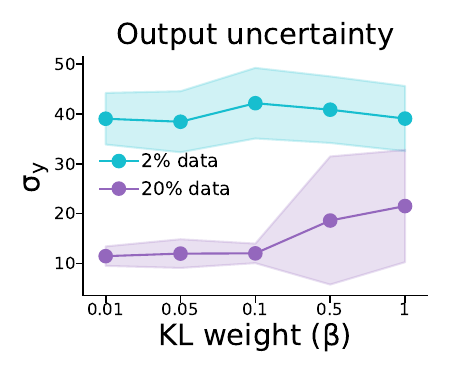}
        \includegraphics[width=0.16\textwidth]{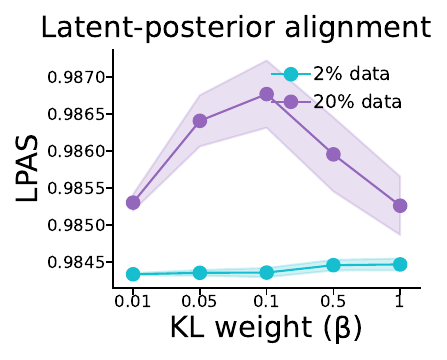}
        \includegraphics[width=0.16\textwidth]{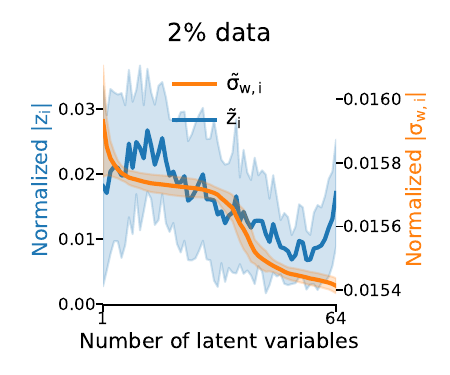}
        \includegraphics[width=0.16\textwidth]{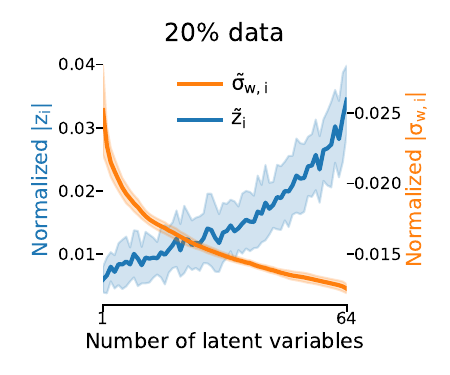}
        \caption{Melting point (mp)} % (b) 
        % \label{si_5_b}
    \end{subfigure}
    % \begin{subfigure}{\textwidth}
    %     \centering
    %     \includegraphics[width=0.16\textwidth]{fig/si/5_kl_test/logP_mae.pdf}
    %     \includegraphics[width=0.16\textwidth]{fig/si/5_kl_test/logP_sigma.pdf}
    %     \includegraphics[width=0.16\textwidth]{fig/si/5_kl_test/logP_pred_sigma.pdf}
    %     \includegraphics[width=0.16\textwidth]{fig/si/5_kl_test/logP_LPAS.pdf}
    %     \includegraphics[width=0.16\textwidth]{fig/si/5_kl_test/logP_02.pdf}
    %     \includegraphics[width=0.16\textwidth]{fig/si/5_kl_test/logP_8.pdf}
    %     \caption{} % (b) 
    %     \label{si_5_c}
    % \end{subfigure}
    \begin{subfigure}{\textwidth}
        \centering
        \includegraphics[width=0.16\textwidth]{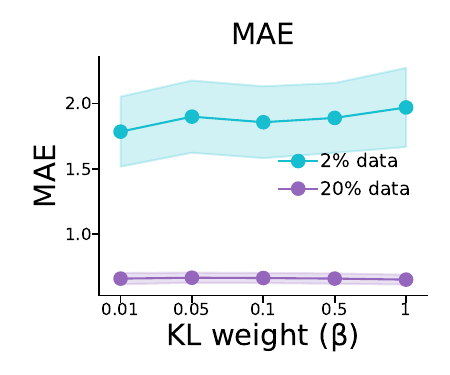}
        \includegraphics[width=0.16\textwidth]{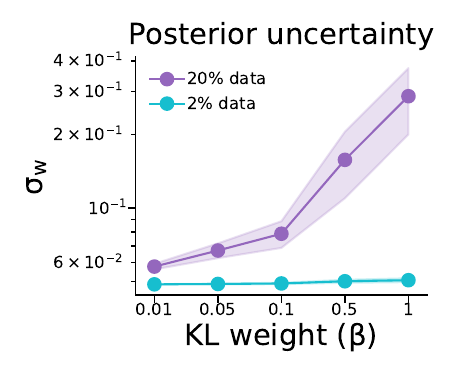}
        \includegraphics[width=0.16\textwidth]{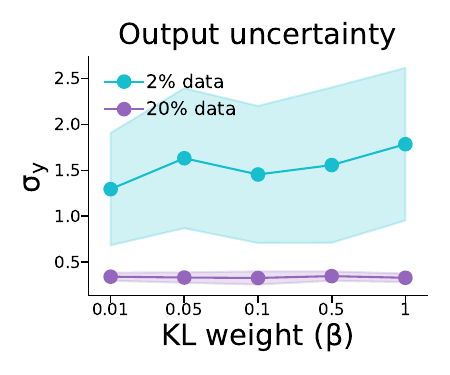}
        \includegraphics[width=0.16\textwidth]{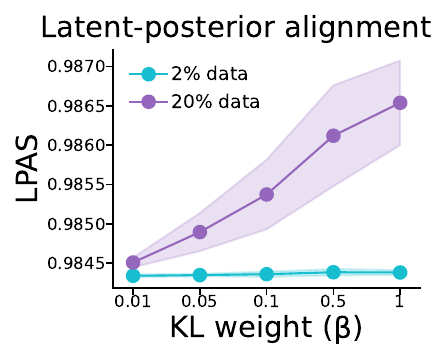}
        \includegraphics[width=0.16\textwidth]{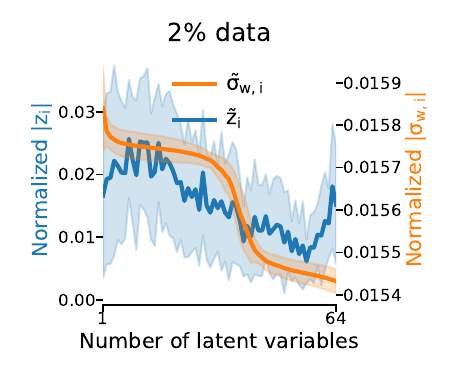}
        \includegraphics[width=0.16\textwidth]{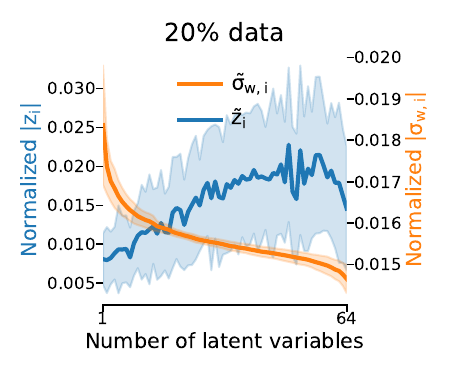}

        \caption{Vapor pressure (Pvap)} % (b) 
        % \label{si_5_d}
    \end{subfigure}
    \begin{subfigure}{\textwidth}
        \centering
        \includegraphics[width=0.16\textwidth]{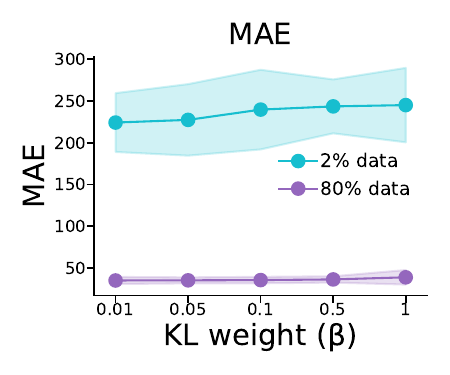}
        \includegraphics[width=0.16\textwidth]{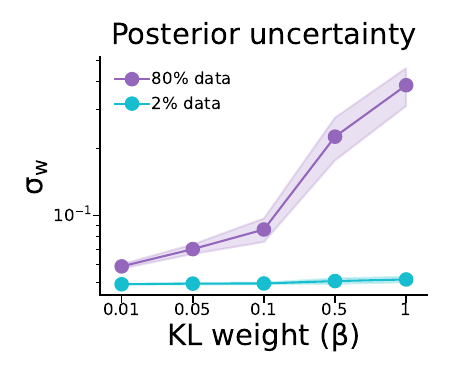}
        \includegraphics[width=0.16\textwidth]{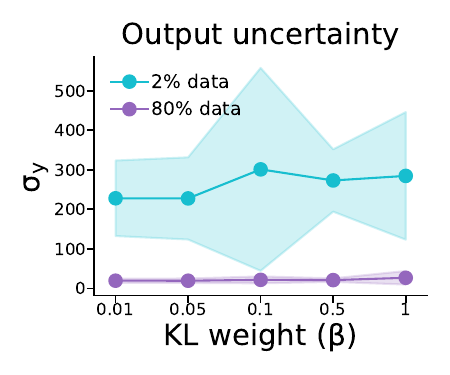}
        \includegraphics[width=0.16\textwidth]{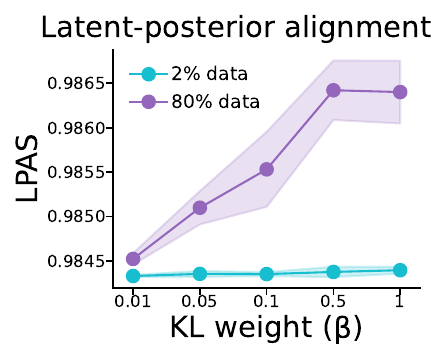}
        \includegraphics[width=0.16\textwidth]{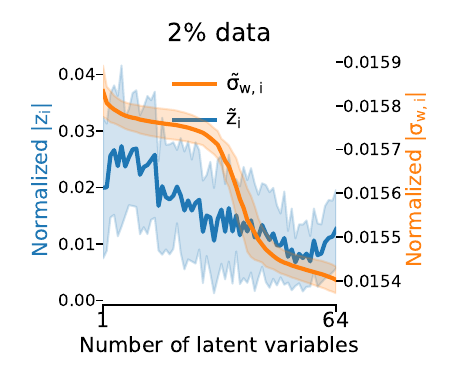}
        \includegraphics[width=0.16\textwidth]{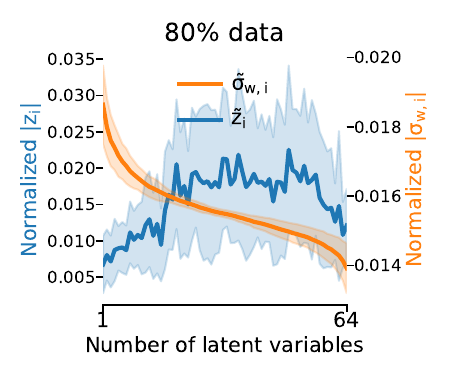}
        \caption{Enthalpy (H)} % (b) 
        % \label{si_5_e}
    \end{subfigure}
    \begin{subfigure}{\textwidth}
        \centering
        \includegraphics[width=0.16\textwidth]{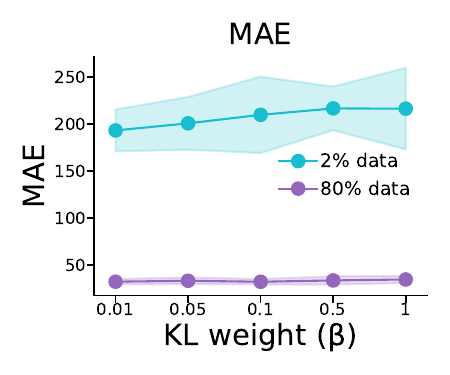}
        \includegraphics[width=0.16\textwidth]{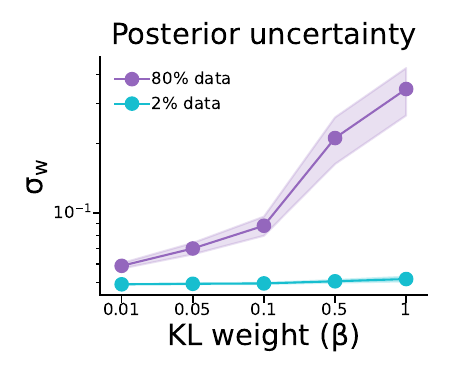}
        \includegraphics[width=0.16\textwidth]{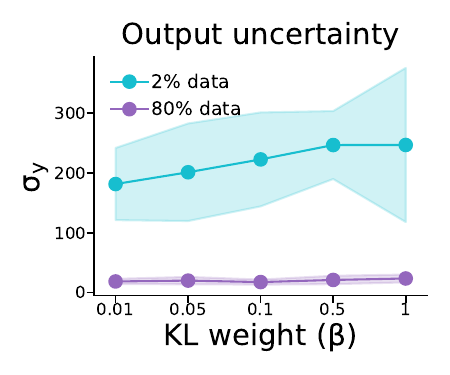}
        \includegraphics[width=0.16\textwidth]{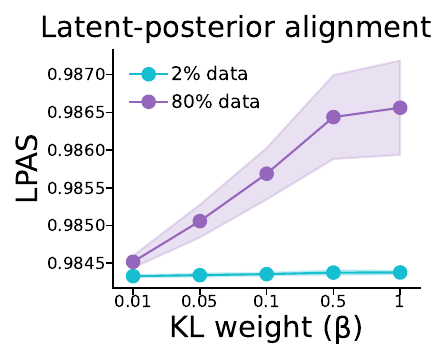}
        \includegraphics[width=0.16\textwidth]{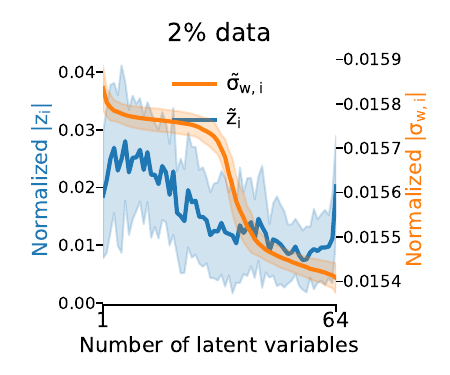}
        \includegraphics[width=0.16\textwidth]{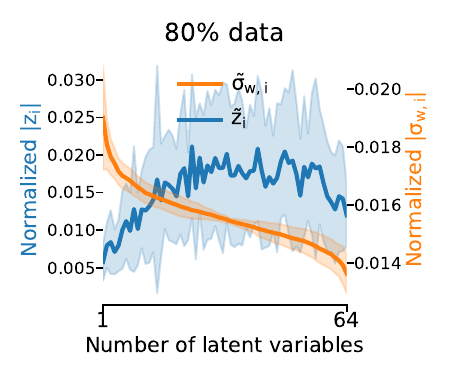}
        \caption{Gibbs free energy (G)} % (b) 
        % \label{si_5_f}
    \end{subfigure}
    
    \caption{Influence of KL-divergence weight on Latent-Posterior Alignment and uncertainty dynamics. Rows (a-e) correspond to five different property datasets. Cyan and purple represent data-poor and data-rich regimes, respectively. Columns 1-4 show the KL-divergence weight impact on MAE, posterior uncertainty ($\sigma_{w}$), predictive uncertainty ($\sigma_{y}$), and LPAS. Columns 5-6 visualize the geometry of latent vectors and posterior standard deviations for data-poor (Column 5) and data-rich (Column 6) models.}
    \label{si_fig_kl_test}
\end{figure}

To examine how robust Latent-Posterior Alignment (LPA) is to posterior regularization, we extended the KL-divergence weight ($\beta$) analysis to five additional physicochemical property datasets (Fig. \ref{si_fig_kl_test}). As expected, lowering $\beta$ weakens the regularization pressure, leading to a notable reduction in posterior uncertainty (Column 2). This effect is regime-dependent: it is more pronounced in the data-rich regime (purple), where baseline variance is larger, compared to the data-poor regime (cyan).

Crucially, however, unlike posterior uncertainty, both predictive accuracy (MAE, Column 1) and output uncertainty (Column 3) remain largely invariant to changes in $\beta$. Specifically, data-rich models maintain their characteristic low-uncertainty profile compared to data-poor models, regardless of whether their posteriors are tight (low $\beta$) or loose (high $\beta$). This invariance strongly suggests that posterior uncertainty itself is not the primary driver of predictive confidence. Instead, the alignment mechanism consistently governs the uncertainty dynamics. The Latent-Posterior Alignment Score (LPAS, Column 4) reveals a fundamental distinction: in the data-poor case, LPAS values remain consistently low and invariant to $\beta$, corresponding to a lack of LPA structure in the latent dimension (Column 5). Conversely, in the data-rich case, the models robustly maintain higher LPAS values, preserving a strong LPA structure (Column 6). These results demonstrate that Latent-Posterior Alignment serves as a robust and dominant strategy for controlling uncertainty, operating independently of the posterior's flexibility.

\clearpage

\subsection{Effect of prior variance on LPA} \label{si_prior_test}
\begin{figure}[H]
    \centering
    \begin{subfigure}{\textwidth}
        \centering
        \includegraphics[width=0.16\textwidth]{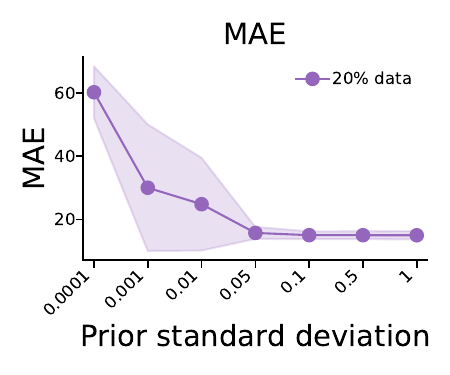}
        \includegraphics[width=0.16\textwidth]{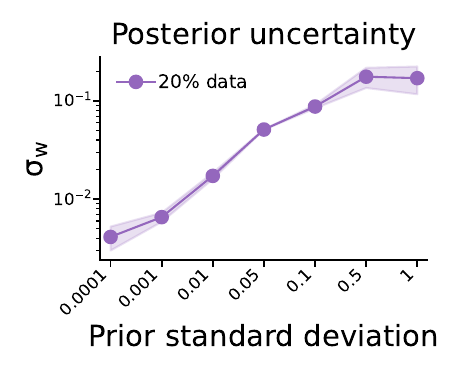}
        \includegraphics[width=0.16\textwidth]{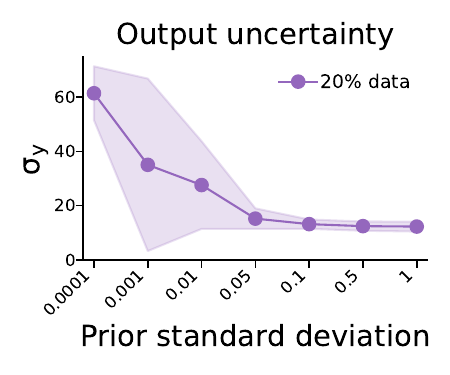}
        \includegraphics[width=0.16\textwidth]{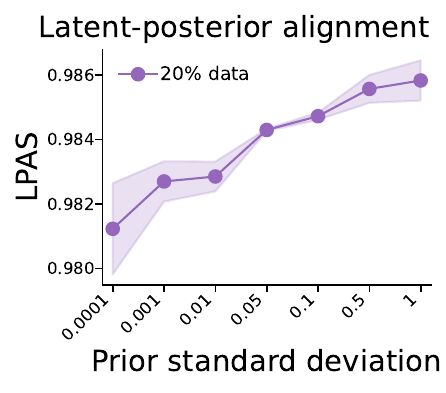}
        \includegraphics[width=0.16\textwidth]{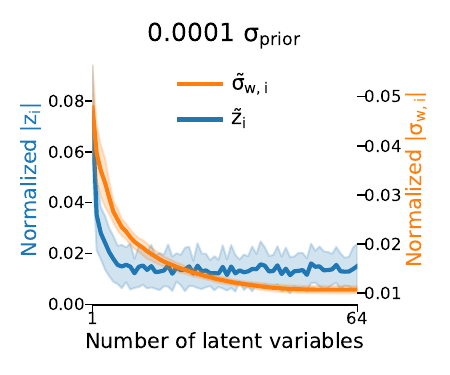}
        \includegraphics[width=0.16\textwidth]{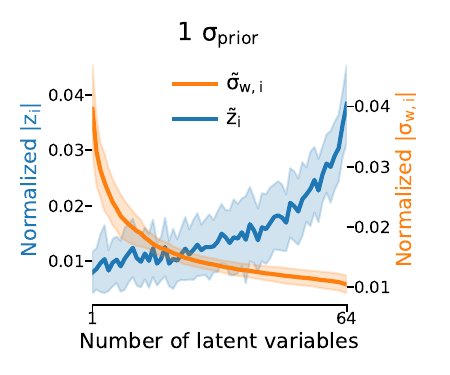}
        \caption{Boiling point (bp)} % (b) 
        % \label{si_5_a}
    \end{subfigure}
    \begin{subfigure}{\textwidth}
        \centering
        \includegraphics[width=0.16\textwidth]{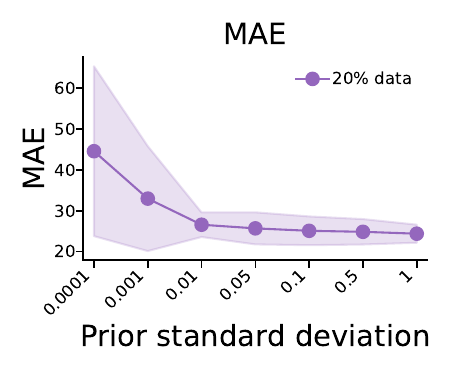}
        \includegraphics[width=0.16\textwidth]{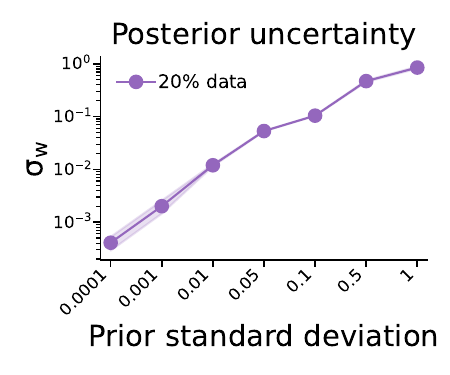}
        \includegraphics[width=0.16\textwidth]{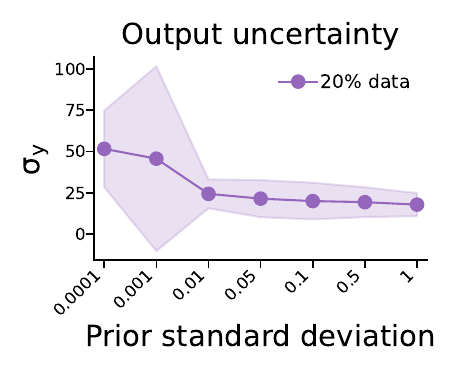}
        \includegraphics[width=0.16\textwidth]{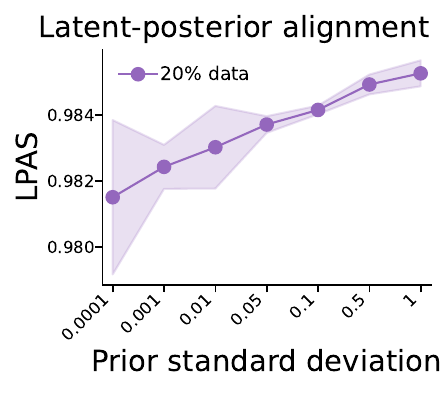}
        \includegraphics[width=0.16\textwidth]{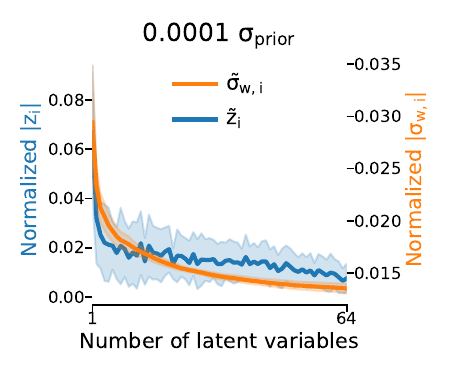}
        \includegraphics[width=0.16\textwidth]{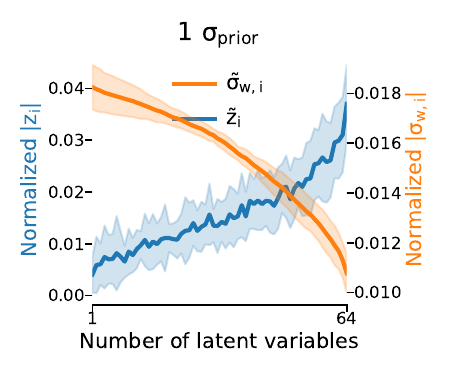}
        \caption{Melting point (mp)} % (b) 
        % \label{si_5_b}
    \end{subfigure}
    % \begin{subfigure}{\textwidth}
    %     \centering
    %     \includegraphics[width=0.16\textwidth]{fig/si/6_prior_test/logP_mae.pdf}
    %     \includegraphics[width=0.16\textwidth]{fig/si/6_prior_test/logP_sigma.pdf}
    %     \includegraphics[width=0.16\textwidth]{fig/si/6_prior_test/logP_pred_sigma.pdf}
    %     \includegraphics[width=0.16\textwidth]{fig/si/6_prior_test/logP_LPAS.pdf}
    %     \includegraphics[width=0.16\textwidth]{fig/si/6_prior_test/logP_0.0001.pdf}
    %     \includegraphics[width=0.16\textwidth]{fig/si/6_prior_test/logP_1.pdf}
    %     \caption{} % (b) 
    %     \label{si_5_c}
    % \end{subfigure}
    \begin{subfigure}{\textwidth}
        \centering
        \includegraphics[width=0.16\textwidth]{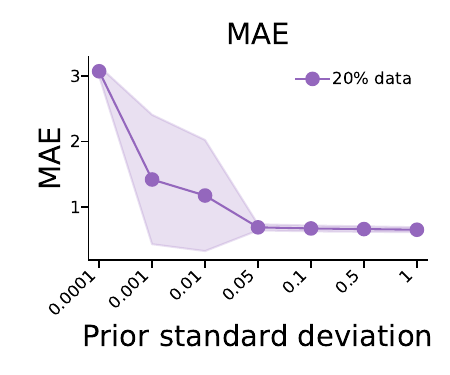}
        \includegraphics[width=0.16\textwidth]{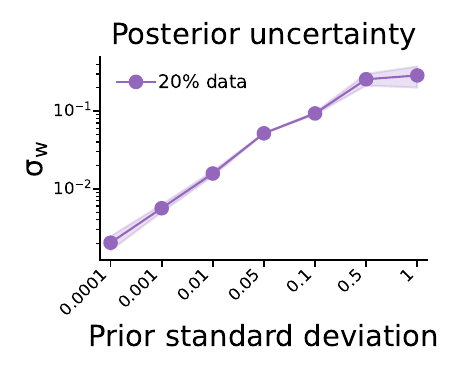}
        \includegraphics[width=0.16\textwidth]{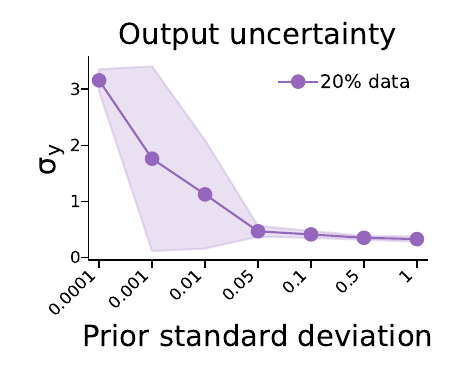}
        \includegraphics[width=0.16\textwidth]{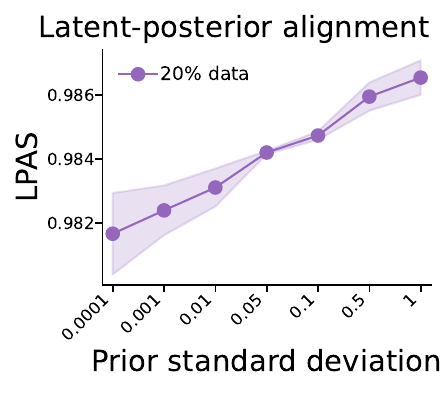}
        \includegraphics[width=0.16\textwidth]{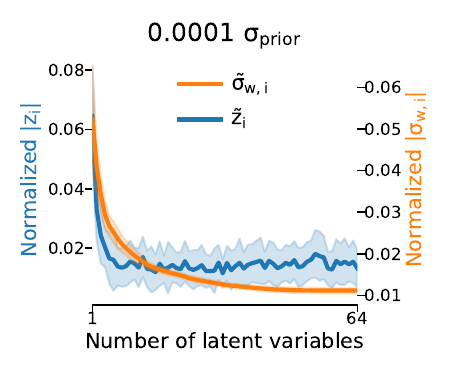}
        \includegraphics[width=0.16\textwidth]{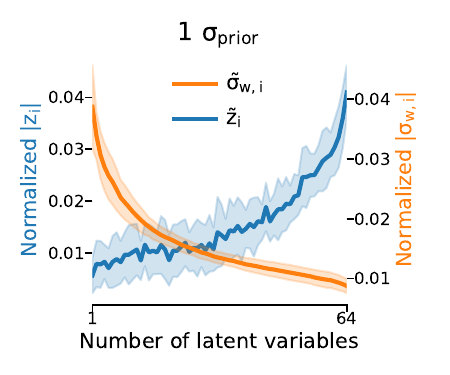}

        \caption{Vapor pressure (Pvap)} % (b) 
        % \label{si_5_d}
    \end{subfigure}
    \begin{subfigure}{\textwidth}
        \centering
        \includegraphics[width=0.16\textwidth]{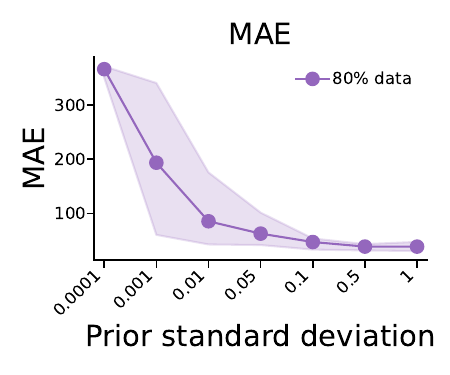}
        \includegraphics[width=0.16\textwidth]{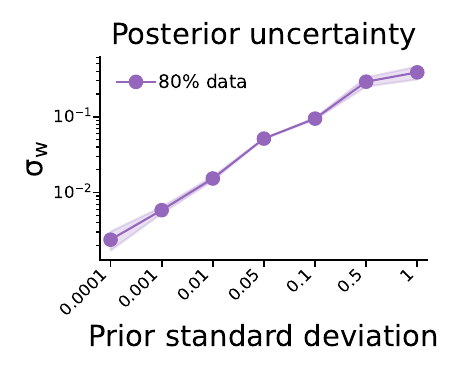}
        \includegraphics[width=0.16\textwidth]{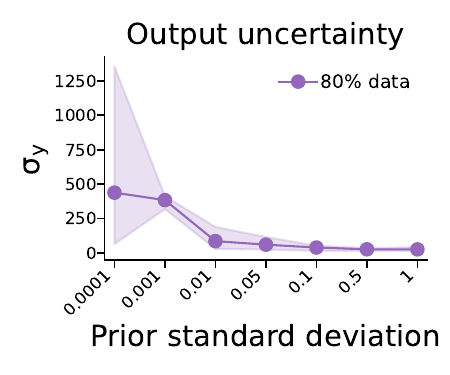}
        \includegraphics[width=0.16\textwidth]{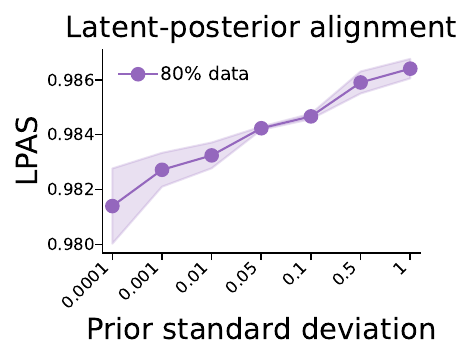}
        \includegraphics[width=0.16\textwidth]{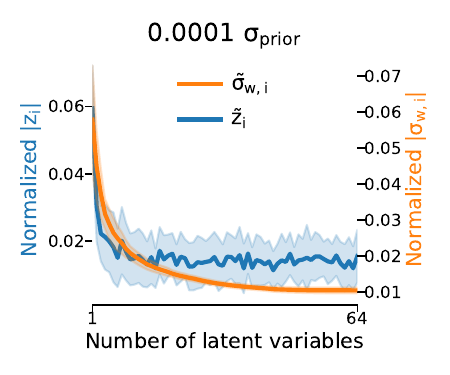}
        \includegraphics[width=0.16\textwidth]{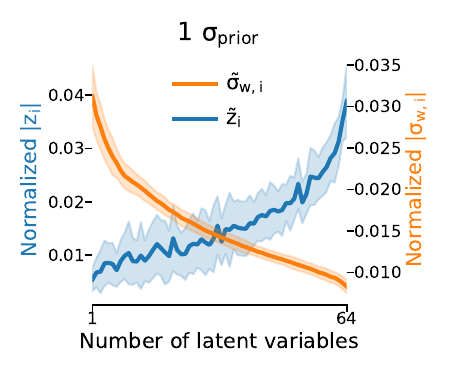}
        \caption{Enthalpy (H)} % (b) 
        % \label{si_5_e}
    \end{subfigure}
    \begin{subfigure}{\textwidth}
        \centering
        \includegraphics[width=0.16\textwidth]{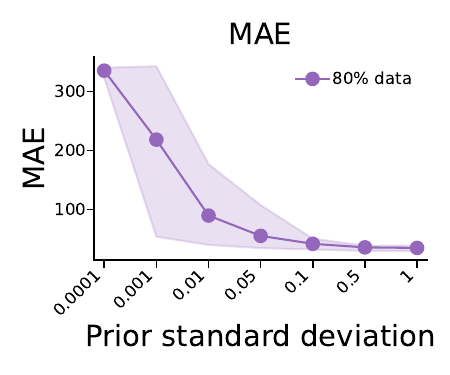}
        \includegraphics[width=0.16\textwidth]{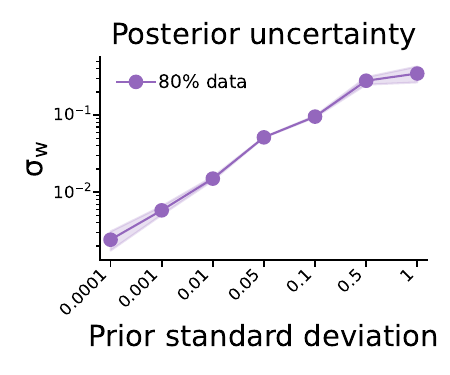}
        \includegraphics[width=0.16\textwidth]{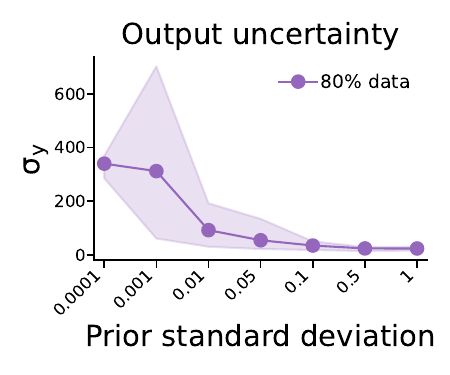}
        \includegraphics[width=0.16\textwidth]{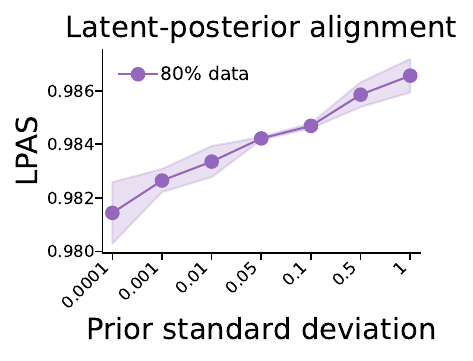}
        \includegraphics[width=0.16\textwidth]{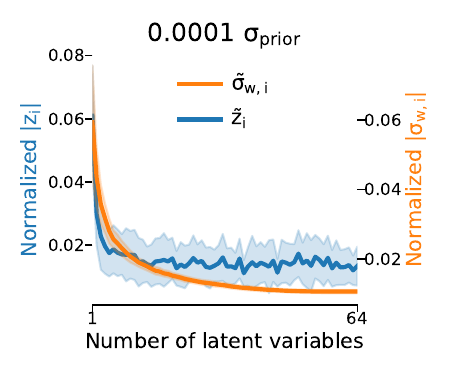}
        \includegraphics[width=0.16\textwidth]{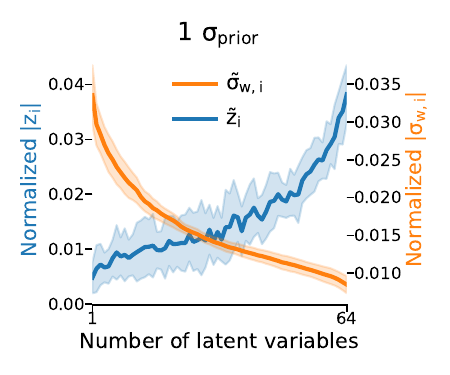}
        \caption{Gibbs free energy (G)} % (b) 
        % \label{si_5_f}
    \end{subfigure}
    
    \caption{Impact of prior distribution on Latent-Posterior Alignment and predictive uncertainty. Analysis is conducted on the data-rich regime (80\% train data) across five physicochemical properties (a-e). Columns 1-4 show the evolution of MAE, posterior uncertainty ($\sigma_{w}$), predictive uncertainty ($\sigma_{y}$), and LPAS as a function of prior standard deviation. Columns 5-6 visualize latent vectors under restrictive prior (Column 5, $\sigma_{prior}=0.0001$) and relaxed prior (Column 6, $\sigma_{prior}=1$).}
    \label{si_fig_prior_test}
\end{figure}

We extended the prior analysis to five property datasets (Fig. \ref{si_fig_prior_test}) to validate the robustness of Latent-Posterior Alignment (LPA) against prior variations. We focused on the data-rich regime (80\% training data) where LPA is active. First, imposing a narrower prior successfully forced the posterior distributions to concentrate (Column 2). However, this contraction followed an inverse trajectory relative to model performance. Both predictive error (MAE, Column 1) and output uncertainty (Column 3) gradually increased as the prior became more restrictive. This paradox reconfirms that the magnitude of posterior variance is not the governing factor for predictive uncertainty.

The deterioration in performance is directly attributable to the disruption of the LPA mechanism. As the prior standard deviation decreases, LPAS decreases (Column 4), indicating weaker alignment under increasingly restrictive priors. Visualizing latent geometry reveals that under strict prior constraints (Column 5), latent vectors are forced into unstable, high-variance dimensions, breaking the alignment structure observed under relaxed priors (Column 6). Furthermore, consistent with the main analysis, we identified a critical failure point across all datasets. While the specific threshold varies by task, excessive suppression of prior variance universally leads to a collapse of the LPA mechanism, resulting in a surge in model error and uncertainty. This finding demonstrates that a sufficient degree of variance in the parameter space—exploratory variance—is indispensable for the model to discover and maintain effective alignment pathways.

\clearpage

\section{Cross-dataset evaluation of AGL and density awareness} \label{si_agl}
\begin{figure}[H]
    \centering
    \begin{subfigure}{0.7\textwidth}
        \centering
        \includegraphics[width=\textwidth]{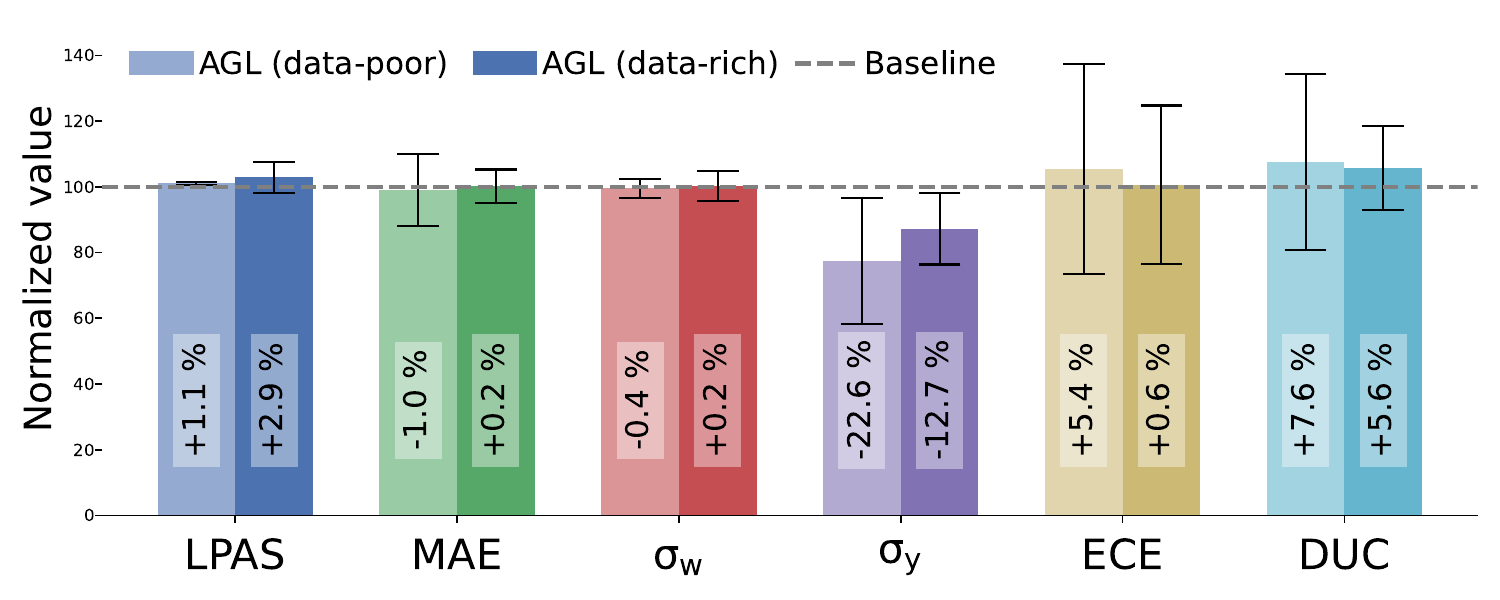}
        \caption{Boiling point (bp)} % (b) 
        \label{si_7_a}
    \end{subfigure}
    \begin{subfigure}{0.7\textwidth}
        \centering
        \includegraphics[width=\textwidth]{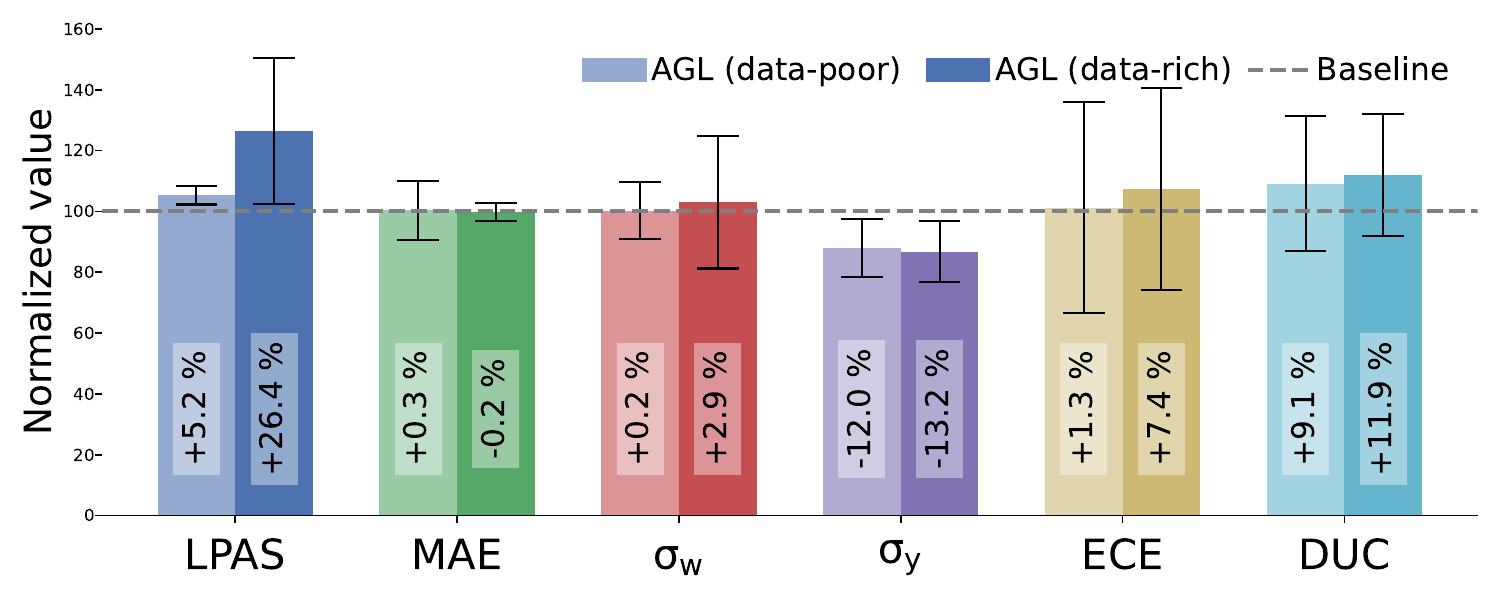}
        \caption{Melting point (mp)} % (b) 
        \label{si_7_b}
    \end{subfigure}
    % \begin{subfigure}{0.48\textwidth}
    %     \centering
    %     \includegraphics[width=\textwidth]{fig/si/7_alignment/logP.pdf}
    %     \caption{} % (b) 
    %     \label{si_7_c}
    % \end{subfigure}
    \begin{subfigure}{0.7\textwidth}
        \centering
        \includegraphics[width=\textwidth]{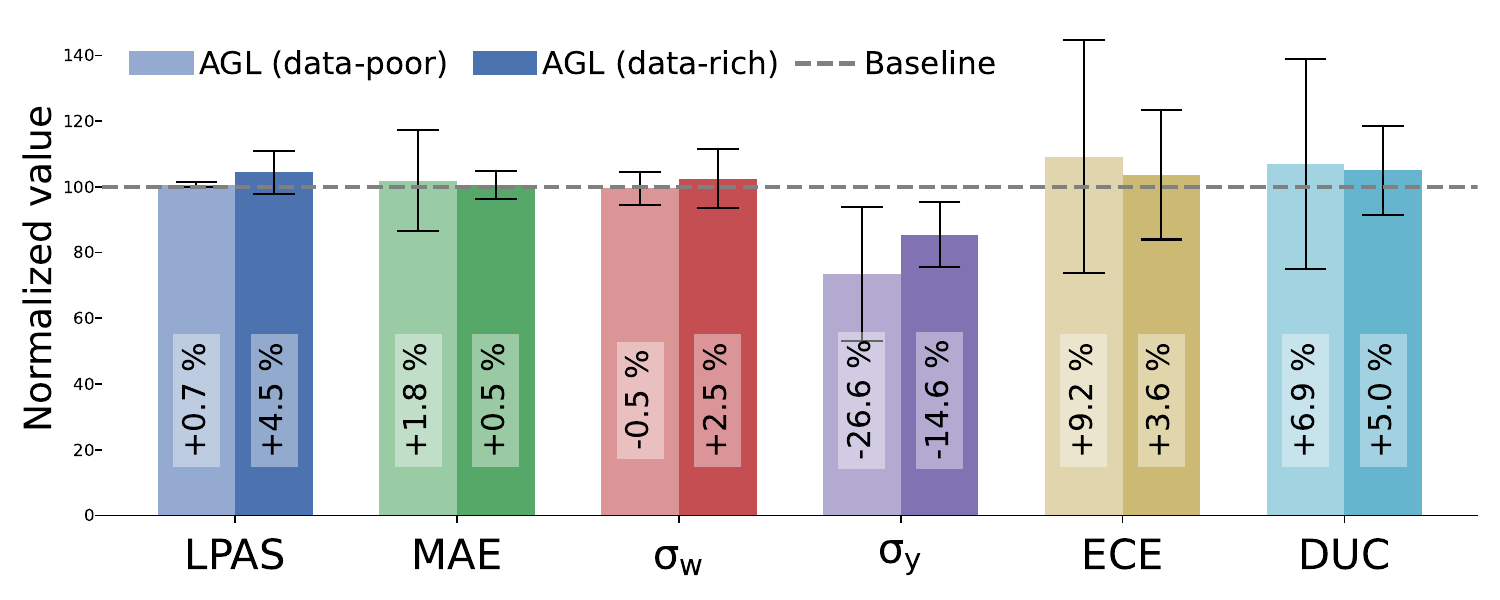}
        \caption{Vapor pressure (Pvap)} % (b) 
        \label{si_7_d}
    \end{subfigure}
    \begin{subfigure}{0.7\textwidth}
        \centering
        \includegraphics[width=\textwidth]{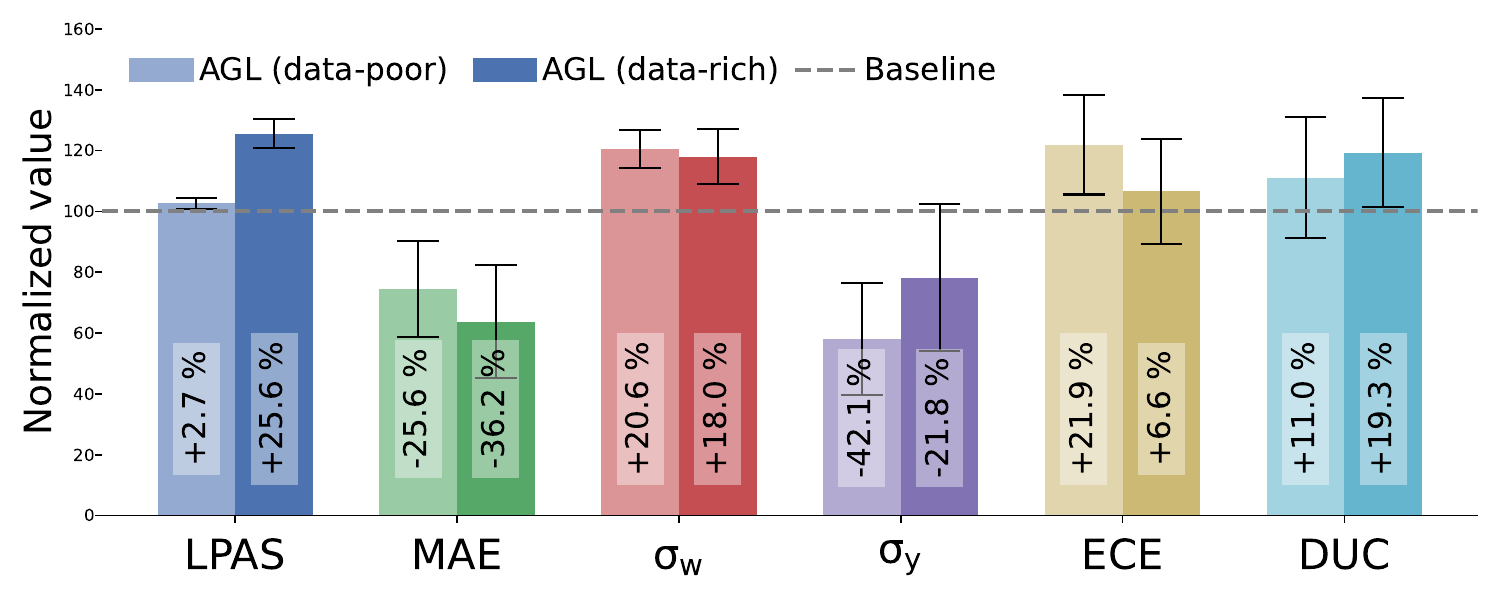}
        \caption{Enthalpy (H)} % (b) 
        \label{si_7_e}
    \end{subfigure}
    \begin{subfigure}{0.7\textwidth}
        \centering
        \includegraphics[width=\textwidth]{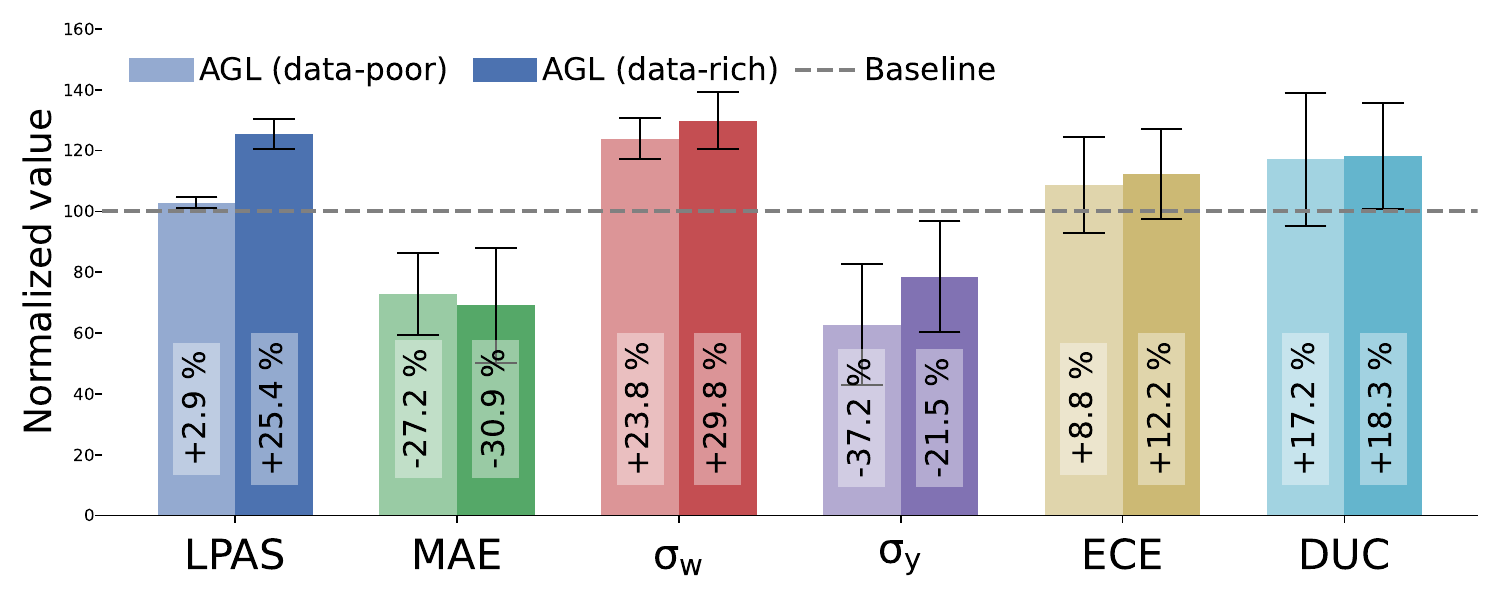}
        \caption{Gibbs free energy (G)} % (b) 
        \label{si_7_f}
    \end{subfigure}
    
    \caption{Performance evaluation of Alignment-Guided Learning (AGL) across diverse datasets. Results are shown for five different properties (a-e). The AGL model (colored bars) is compared with the baseline (gray dashed line) for six metrics, with values normalized relative to the baseline scaled to 100. Light and dark shaded bars represent the data-poor and data-rich regimes, respectively.}
    \label{si_fig_agl}
\end{figure}

\begin{figure}[H]
    \centering

    \begin{subfigure}{0.48\textwidth}
        \centering
        \includegraphics[width=0.46\linewidth]{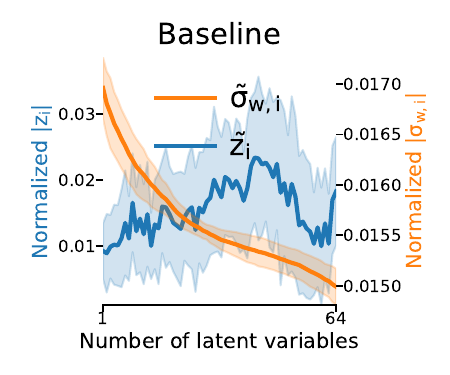}
        \hfill
        \raisebox{3em}{$\scriptstyle \rightarrow$}
        \hfill
        \includegraphics[width=0.46\linewidth]{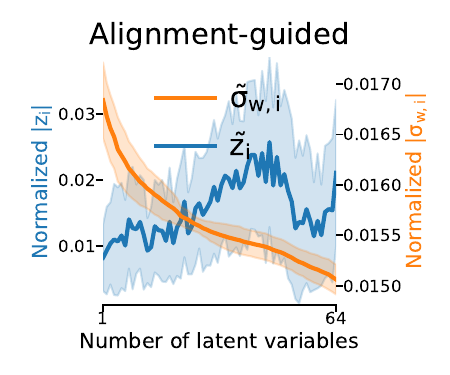}
        \rule{\linewidth}{0.4pt}
        \caption{Boiling point (bp)} %
    \end{subfigure}
    \begin{subfigure}{0.48\textwidth}
        \centering
        \includegraphics[width=0.46\linewidth]{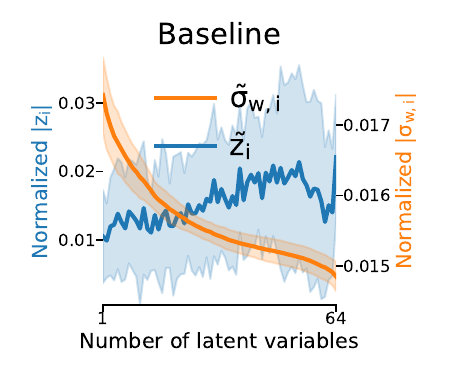}
        \hfill
        \raisebox{3em}{$\scriptstyle \rightarrow$}
        \hfill
        \includegraphics[width=0.46\linewidth]{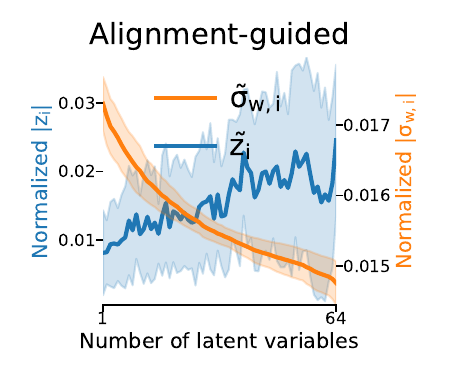}
        \rule{\linewidth}{0.4pt}
        \caption{Melting point (mp)} %
    \end{subfigure}

    \begin{subfigure}{0.48\textwidth}
        \centering
        \includegraphics[width=0.46\linewidth]{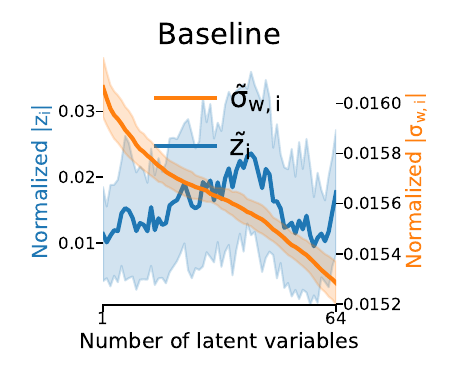}
        \hfill
        \raisebox{3em}{$\scriptstyle \rightarrow$}
        \hfill
        \includegraphics[width=0.46\linewidth]{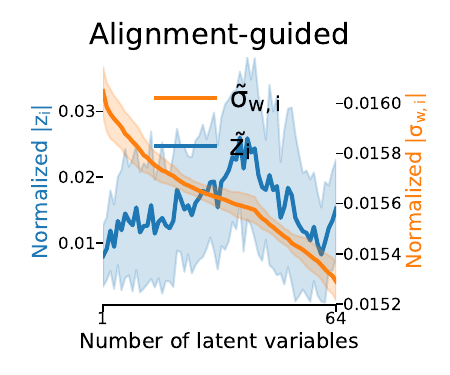}
        \rule{\linewidth}{0.4pt}
        \caption{Vapor pressure (Pvap)} %
    \end{subfigure}
    \begin{subfigure}{0.48\textwidth}
        \centering
        \includegraphics[width=0.46\linewidth]{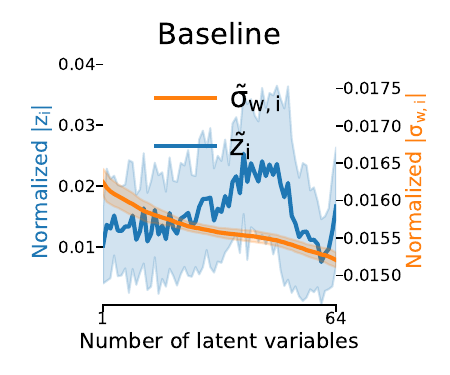}
        \hfill
        \raisebox{3em}{$\scriptstyle \rightarrow$}
        \hfill
        \includegraphics[width=0.46\linewidth]{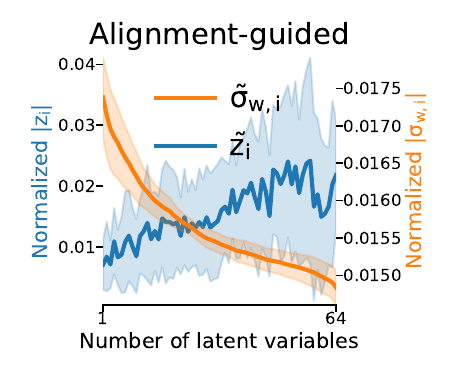}
        \rule{\linewidth}{0.4pt}
        \caption{Enthalpy (H)} %
    \end{subfigure}

    \begin{subfigure}{0.48\textwidth}
        \centering
        \includegraphics[width=0.46\linewidth]{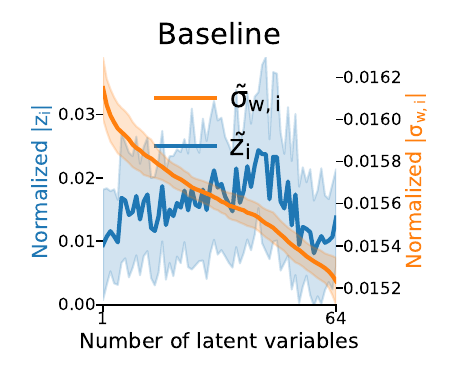}
        \hfill
        \raisebox{3em}{$\scriptstyle \rightarrow$}
        \hfill
        \includegraphics[width=0.46\linewidth]{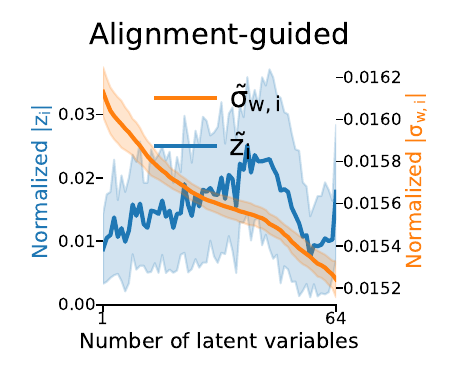}
        \rule{\linewidth}{0.4pt}
        \caption{Gibbs free energy (G)} %
    \end{subfigure}
        
    \caption{Impact of Alignment-Guided Learning (AGL) on latent geometry in the data-poor regime. Visualizations of normalized latent vector magnitudes ($\tilde{\mathbf{z}}$, blue line) and posterior standard deviations ($\tilde{\sigma}_{w}$, orange line) across five physicochemical datasets (a–e). Each panel compares the unconstrained baseline model (left) with the AGL model (right). The arrows indicate the structural shift induced by the alignment objective. In this data-scarce regime, while AGL induces a tendency toward the alignment, the resulting structural reorganization is not clearly visible (or remains visually subtle) due to severe information scarcity. Shaded areas represent the standard deviation across $n=30$ independent runs.}
    \label{si_fig_align_latent_poor}
\end{figure}

\begin{figure}[H]
    \centering

    \begin{subfigure}{0.48\textwidth}
        \centering
        \includegraphics[width=0.46\linewidth]{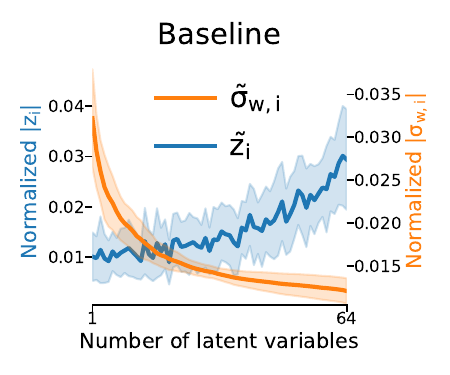}
        \hfill
        \raisebox{3em}{$\scriptstyle \rightarrow$}
        \hfill
        \includegraphics[width=0.46\linewidth]{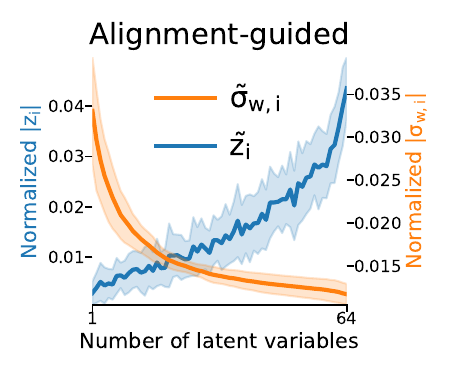}
        \rule{\linewidth}{0.4pt}
        \caption{Boiling point (bp)} %
    \end{subfigure}
    \begin{subfigure}{0.48\textwidth}
        \centering
        \includegraphics[width=0.46\linewidth]{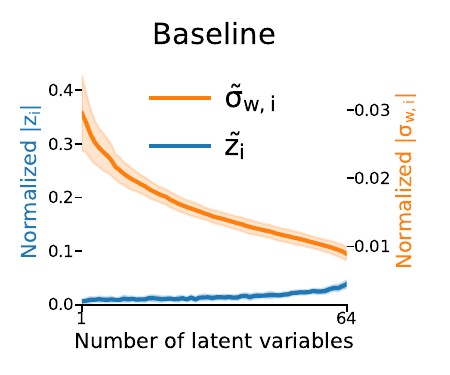}
        \hfill
        \raisebox{3em}{$\scriptstyle \rightarrow$}
        \hfill
        \includegraphics[width=0.46\linewidth]{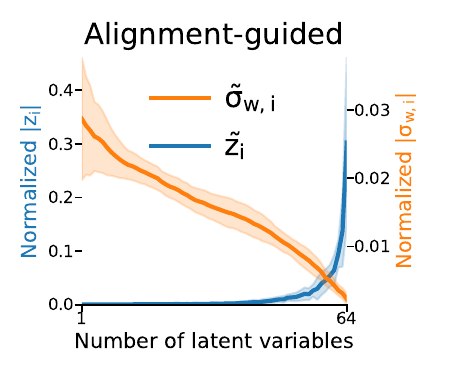}
        \rule{\linewidth}{0.4pt}
        \caption{Melting point (mp)} %
    \end{subfigure}

    \begin{subfigure}{0.48\textwidth}
        \centering
        \includegraphics[width=0.46\linewidth]{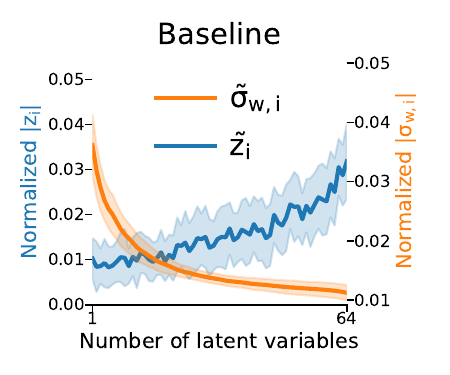}
        \hfill
        \raisebox{3em}{$\scriptstyle \rightarrow$}
        \hfill
        \includegraphics[width=0.46\linewidth]{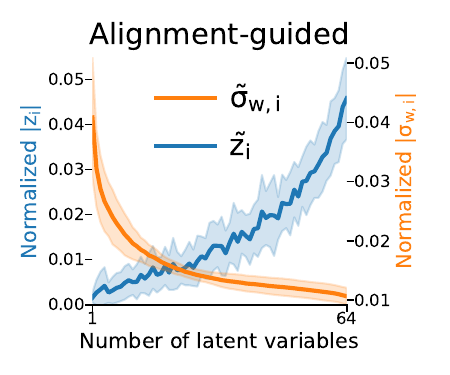}
        \rule{\linewidth}{0.4pt}
        \caption{Vapor pressure (Pvap)} %
    \end{subfigure}
    \begin{subfigure}{0.48\textwidth}
        \centering
        \includegraphics[width=0.46\linewidth]{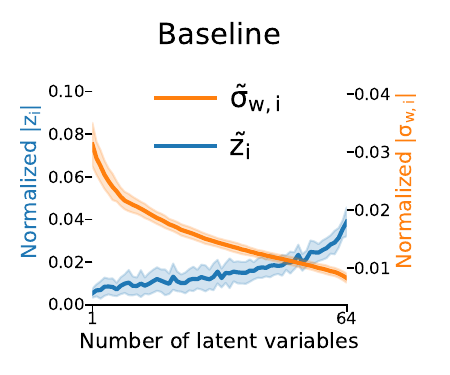}
        \hfill
        \raisebox{3em}{$\scriptstyle \rightarrow$}
        \hfill
        \includegraphics[width=0.46\linewidth]{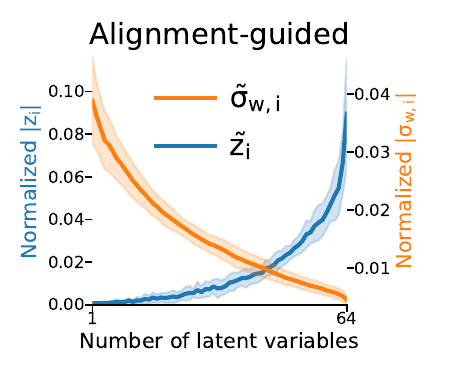}
        \rule{\linewidth}{0.4pt}
        \caption{Enthalpy (H)} %
    \end{subfigure}

    \begin{subfigure}{0.48\textwidth}
        \centering
        \includegraphics[width=0.46\linewidth]{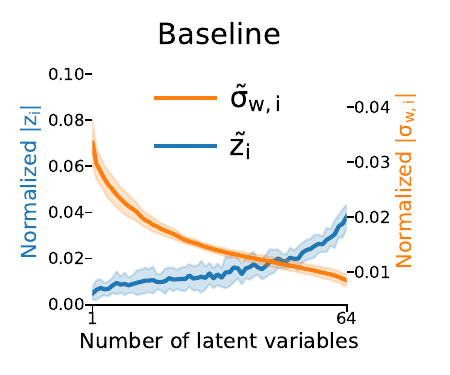}
        \hfill
        \raisebox{3em}{$\scriptstyle \rightarrow$}
        \hfill
        \includegraphics[width=0.46\linewidth]{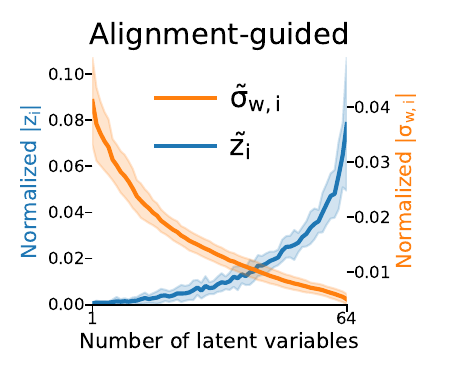}
        \rule{\linewidth}{0.4pt}
        \caption{Gibbs free energy (G)} %
    \end{subfigure}
        
    \caption{Impact of Latent-Posterior Alignment via AGL in the data-rich regime. Comparison of latent geometry between Baseline (left) and AGL (right) models trained in a data-abundant setting across five datasets (a–e). Consistent with the data-poor regime, AGL enforces alignment; however, the extent of this reorganization is more pronounced. With sufficient data, AGL actively reconfigures the latent space, creating a sharp boundary where latent activity is strictly confined to stable, low-variance axes (forming a distinct ``X'' shape). This confirms that the availability of information facilitates more precise geometric calibration. Shaded areas denote standard deviation ($n=30$).}
    \label{si_fig_align_latent_rich}
\end{figure}

\begin{figure}[H]
    \centering
    \begin{center}
    \begin{tabular}{@{}m{0.46\textwidth} @{}m{0.04\textwidth} @{}m{0.46\textwidth}@{}}
    \centering
    \begin{subfigure}{\linewidth}
        \centering
        \includegraphics[width=0.48\linewidth]{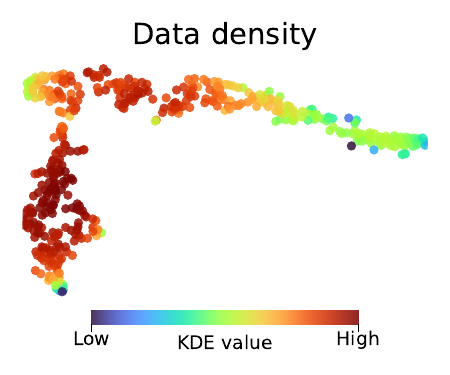}
        \includegraphics[width=0.48\linewidth]{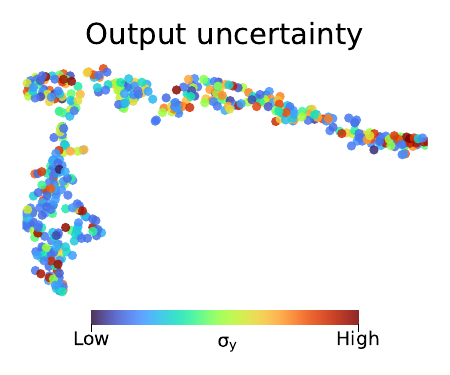}
        \rule{\linewidth}{0.4pt}
        \caption{bp: Baseline } 
    \end{subfigure}
    &
    \centering{\raisebox{2.4em}{$\rightarrow$}}
    &
    \centering
    \begin{subfigure}{\linewidth}
        \centering
        \includegraphics[width=0.48\linewidth]{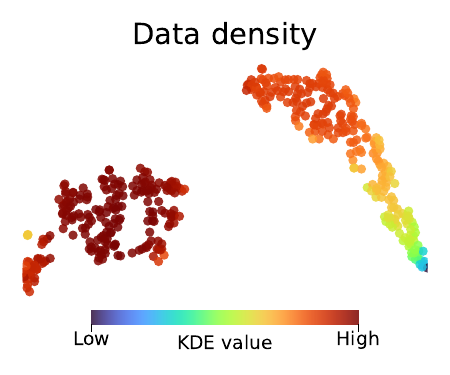}
        \includegraphics[width=0.48\linewidth]{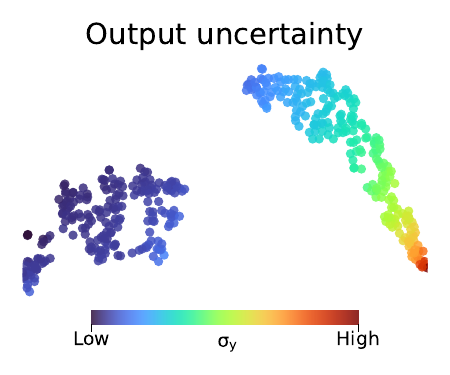}
        \rule{\linewidth}{0.4pt}
        \caption{bp: AGL}  
    \end{subfigure}
    \end{tabular}
    \end{center}

    \begin{center}
    \begin{tabular}{@{}m{0.46\textwidth} @{}m{0.04\textwidth} @{}m{0.46\textwidth}@{}}
    \centering
    \begin{subfigure}{\linewidth}
        \centering
        \includegraphics[width=0.48\linewidth]{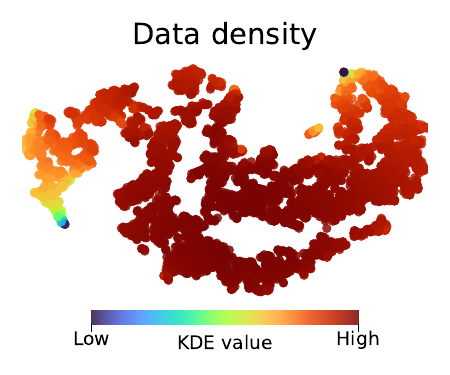}
        \includegraphics[width=0.48\linewidth]{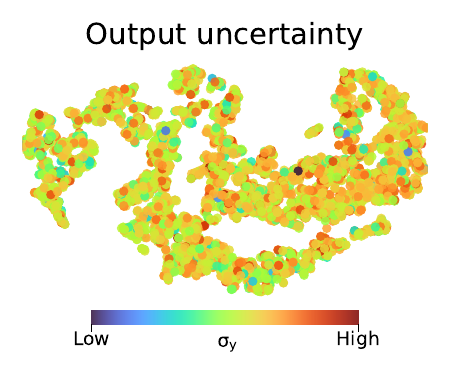}
        \rule{\linewidth}{0.4pt}
        \caption{mp: Baseline} 
    \end{subfigure}
    &
    \centering{\raisebox{2.4em}{$\rightarrow$}}
    &
    \centering
    \begin{subfigure}{\linewidth}
        \centering
        \includegraphics[width=0.48\linewidth]{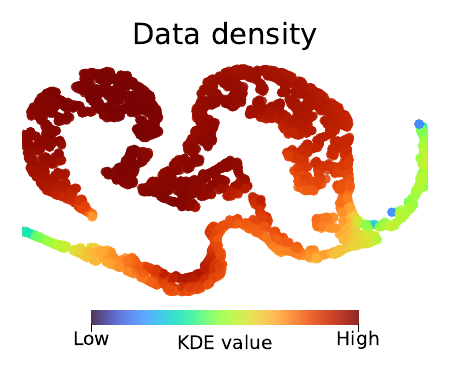}
        \includegraphics[width=0.48\linewidth]{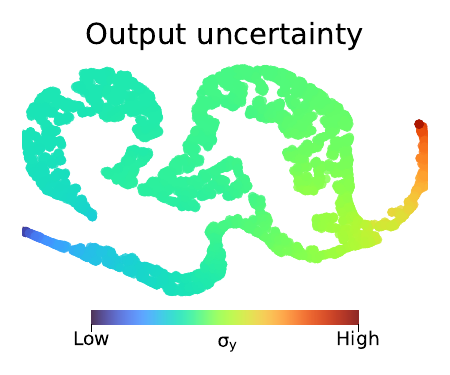}
        \rule{\linewidth}{0.4pt}
        \caption{mp: AGL}  
    \end{subfigure}
    \end{tabular}
    \end{center}

    % \begin{center}
    % \begin{tabular}{@{}m{0.46\textwidth} @{}m{0.04\textwidth} @{}m{0.46\textwidth}@{}}
    % \centering
    % \begin{subfigure}{\linewidth}
    %     \centering
    %     \includegraphics[width=0.48\linewidth]{fig/si/8_latent_tsne/logP_base_poor_density.pdf}
    %     \includegraphics[width=0.48\linewidth]{fig/si/8_latent_tsne/logP_base_poor_uncertainty.pdf}
    %     \rule{\linewidth}{0.4pt}
    %     \caption{Baseline} 
    % \end{subfigure}
    % &
    % \centering{\raisebox{2.4em}{$\rightarrow$}}
    % &
    % \centering
    % \begin{subfigure}{\linewidth}
    %     \centering
    %     \includegraphics[width=0.48\linewidth]{fig/si/8_latent_tsne/logP_align_poor_density.pdf}
    %     \includegraphics[width=0.48\linewidth]{fig/si/8_latent_tsne/logP_align_poor_uncertainty.pdf}
    %     \rule{\linewidth}{0.4pt}
    %     \caption{Alignment-guided}  
    % \end{subfigure}
    % \end{tabular}
    % \end{center}

    \begin{center}
    \begin{tabular}{@{}m{0.46\textwidth} @{}m{0.04\textwidth} @{}m{0.46\textwidth}@{}}
    \centering
    \begin{subfigure}{\linewidth}
        \centering
        \includegraphics[width=0.48\linewidth]{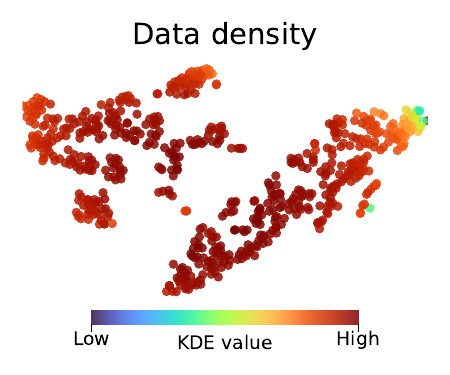}
        \includegraphics[width=0.48\linewidth]{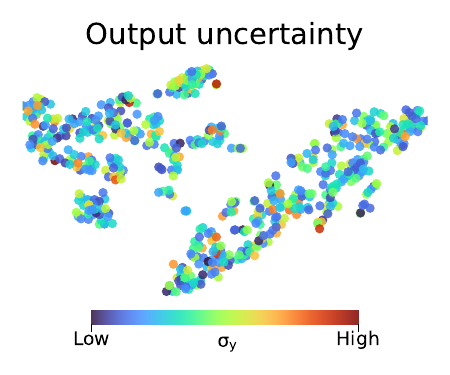}
        \rule{\linewidth}{0.4pt}
        \caption{Pvap: Baseline} 
    \end{subfigure}
    &
    \centering{\raisebox{2.4em}{$\rightarrow$}}
    &
    \centering
    \begin{subfigure}{\linewidth}
        \centering
        \includegraphics[width=0.48\linewidth]{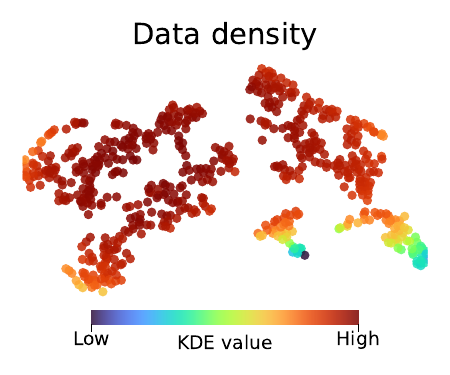}
        \includegraphics[width=0.48\linewidth]{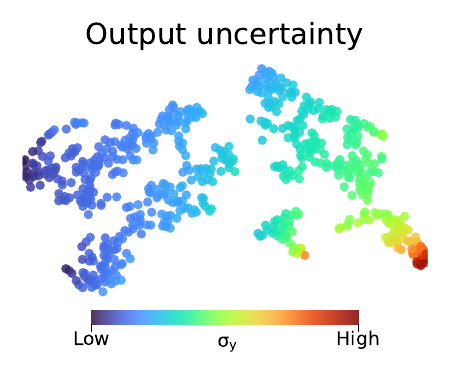}
        \rule{\linewidth}{0.4pt}
        \caption{Pvap: AGL}  
    \end{subfigure}
    \end{tabular}
    \end{center}

    \begin{center}
    \begin{tabular}{@{}m{0.46\textwidth} @{}m{0.04\textwidth} @{}m{0.46\textwidth}@{}}
    \centering
    \begin{subfigure}{\linewidth}
        \centering
        \includegraphics[width=0.48\linewidth]{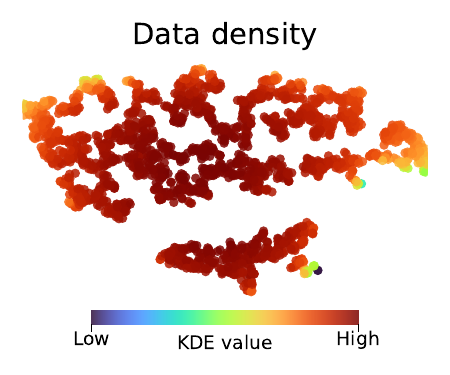}
        \includegraphics[width=0.48\linewidth]{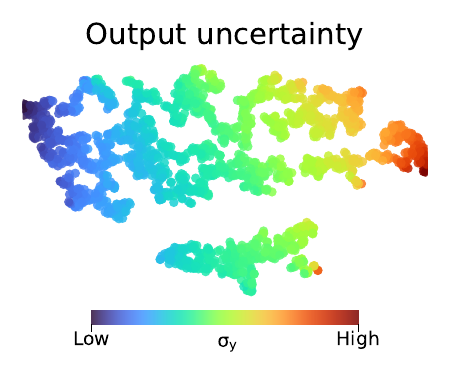}
        \rule{\linewidth}{0.4pt}
        \caption{H: Baseline} 
    \end{subfigure}
    &
    \centering{\raisebox{2.4em}{$\rightarrow$}}
    &
    \centering
    \begin{subfigure}{\linewidth}
        \centering
        \includegraphics[width=0.48\linewidth]{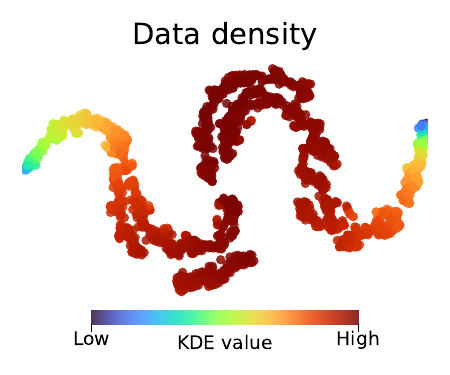}
        \includegraphics[width=0.48\linewidth]{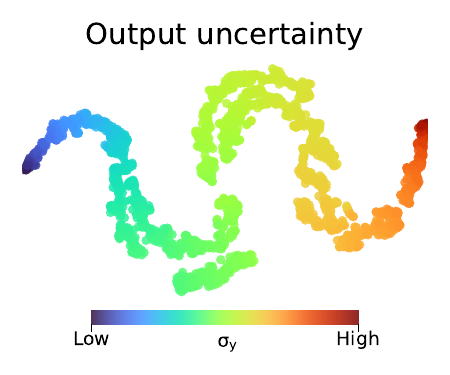}
        \rule{\linewidth}{0.4pt}
        \caption{H: AGL}  
    \end{subfigure}
    \end{tabular}
    \end{center}

    \begin{center}
    \begin{tabular}{@{}m{0.46\textwidth} @{}m{0.04\textwidth} @{}m{0.46\textwidth}@{}}
    \centering
    \begin{subfigure}{\linewidth}
        \centering
        \includegraphics[width=0.48\linewidth]{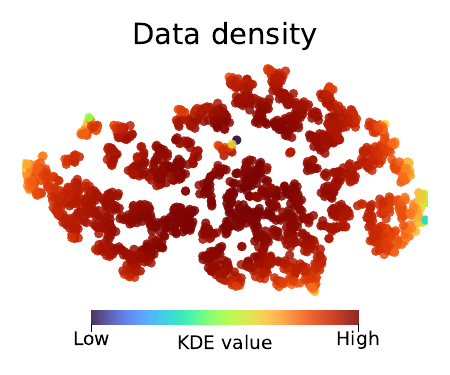}
        \includegraphics[width=0.48\linewidth]{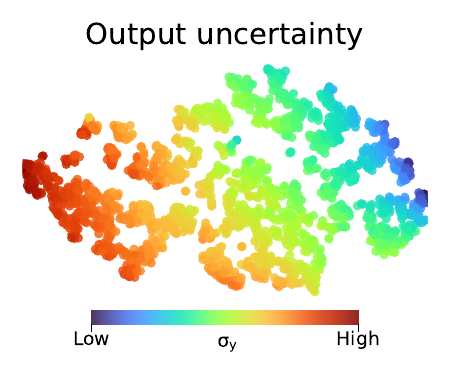}
        \rule{\linewidth}{0.4pt}
        \caption{G: Baseline} 
    \end{subfigure}
    &
    \centering{\raisebox{2.4em}{$\rightarrow$}}
    &
    \centering
    \begin{subfigure}{\linewidth}
        \centering
        \includegraphics[width=0.48\linewidth]{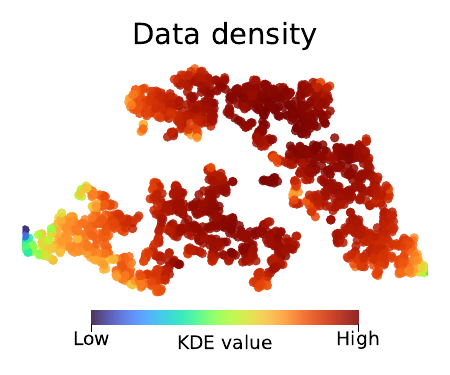}
        \includegraphics[width=0.48\linewidth]{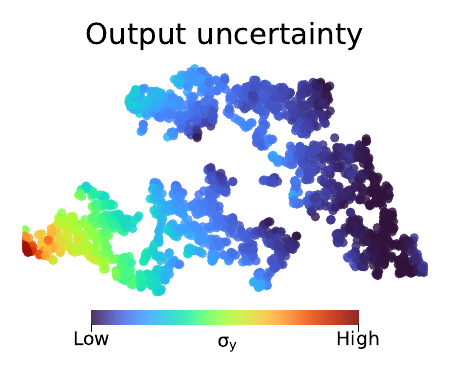}
        \rule{\linewidth}{0.4pt}
        \caption{G: AGL}  
    \end{subfigure}
    \end{tabular}
    \end{center}
    \caption{Structural calibration in data-poor regime. t-SNE dimensional reduction of latent representations is used to visualize latent geometry and compare baseline models against AGL models for five different property datasets. Each sub-figure displays two maps: data density (left, estimated via KDE) and output uncertainty (right, $\sigma_y$). Colors range from low (blue) to high (red). Left columns (a, c, e, g, i) and right columns (b, d, f, h, j) represent baseline and AGL models, respectively.}
    \label{si_fig_agl_data_poor}
\end{figure}

\begin{figure}[H]
    \centering
    \begin{center}
    \begin{tabular}{@{}m{0.46\textwidth} @{}m{0.04\textwidth} @{}m{0.46\textwidth}@{}}
    \centering
    \begin{subfigure}{\linewidth}
        \centering
        \includegraphics[width=0.48\linewidth]{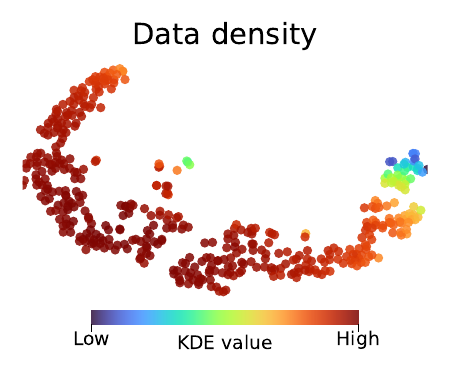}
        \includegraphics[width=0.48\linewidth]{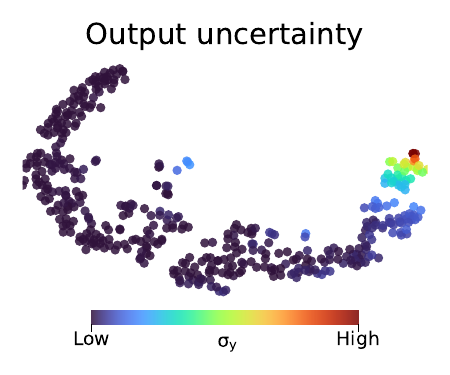}
        \rule{\linewidth}{0.4pt}
        \caption{bp: Baseline} 
    \end{subfigure}
    &
    \centering{\raisebox{2.4em}{$\rightarrow$}}
    &
    \centering
    \begin{subfigure}{\linewidth}
        \centering
        \includegraphics[width=0.48\linewidth]{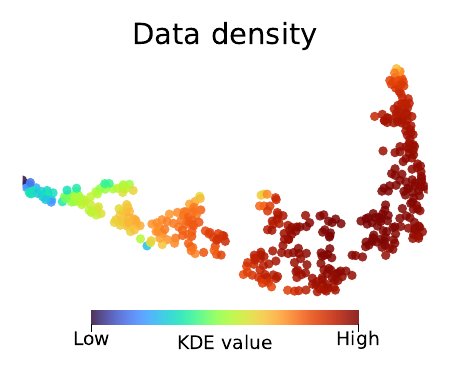}
        \includegraphics[width=0.48\linewidth]{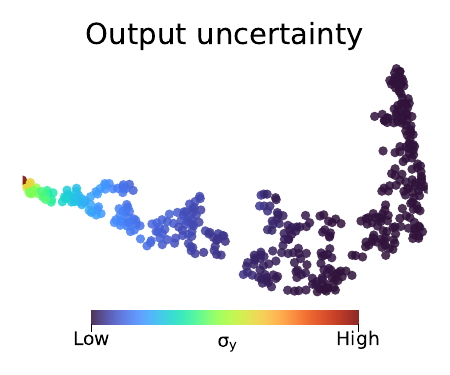}
        \rule{\linewidth}{0.4pt}
        \caption{bp: AGL}  
    \end{subfigure}
    \end{tabular}
    \end{center}

    \begin{center}
    \begin{tabular}{@{}m{0.46\textwidth} @{}m{0.04\textwidth} @{}m{0.46\textwidth}@{}}
    \centering
    \begin{subfigure}{\linewidth}
        \centering
        \includegraphics[width=0.48\linewidth]{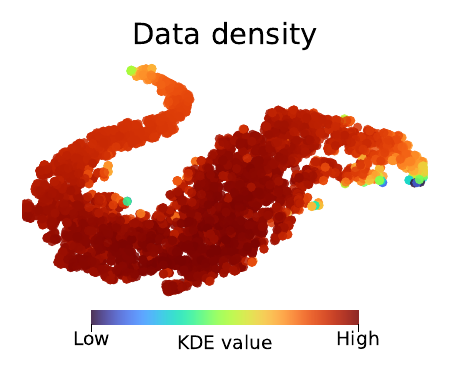}
        \includegraphics[width=0.48\linewidth]{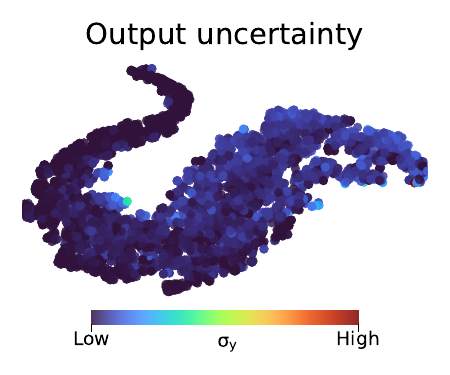}
        \rule{\linewidth}{0.4pt}
        \caption{mp: Baseline} 
    \end{subfigure}
    &
    \centering{\raisebox{2.4em}{$\rightarrow$}}
    &
    \centering
    \begin{subfigure}{\linewidth}
        \centering
        \includegraphics[width=0.48\linewidth]{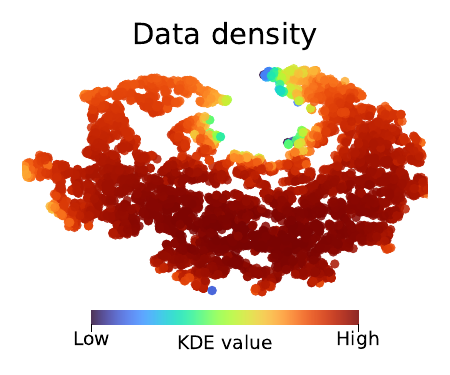}
        \includegraphics[width=0.48\linewidth]{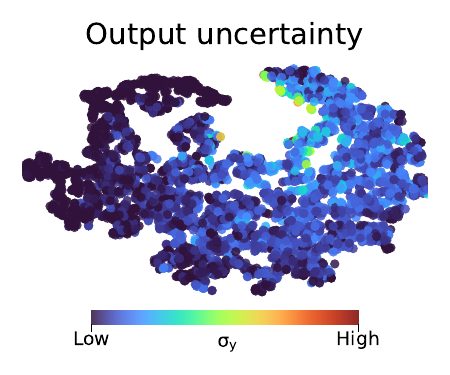}
        \rule{\linewidth}{0.4pt}
        \caption{mp: AGL}  
    \end{subfigure}
    \end{tabular}
    \end{center}

    % \begin{center}
    % \begin{tabular}{@{}m{0.46\textwidth} @{}m{0.04\textwidth} @{}m{0.46\textwidth}@{}}
    % \centering
    % \begin{subfigure}{\linewidth}
    %     \centering
    %     \includegraphics[width=0.48\linewidth]{fig/si/8_latent_tsne/logP_base_rich_density.pdf}
    %     \includegraphics[width=0.48\linewidth]{fig/si/8_latent_tsne/logP_base_rich_uncertainty.pdf}
    %     \rule{\linewidth}{0.4pt}
    %     \caption{Baseline} 
    % \end{subfigure}
    % &
    % \centering{\raisebox{2.4em}{$\rightarrow$}}
    % &
    % \centering
    % \begin{subfigure}{\linewidth}
    %     \centering
    %     \includegraphics[width=0.48\linewidth]{fig/si/8_latent_tsne/logP_align_rich_density.pdf}
    %     \includegraphics[width=0.48\linewidth]{fig/si/8_latent_tsne/logP_align_rich_uncertainty.pdf}
    %     \rule{\linewidth}{0.4pt}
    %     \caption{Alignment-guided}  
    % \end{subfigure}
    % \end{tabular}
    % \end{center}

    \begin{center}
    \begin{tabular}{@{}m{0.46\textwidth} @{}m{0.04\textwidth} @{}m{0.46\textwidth}@{}}
    \centering
    \begin{subfigure}{\linewidth}
        \centering
        \includegraphics[width=0.48\linewidth]{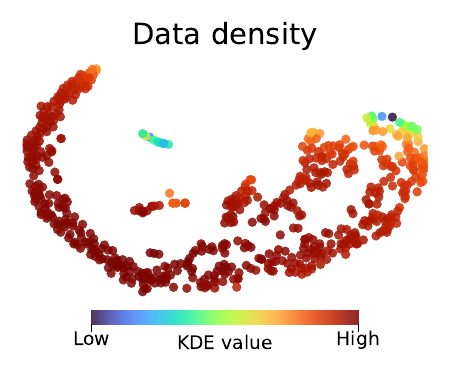}
        \includegraphics[width=0.48\linewidth]{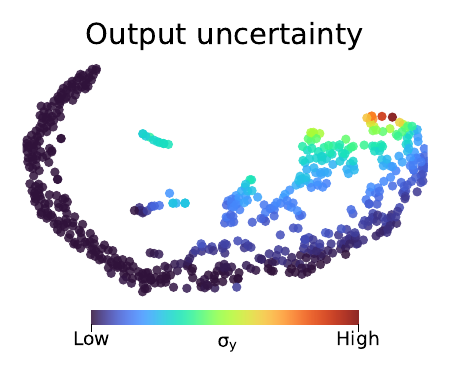}
        \rule{\linewidth}{0.4pt}
        \caption{Pvap: Baseline} 
    \end{subfigure}
    &
    \centering{\raisebox{2.4em}{$\rightarrow$}}
    &
    \centering
    \begin{subfigure}{\linewidth}
        \centering
        \includegraphics[width=0.48\linewidth]{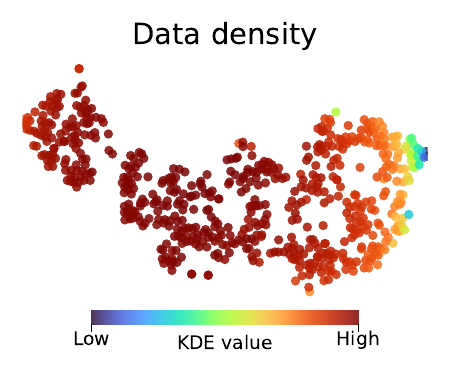}
        \includegraphics[width=0.48\linewidth]{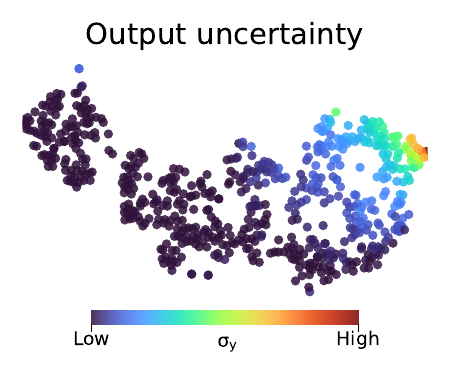}
        \rule{\linewidth}{0.4pt}
        \caption{Pvap: AGL}  
    \end{subfigure}
    \end{tabular}
    \end{center}

    \begin{center}
    \begin{tabular}{@{}m{0.46\textwidth} @{}m{0.04\textwidth} @{}m{0.46\textwidth}@{}}
    \centering
    \begin{subfigure}{\linewidth}
        \centering
        \includegraphics[width=0.48\linewidth]{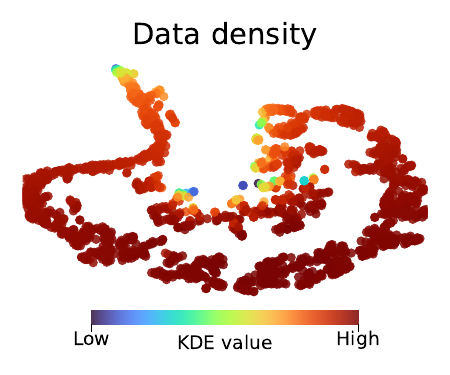}
        \includegraphics[width=0.48\linewidth]{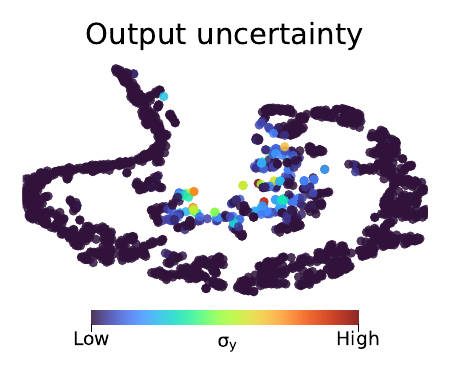}
        \rule{\linewidth}{0.4pt}
        \caption{H: Baseline} 
    \end{subfigure}
    &
    \centering{\raisebox{2.4em}{$\rightarrow$}}
    &
    \centering
    \begin{subfigure}{\linewidth}
        \centering
        \includegraphics[width=0.48\linewidth]{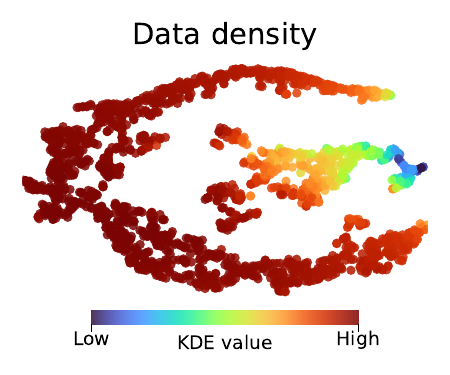}
        \includegraphics[width=0.48\linewidth]{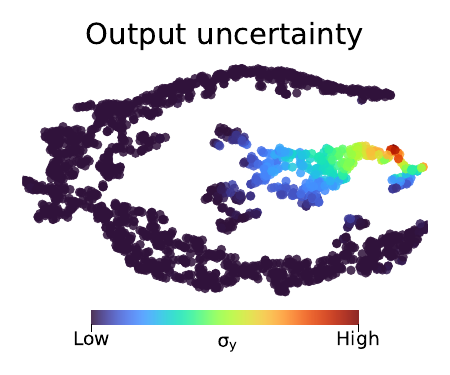}
        \rule{\linewidth}{0.4pt}
        \caption{H: AGL}  
    \end{subfigure}
    \end{tabular}
    \end{center}

    \begin{center}
    \begin{tabular}{@{}m{0.46\textwidth} @{}m{0.04\textwidth} @{}m{0.46\textwidth}@{}}
    \centering
    \begin{subfigure}{\linewidth}
        \centering
        \includegraphics[width=0.48\linewidth]{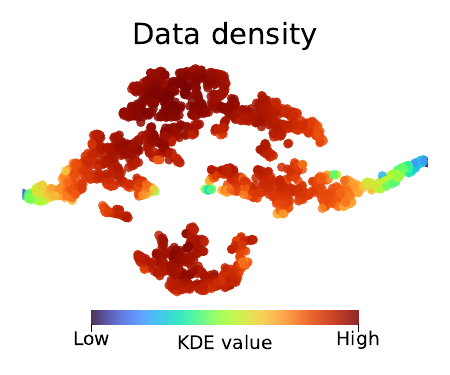}
        \includegraphics[width=0.48\linewidth]{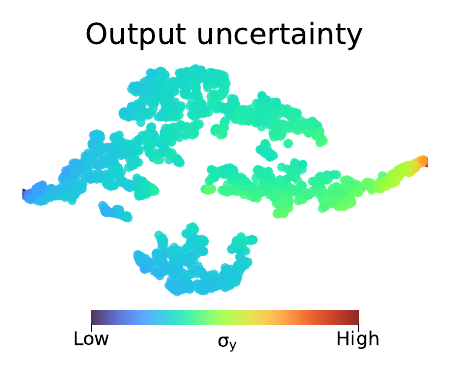}
        \rule{\linewidth}{0.4pt}
        \caption{G: Baseline} 
    \end{subfigure}
    &
    \centering{\raisebox{2.4em}{$\rightarrow$}}
    &
    \centering
    \begin{subfigure}{\linewidth}
        \centering
        \includegraphics[width=0.48\linewidth]{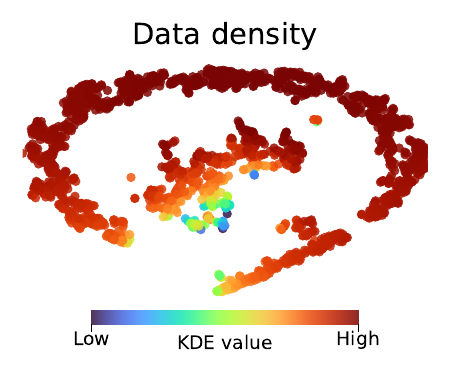}
        \includegraphics[width=0.48\linewidth]{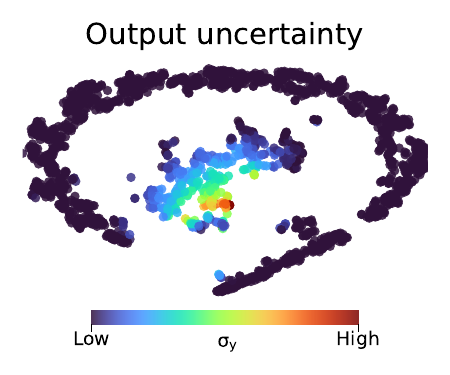}
        \rule{\linewidth}{0.4pt}
        \caption{G: AGL}  
    \end{subfigure}
    \end{tabular}
    \end{center}
    
    \caption{Structural calibration in data-rich regime. t-SNE dimensional reduction of latent representations is used to visualize latent geometry and compare baseline models against AGL models for five different property datasets. Each sub-figure displays two maps: data density (left, estimated via KDE) and output uncertainty (right, $\sigma_y$). Colors range from low (blue) to high (red). Left columns (a, c, e, g, i) and right columns (b, d, f, h, j) represent baseline and AGL models, respectively.}
    \label{si_fig_agl_data_rich}
\end{figure}

To confirm the generalizability of Alignment-Guided Learning (AGL), we applied the proposed training scheme to five additional physicochemical property datasets. We evaluated AGL using the same metrics—LPAS, MAE, posterior standard deviation $\sigma_{w}$, output uncertainty $\sigma_{y}$, and calibration indicators (ECE, DUC)—comparing results across data-poor and data-rich regimes (Figs. \ref{si_fig_agl} -- \ref{si_fig_agl_data_rich}). All metrics are normalized relative to the baseline, which is scaled to 100. Hyperparameter $\gamma$ for each dataset is given in Table \ref{si_table_lambda}. For LPAS, normalized values represent the proportional reduction in $1-\mathrm{LPAS}$ relative to the baseline, rather than the direct percentage change in raw LPAS.

As shown in Fig. \ref{si_fig_agl}, AGL consistently increases the LPAS relative to the baseline (gray dashed line) across both data-poor (light shaded bars) and data-rich (dark shaded bars) regimes, successfully reinforcing the Latent-Posterior Alignment (LPA) mechanism as intended (Figs. \ref{si_fig_align_latent_poor}-\ref{si_fig_align_latent_rich}). Consistent with the main analysis, the magnitude of this increase is more pronounced in the data-rich case (Fig. \ref{si_fig_align_latent_rich}) compared to the data-poor case (Fig. \ref{si_fig_align_latent_poor}), reflecting the availability of sufficient information to guide geometric structure.

Crucially, while predictive accuracy (MAE) remains stable without degradation, predictive uncertainty $\sigma_{y}$ is substantially reduced. This reduction occurs even though posterior uncertainty $\sigma_{w}$ remains unchanged or slightly increases. This discrepancy further demonstrates that the geometric alignment mechanism, rather than the posterior itself, is the governing factor of predictive uncertainty.

The calibration analysis reveals a distinct trade-off. While probabilistic calibration (ECE) shows mixed results or slight degradation in some cases, structural calibration (DUC) exhibits a consistent and significant enhancement. This suggests that AGL specializes in geometric calibration—aligning high uncertainty with low data density—over simple probability matching. To visualize this geometric calibration, we projected latent representations via t-SNE, mapping data density measured by Kernel Density Estimation (KDE) against output uncertainty ($\sigma_{y}$) as shown in Figs. \ref{si_fig_agl_data_poor}-\ref{si_fig_agl_data_rich}. The projections reveal that AGL harmonizes the distributions: regions of low data density (blue) consistently correspond to high uncertainty (red), and vice versa.

Crucially, the dynamics of this improvement differ by regime. In the data-poor case (Fig. \ref{si_fig_agl_data_poor}), the baseline model exhibits a disjoint and quasi-random uncertainty landscape compared to data density. Here, AGL fundamentally corrects this relationship, enforcing an alignment between uncertainty and data density. Meanwhile, in the data-rich case (Fig. \ref{si_fig_agl_data_rich}), the baseline models already show slight alignment. Here, AGL refines and sharpens the relationship in boundary regions, yielding a stricter correspondence between uncertainty and data sparsity. These strong improvements in DUC and structural alignment imply that AGL is particularly advantageous for downstream tasks like Bayesian optimization and active learning, which rely on the fundamental assumption that uncertainty serves as a reliable proxy for data scarcity (epistemic ignorance).

\end{document}